%% file: journal.tex
\documentclass[preprint]{elsarticle}

\newcommand{\etal}{\textit{et al.~}}
\newcommand{\norm}[1]{\left\lVert#1\right\rVert}
\usepackage{hyperref}
\usepackage{subfig}
\usepackage{graphicx}
\usepackage{url}
\usepackage{color}
\usepackage{multirow}
\usepackage{amsmath}
\usepackage{xspace}
\usepackage{rotating}
\usepackage{caption}
\usepackage{floatrow}
\usepackage{url}
\usepackage{lettrine}

\biboptions{sort}

\usepackage{tikz}
\usetikzlibrary{shapes,arrows}

\usepackage{enumerate}

\usepackage{tikz}
\def\checkmark{\tikz\fill[scale=0.4](0,.35) -- (.25,0) -- (1,.7) -- (.25,.15) -- cycle;}

\begin{document}
	
\begin{frontmatter}
\title{A Large-scale Dataset and Benchmark for Similar Trademark Retrieval}

\author[myu]{Osman Tursun\corref{cor1}}
\ead{wusiman.tuerxun@ceng.metu.edu.tr}

\author[myu]{Cemal Aker}
\ead{cemal@ceng.metu.edu.tr}

\author[myu]{Sinan Kalkan}
\ead{skalkan@ceng.metu.edu.tr}

\address[myu]{KOVAN Research Lab, Computer Engineering, Middle East Technical University}

\cortext[cor1]{Corresponding author}

\begin{abstract}
Trademark retrieval (TR) has become an important yet challenging problem due to an ever increasing trend in trademark applications and infringement incidents. There have been many promising attempts for the TR problem, which, however, fell impracticable since they were evaluated with limited and mostly trivial datasets. 
In this paper, we provide a large-scale dataset with benchmark queries with which different TR approaches can be evaluated systematically. Moreover, we provide a baseline on this benchmark using the widely-used methods applied to TR in the literature. Furthermore, we identify and correct two important issues in TR approaches that were not addressed before: reversal of contrast, and presence of irrelevant text in trademarks severely affect the TR methods. Lastly, we applied deep learning, namely, several popular Convolutional Neural Network models, to the TR problem. To the best of the authors, this is the first attempt to do so.

%
%
\end{abstract}


\end{frontmatter}
%

\input{section/introduction.tex}
\input{section/the_metu_dataset.tex}
\input{section/methods.tex}
\input{section/experiments_and_results.tex}
\input{section/conclusion.tex}

\appendix
\input{section/appendix_dist.tex}

\section*{References}
\bibliographystyle{abbrv}
\bibliography{reference}

\end{document}

%% file: section/introduction.tex
\section{Introduction}
A trademark is a recognizable symbol or associated text that identifies products or services
of an individual, a business organization or a legal entity from those of others. Registered
trademarks are viewed as a form of legitimate property and needs to be protected from brand piracy and
trademark infringement. To protect and legalize their trademarks, owners have to register
their trademarks in patent offices in many countries. More than 100 million companies are
known to exist in local and global markets\footnote{See \url{http://www.econstats.com/wdi/wdiv_494.html} for related statistics.}, and many of them own at least one registered
trademark. According to Word Intellectual Property Organization \cite{WIPO}, 3 million trademark
registrations exist worldwide and trademark applications keep increasing at a rate of
6-8\% in recent years.

\begin{figure*}[!hbt]
	\centering
	\subfloat[]{
		\includegraphics[width = 0.10\linewidth]{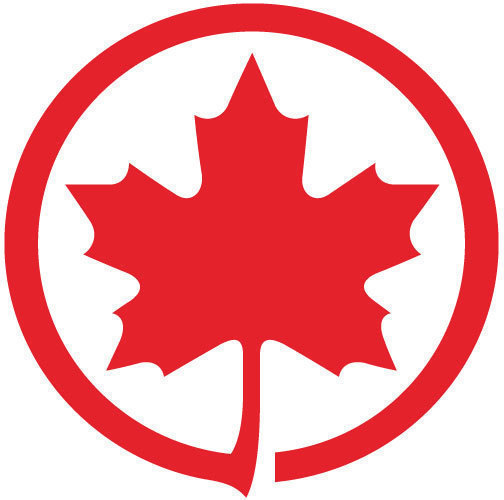}
		\includegraphics[width = 0.10\linewidth]{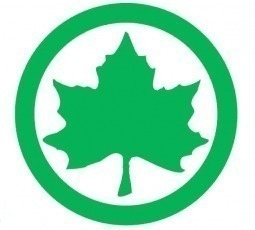}
	}
	\hspace {0.8cm}
	\subfloat[]{
		\includegraphics[width = 0.10\linewidth]{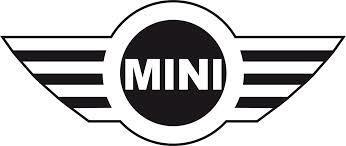}
		\includegraphics[width = 0.10\linewidth]{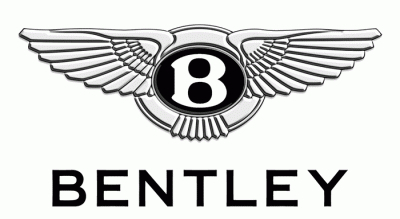}
	}
	\hspace {0.8cm}
	\subfloat[]{
		\includegraphics[width = 0.12\linewidth]{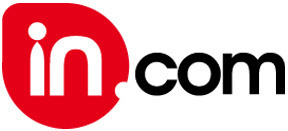}
		\includegraphics[width = 0.12\linewidth]{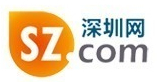}
	}
	\\
	\subfloat[]{
		\includegraphics[width = 0.12\linewidth]{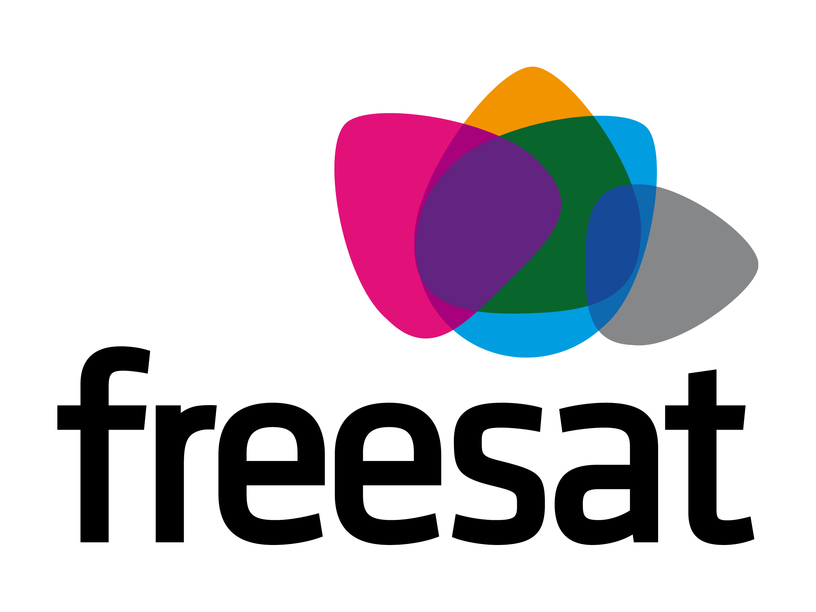}
		\includegraphics[width = 0.12\linewidth]{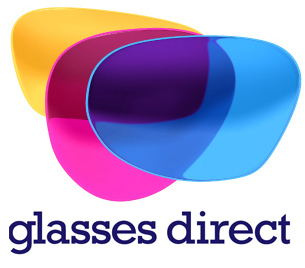}
	}
	\hspace {0.8cm}
	\subfloat[]{
		\includegraphics[width = 0.12\linewidth]{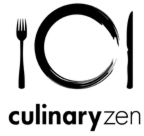}
		\includegraphics[width = 0.12\linewidth]{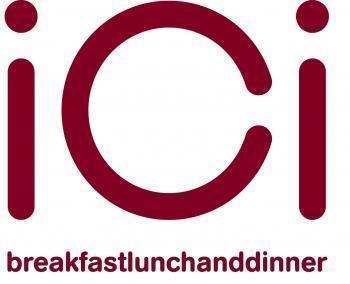}
	}
	\hspace {0.8cm}
	\subfloat[]{
		\includegraphics[width = 0.12\linewidth]{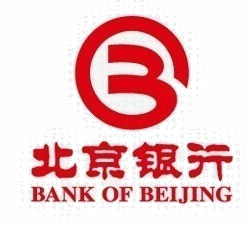}
		\includegraphics[width = 0.12\linewidth]{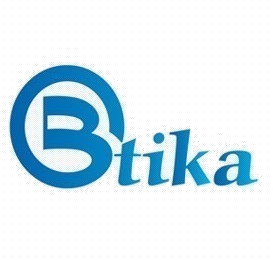}
	}
	\\
	\subfloat[]{
		\includegraphics[width = 0.12\linewidth]{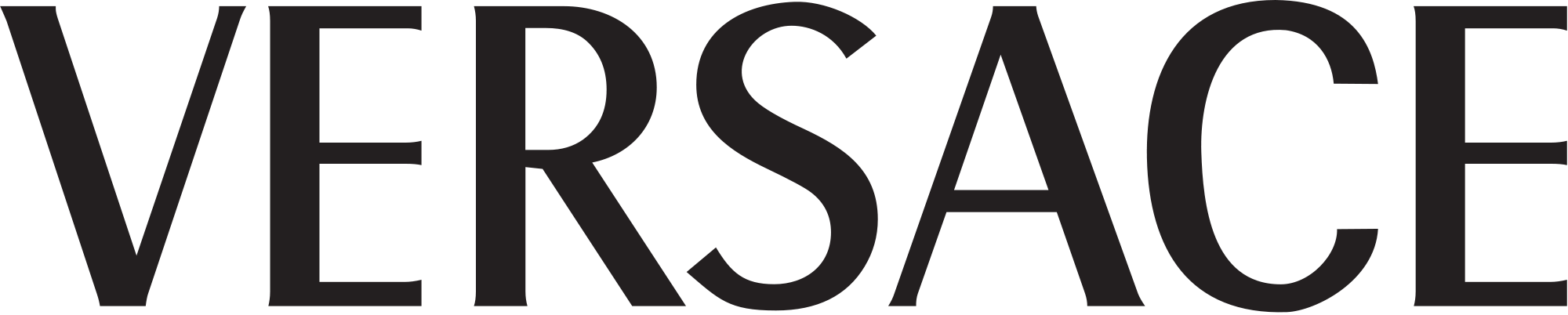}
		\includegraphics[width = 0.12\linewidth]{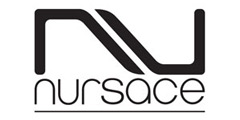}
	}
	\hspace {0.8cm}
	\subfloat[]{
		\includegraphics[width = 0.08\linewidth]{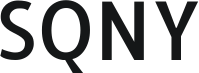}
		\includegraphics[width = 0.15\linewidth]{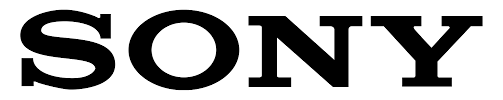}
	}
	\hspace {0.8cm}
	\subfloat[]{
		\includegraphics[width = 0.10\linewidth]{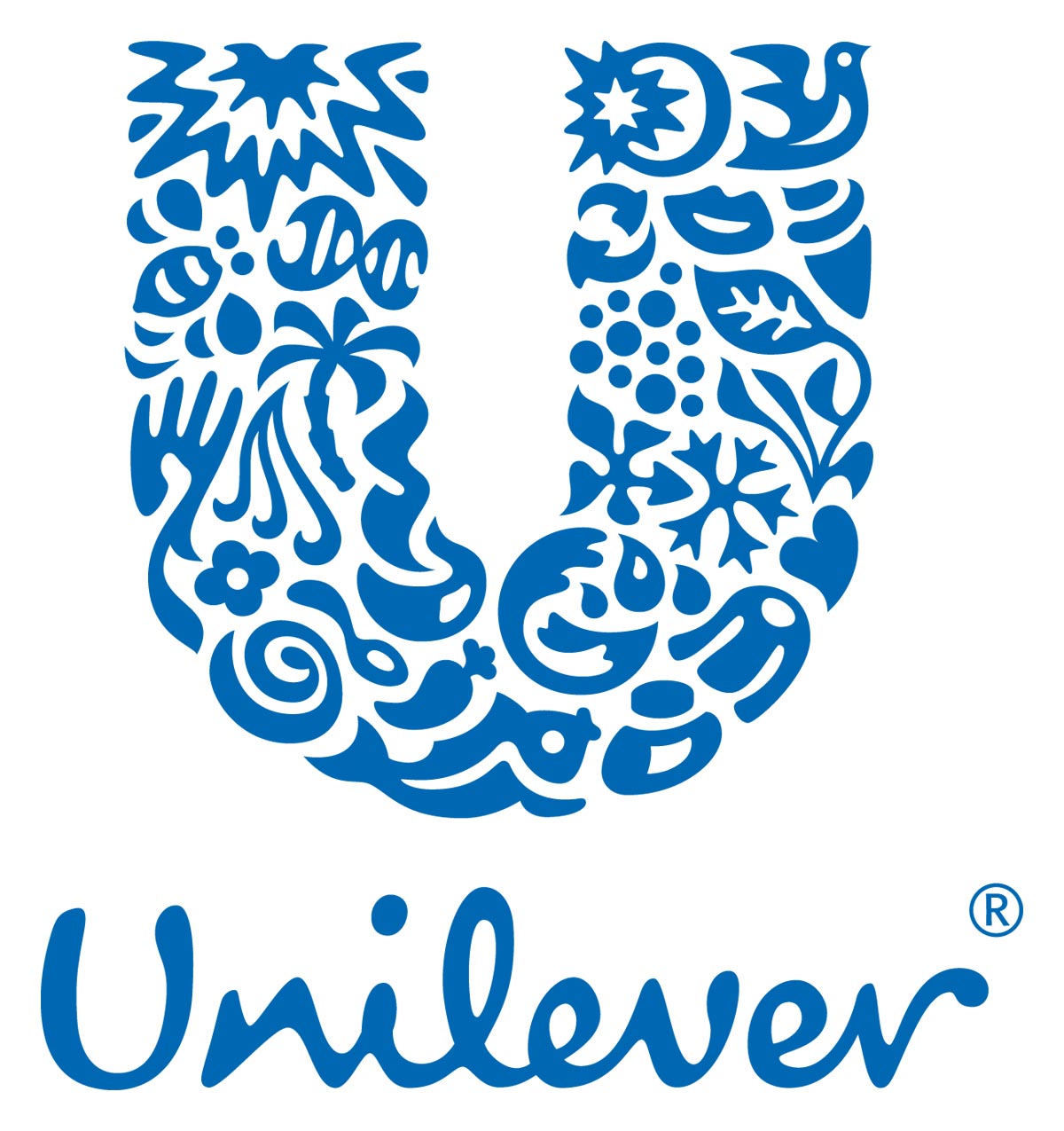}
		\includegraphics[width = 0.10\linewidth]{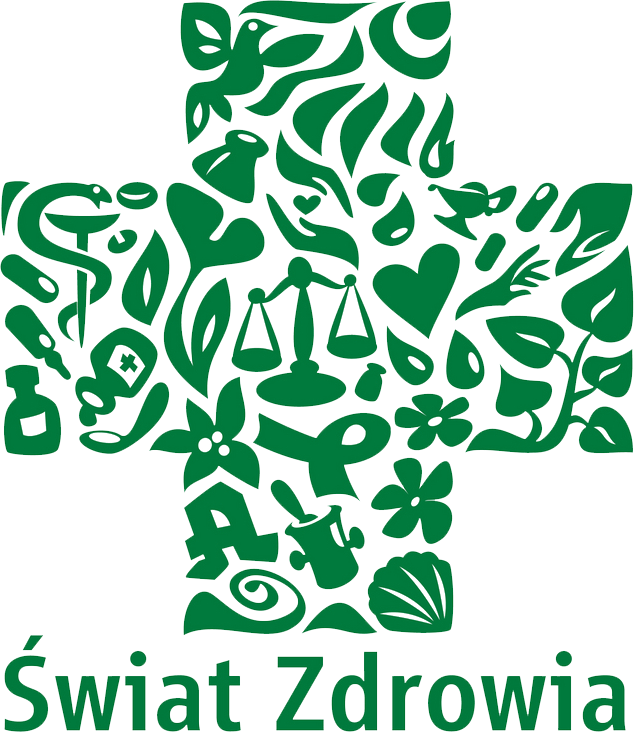}
	}
		
	\caption{Sample trademarks and trademark similarities.}
	\label{fig:vis_sam}
\end{figure*}

Upon application of a new trademark, it needs to be made sure that the new trademark does not imitate or is dissimilar enough from existing trademarks. In most developed countries, organizations like patent offices take the responsibility of protecting trademarks from encroachment. To avoid various infringements, they exclude registration
of near-duplicate or intentionally imitated trademarks by manually checking trademarks in
the database or by using TR systems. Massive amounts of registration have overwhelmed
both manual and automatic operations and reduced service quality of patent offices, which
leaves an open space for trademark infringements. What is worse, two mistakenly registered
similar trademarks will increase the complexity of handling legal disputation between owners.
To ease the burdens of patent offices, a robust automated trademark retrieval (TR) system with intelligent image analyzing techniques is imperative.

\begin{figure}[ht]
	\centering
	\subfloat[Text only mark]{
		\includegraphics[width = 0.3\textwidth]{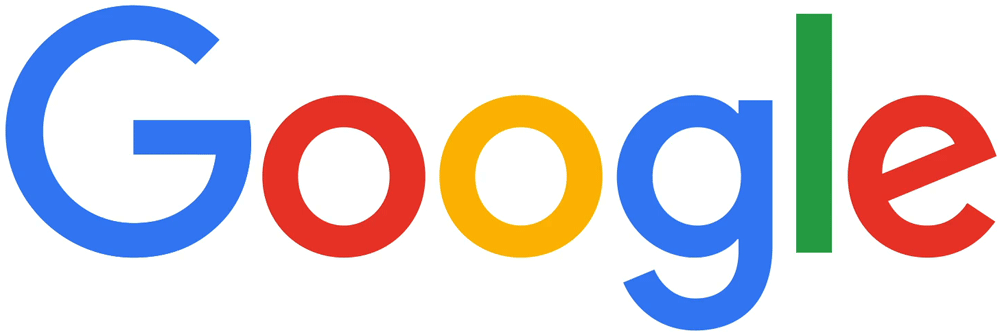}
		\label{fig:yasmeen}}
	\subfloat[Figure only mark]{
		\includegraphics[width = 0.3\textwidth]{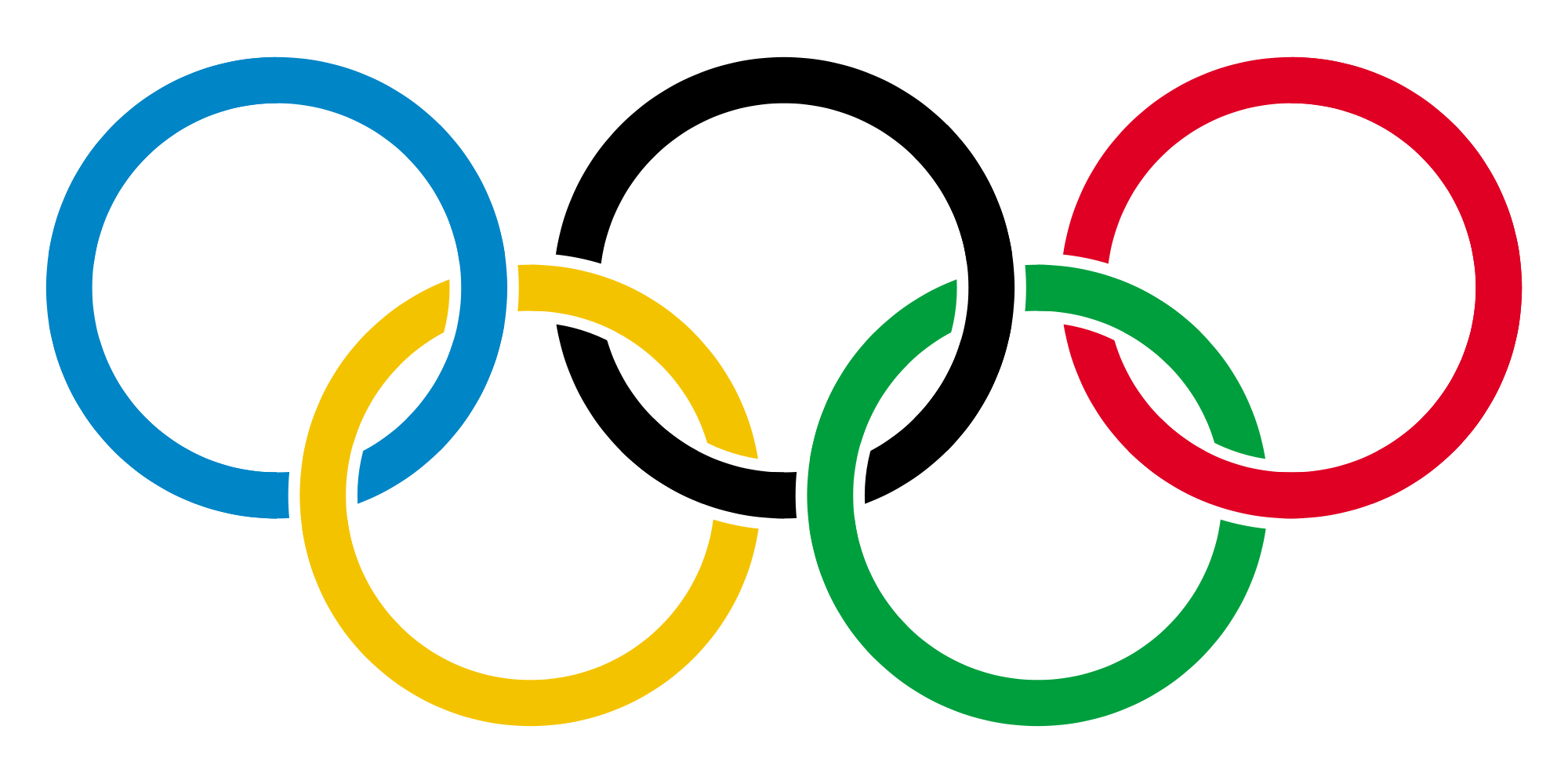}
		\label{fig:apple}}
	\subfloat[Figure and text mark]{
		\includegraphics[width = 0.3\textwidth]{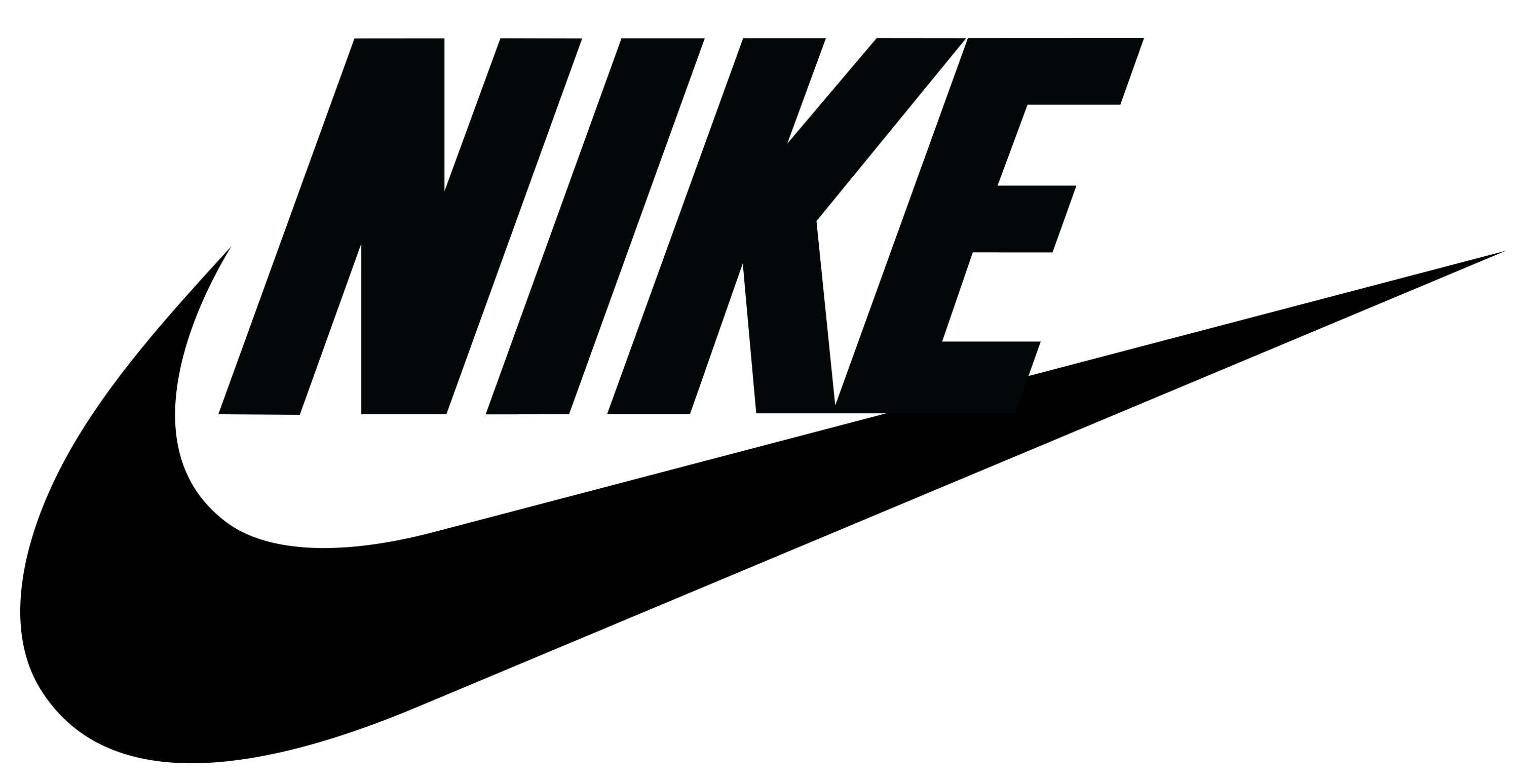}
		\label{fig:bass}}
	\caption{Examples of different trademark types.}
	\label{fig:logo sample}
\end{figure}

However, retrieving all trademark similarities in an efficient and effective way is challenging since: 
\begin{enumerate}[i.]
	\item similarity, even when constrained to visual aspects, is eluding since it can occur at many different levels, either visually or semantically -- see Figure \ref{fig:vis_sam} for some samples. For example, two trademarks can be deemed similar based on the textu(r)al content, the way a line is shaped or placed, or the combinations of such low-level visual content -- see Figure \ref{fig:logo sample}. 
	
	\item similarity is subjective mainly due to the lack of clear criteria for deciding similarity. Visual similarity, especially in the case of trademark similarity, can be influenced greatly by many aspects including education background, religion, hobbies etc. 
		
	Another important factor affecting similarity is the fact that the amount of existing trademarks is tremendous and rapidly increasing, which poses a challenge for the creation of new, substantially different trademark for expressing the same content. This, in time, may lead to a shift in deciding similarity since we may run out of ways to express a meaning.
	
	\item until recently \cite{tursun2015metu}, there was no large trademark dataset available to see the challenges of the problem and evaluate the methods. With this paper, we hope to extend our previous work \cite{tursun2015metu} -- see Section \ref{sect:contributions} for the details.
	
	\item Available image retrieval methods, which are mostly tailored towards defining similarity in terms of object-related features, are not optimal solutions for trademark retrieval problems, since figures of trademarks mostly incorporate abstract information with various transformations and amounts of detail. 	In fact, trademark retrieval systems should be equipped with high-level visual capabilities like visual grouping, object recognition, scene/content understanding etc. to be able to handle cases like the ones in Figure \ref{fig:gestalt}.
	
\end{enumerate}

\subsection{Contributions}
\label{sect:contributions}
In this paper, we focus on trademark similarity defined in terms of visual similarity (see Figure \ref{fig:vis_sam} for examples) and skip the conceptual/semantic similarities (see, e.g., \cite{kesidis2014logo,anuar2013conceptual} for some attempts). Visual similarities of trademarks includes color, shape and texture aspects.

In this work, we extend our previous work \cite{tursun2015metu,tuerxun2015comparison} and make the following contributions:

\begin{itemize}
	\item \textbf{A large-scale dataset and a benchmark}:  We had already introduced the dataset in our previous work \cite{tursun2015metu}. However, the dataset has been extended with more trademarks and better query samples with which trademark retrieval systems can be tested and compared.
	
	\item \textbf{An analysis of visual features and a baseline}: We apply on our dataset many widely-used hand-crafted features (including local and global descriptors based on color, texture and shape -- including color histogram, shape context, LBP, SIFT, SURF, GIST, etc.) as well as deep features (AlexNet \cite{NIPS2012_4824}, GoogleNet \cite{GoogLeNet} and VGG-net \cite{Simonyan14c}) that have been shown to perform well on many challenging image recognition tasks. In fact, to the best of our knowledge, this is the first study that has applied deep learning to the trademark retrieval problem. Moreover, we have tested fusion of the best features to see whether they can perform better when combined.
	
	The performances of the methods reveal that the trademark retrieval problem is very challenging (even for deep learning), and in fact, it should attract more attention than it does in the computer vision and pattern recognition community. 
	
	\item \textbf{An analysis of the aspects}: We identified that the methods were impeded by the presence of text, or inverse contrast change. To overcome these limitations, we have proposed and tested several methods.
\end{itemize}

To be more specific, the current paper differs from our previous work \cite{tursun2015metu,tuerxun2015comparison} in (i) the dataset, and (ii) the methods tested. Namely, the current paper includes deep learning methods, and the improvement of performance of the methods through handling text and contrast change separately.

\subsection{Organization}

Section \ref{sect:dataset} introduces the METU trademark dataset. We present our large scale trademark dataset and compare it with other related datasets. Section \ref{sect:methods} describes the methods evaluated in this work. These methods are divided into two main groups: traditional hand-crafted features and deep features.
In Section \ref{sect:experiments}, the setup and configurations of our experiments are given. Finally, Section \ref{sect:conclusion} concludes the paper with an outline of future work.

%
%
%
%

\begin{figure}
	\centering
	\subfloat[WWF logo] {
		\includegraphics[width = 0.4\textwidth]{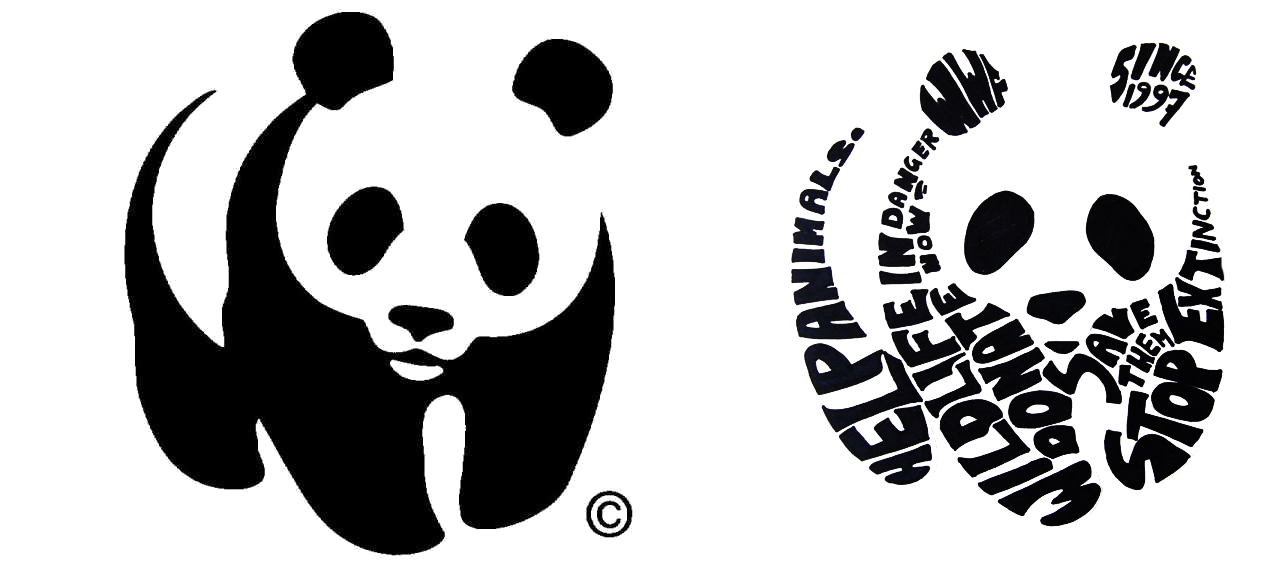}
		\label{fig:tripan}
	}
	\hspace{1cm}
	\subfloat[IBM logo] {
		\includegraphics[width = 0.4\textwidth]{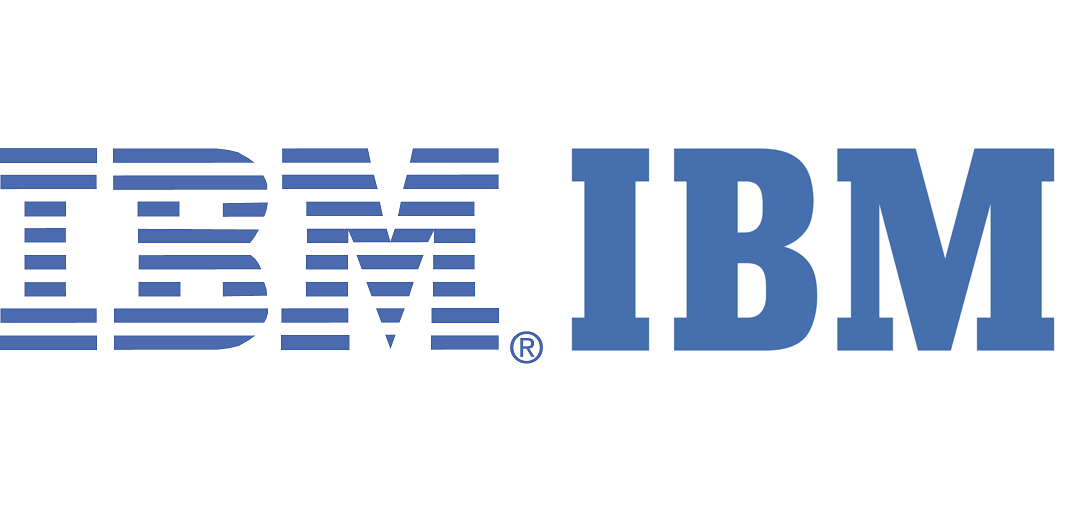}
		\label{fig:ibm}
	}
	\caption{Example of how Gestalt principles affect trademark perception.}
	\label{fig:gestalt}
\end{figure}
  
%

%

\section{Related Studies}
In this section, we discuss the current approaches to trademark similarity, including the manual methods.

\begin{figure}[htb]
	\centering
	\subfloat[Sample part of Vienna classification categories.] {
	\includegraphics[width = 0.99\textwidth]{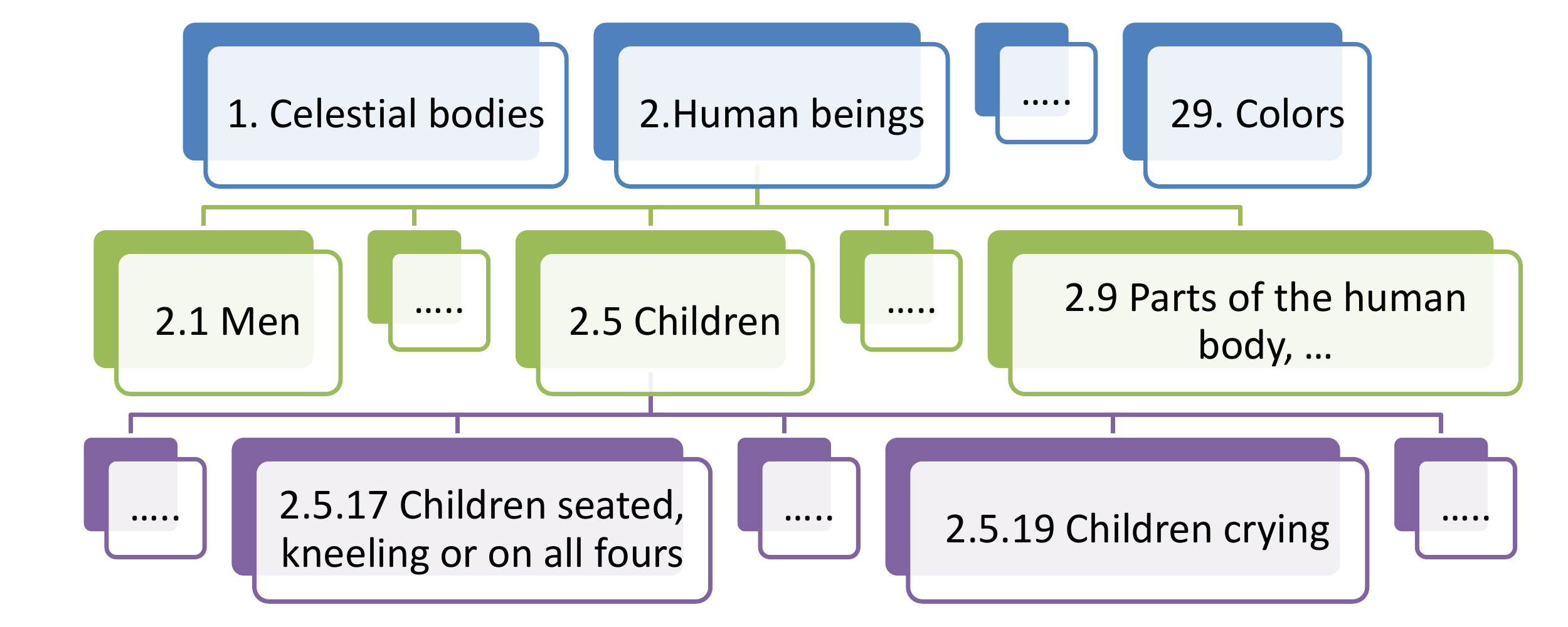}
	\label{fig:vcat1}
	}
	\\
	\subfloat[Vienna code: 2.5.19] {
		\includegraphics[width = 0.3\textwidth]{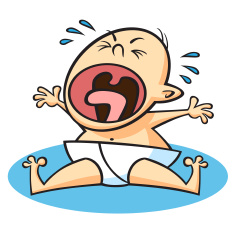}
		\label{fig:baby1}
	}
	\hspace{1cm}
	\subfloat[Vienna code: 2.5.17] {
		\includegraphics[width = 0.35\textwidth]{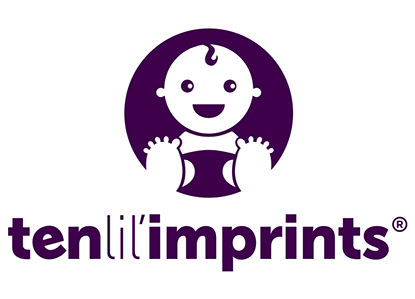}
		\label{fig:baby2}
	}
	\caption{Vienna classification categories (a) and sample codes (b-c).
	\label{fig:vcat}}
\end{figure}

\subsection{Checking Trademark Similarity Using Manual Methods}

All patent offices still rely on manual effort for evaluating trademark similarity. Such efforts can be fully manual or semi-manual: In a fully manual approach, a human first memorizes a trademark and then skims through the whole collection of trademarks to hopefully spot similarities. In the semi-manual approach, a human first labels the trademarks, retrieves trademarks with the same labels, and visually inspects similarity among the retrieved trademarks. 

The accepted standard for the labeling approach is the Vienna classification system, which uses a hierarchy of categories (as displayed in Figure \ref{fig:vcat1}) for labeling the trademarks. For example, the Vienna code (category) for a trademark including human-beings is 2, and 5 as a sub-category for a baby. Based on what the baby is doing, further sub-categories can be attached: e.g., 17 and 19 stand for sitting and crying babies respectively -- see Figure \ref{fig:vcat}. When queried with a trademark for similarity, first the trademark is labeled with the Vienna categories, then the trademarks with these categories are retrieved, and similarity is evaluated by an expert using visual inspection.

Although Vienna classification system is practical compared to a fully-manual approach, it bares several disadvantages: (i) The classification process is subjective since the detail of labeling can depend on the observer. (ii) The categories are fixed and not expendable. (iii) It is not possible to describe in words all the content of a trademark, as in the famous saying ``a picture is worth a thousand words''. 

In short, manual approaches, even sophisticated labeling systems such as the Vienna classification, are (i) unreasonably time-demanding (in the fully-manual case, it takes 3-4 days for a human expert to visually inspect a trademark among approx. 1 million trademarks), (ii) quite error prone since humans are involved in the process, and (iii) unpractical for a trademark system that is rapidly growing with new trademarks. Therefore, automatizing trademark similarity is necessary.


\subsection{Checking Trademark Similarity Using Automated Methods}
Although patent offices still rely on manual methods, researchers have been working on fully automated methods for similar trademark retrieval for around two decades. 

Early attempts applied low/medium level global features, including graphic feature vectors \cite{kato1992}, Fourier descriptors \cite{wu1996content, hsieh2001multiple, eakins2001comparison}, image moments \cite{wu1996content,jain1998shape, eakins2001comparison, ciocca2001content}, Zernike moments \cite{Wei2009, Kim1998, zhang2012technique} as well as simpler and lower-cost shape features such as aspect ratio \cite{eakins2001comparison}, circularity \cite{eakins2001comparison}, Rosin descriptor \cite{eakins2001comparison}, angular radial transform \cite{eakins2001comparison}, gray level projection \cite{wu1996content}, gradient orientation histogram \cite{jain1998shape, ciocca2001content}, wavelets \cite{ciocca2001content}, triangle area representation \cite{alajlan2007retrieval, alajlan2006multi} -- see Table \ref{tab:literature} for an overview. In addition to shape and texture related features, color-feature based approaches have also been applied for trademark retrieval \cite{lam1996star, phan2010content, rusinol2011interactive, zeggari2010trademarks, rusinol2010perceptual}. 

Jiang \etal \cite{jiang2006gestalt} pointed out that the aforementioned descriptors do not incorporate geometric information of the extracted features. These descriptors will fail in cases where trademarks match each other at partial parts or unrelated trademarks lead to similar global descriptors. To improve retrieval results, various combinations of these features have been applied. Although there is contrasting evidence \cite{eakins2003shape}, effective integration of multiple features has been shown to improve retrieval performance \cite{her2011hybrid, jain1998shape}. 

To improve retrieval results and the partial matching problem, one approach is to segment trademarks to several sub-objects and match trademarks by comparing their part descriptors \cite{Alwis1998, alwis1999trademark, Alwis2000, Eakins1998, eakins2001comparison, eakins2003shape, jain1998shape, Leung}. However, segmentation is an ill-posed problem, and looking at cases like those in Figure \ref{fig:gestalt}, a promising approach should rely on employing perceptual organization and grouping mechanisms similar to Gestalt principles \cite{wertheimer1938laws}. Some common Gestalt principles like similarity, continuation, closeness, proximity and etc. have already been incorporated into trademark retrieval systems by Eakins \etal \cite{eakins2001comparison, eakins2003shape, Eakins1998}, Alwis \etal \cite{Alwis1998, Alwis2000}, and Jiang \etal \cite{jiang2006gestalt}.


Describing trademarks with global features extracted either from the whole trademark or its parts is time and memory efficient. However, these methods ignore local information, which can be important in addressing partially infringement issues. In order to include local information for addressing partial matching, key-point based methods such as SIFT \cite{kochakornjarupong2011trademark, wei2009trademark, lin2008trademark}, Harris corners \cite{zeggari2010trademarks}, etc. have been tested.

\begin{table}[!ht]
	\centering
	\caption{Shape-based trademark retrieval methods in the literature.}
	\vspace{-0.2cm}
	\resizebox{\textwidth}{!}{
		\begin{tabular}{l|l|l}
			\hline
			{\bf Group} & {\bf Approach}  & {\bf Study}\\ 
			\hline \hline
			\multirow{5}{4cm}{\it{Transform- and moment-based shape features}}  & Fourier descriptors 	  & \cite{wu1996content, hsieh2001multiple, eakins2001comparison} \\ 
			& Moment variants 		  & \cite{wu1996content,jain1998shape, eakins2001comparison, ciocca2001content} \\ 
			& Zernike moments 		  & \cite{Wei2009, Kim1998, zhang2012technique} \\ 
			& Wavelets			&\cite{ciocca2001content}\\
			& Angular radial transform			&\cite{eakins2001comparison} \\
			\cline{1-3}
			\multirow{6}{4cm}{\it{Simple and low-cost shape features}}  	& Aspect ratio	& \cite{eakins2001comparison} \\ 
			& Circularity	& \cite{eakins2001comparison,Wei2009} \\
			& Convexity		& \cite{eakins2001comparison, zhang2012technique}\\
			& Compactness	& \cite{zhang2012technique}\\
			& Eccentricity	& \cite{zhang2012technique, alajlan2007retrieval} \\
			& Distance to centroid	& \cite{Wei2009} \\
			& Rosin descriptor	(triangularity,  				&\multirow{2}{*}[0cm]{\cite{eakins2001comparison}} \\
			& rectangularity and ellipticity)						& 		\\
			& Triangle area representation (TAR)  &\cite{alajlan2007retrieval, alajlan2006multi} \\
			\cline{1-3}
			\multirow{3}{4cm}{\it{Histogram or relation-based shape features}}	& Gray level projection			&\cite{wu1996content}\\
			& Gradient orientation histogram (edge direction)	&\cite{jain1998shape, ciocca2001content} \\
			& Shape-context		&\cite{rusinol2010perceptual, rusinol2010efficient}\\
			\hline
		\end{tabular}
	}
	\label{tab:literature}
\end{table}
\subsection{Related Problems: Trademark Detection and Recognition}

Trademark detection and recognition are two problems, which are related to trademark retrieval. Trademark detection is the problem of finding all trademarks in a scene. On the other hand, trademark recognition is interested in finding a specific trademark in the scene -- see Kesidis \etal \cite{kesidis2014logo} for a very detailed survey about these problems. 

Kesidis \etal \cite{kesidis2014logo} point out that the difference between similarity and matching is subtle but critical to trademark retrieval, since most of the image retrieval methods are designed for exact match rather than detecting similarity. For example, keypoint-based methods rely on having the same keypoints being detected and matched. However, in a similarity problem, two trademarks may not own any common key-points.  

%% file: section/the_metu_dataset.tex
\section{The METU Trademark Dataset}
\label{sect:dataset}

Existing trademark retrieval studies were conducted on small scale and limited (only consist of special types of logos) datasets, some of which are listed in Table \ref{tbl:datasets}. Despite their valuable contributions and prominent results, their practicality, efficiency and reliability can only be confirmed on large scale datasets. For this end, in \cite{tursun2015metu}, we shared a very challenging trademark dataset, the METU Trademark dataset, for benchmarking the trademark retrieval problem.

\begin{figure*}[hbt!]
	\centering
	\subfloat[Dataset samples]{
		\includegraphics[width = 0.9\linewidth]{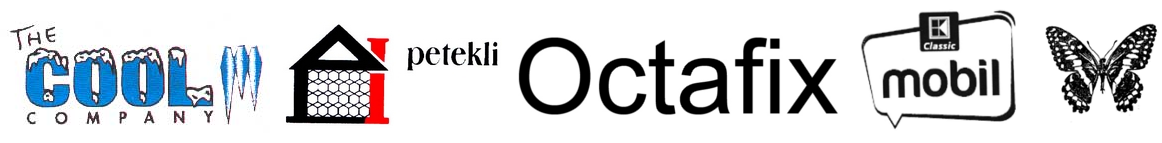}
		\label{fig:ds}} \\
	\subfloat[Example set for similar trademarks]{
		\includegraphics[width = 0.9\linewidth]{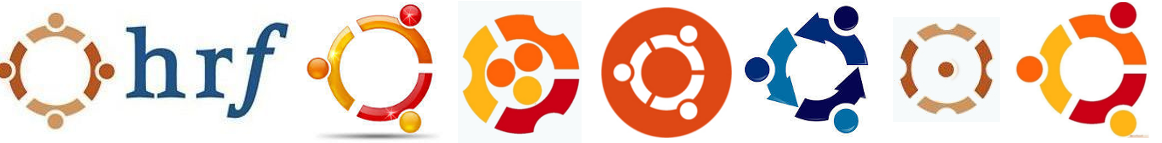}
		\label{fig:qs1}} \\
	\subfloat[Another example set for similar trademarks]{
		\includegraphics[width = 0.9\linewidth]{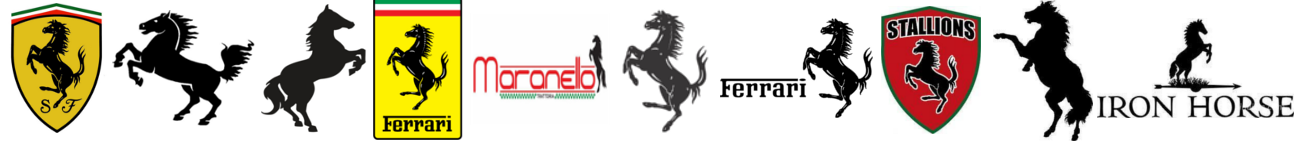}
		\label{fig:qs2}}
	\caption{Logo samples from the METU dataset. (a) Arbitrary samples. (b) Sample set for similar trademarks. 
		(c) Another Sample set for similar trademarks. 
		\label{fig:metu_dataset}}
\end{figure*}

In our previous works \cite{tuerxun2015comparison, tursun2015metu}, we shared the first version of the dataset and conducted several experiments on it. The first version included 930,328 logos, 320 of which belonged to a ``query set" for which an expert had identified similar logos already. These query logos are divided into 32 groups. Query logos in the same group are similar to each other. For convenience, here, we name the 930,008 logos as test-set and the 320 query logos as query-set. Figure \ref{fig:ds} shows that various types of logos from query-set and test-set. The METU trademark dataset is composed of logos belonging to around 410,000 companies. The test-part of the dataset is provided by the patent office ``Grup Ofis Marka Patent A.\c{S}." \footnote{\url{http://www.grupofis.com.tr}}, and ``query set" is constructed through collecting and enriching trademark infringement cases appearing in the market.  We have performed ``cleaning'' operations like auto-cropping, filtering corrupted and low-quality trademarks to make our dataset more suitable for academic research.

With this article, we share the second version of the METU trademark dataset. The update includes removal of duplicate logos, and addition of new similar logos in the test-set. As a result, 6,985 logos were removed from the dataset, and the query set is extended to 35 groups, where each group contains around 10-15 trademarks. In total, the query-set contains 417 logos. Figure \ref{fig:qs1} and \ref{fig:qs2} are examples of query samples. Detailed comparison of the first and second versions is given in Table \ref{tbl:details}.

The updated dataset is available on-line for research purposes \cite{metudataset}.

%
\begin{table}[hbt!]
	\begin{center}
		\caption{Details of the METU dataset. \label{tbl:details}}
		\vspace{-0.2cm}
		\footnotesize
		\begin{tabular}{l l l}\hline
			\textbf{Aspect}  &  \textbf{Version 2} & \textbf{Version 1}\\
			\hline \hline
			\# trademarks & 923,343 & 930,328\\
			\# query sets & 417 & 320\\
			\# unique registered firms & 409,675 & 410,439\\
			\# unique trademarks & 687,842 & 690,418\\
			\# trademarks containing text only & 583,715 & 589,098\\
			\# trademarks containing figure only &  19,214 & 19,387\\
			\# trademarks containing figure and text  &  310,804 & 311,986\\
			\# trademarks with unknown contents & 9,610 & 9,857\\
			\# file format & JPEG & JPEG\\
			\# Max Resolution & $1,800\times1,800$(px) & $1,800\times1,800$(px)\\
			\# Min Resolution & $30\times30$(px) & $30\times30$(px)\\
			\hline
		\end{tabular}
	\end{center}
\end{table}
%

\subsection{Comparison with Other Datasets}
A comparison with the available datasets is provided in Table \ref{tbl:datasets}. To the best of our knowledge, the METU trademark dataset is the largest, organized and challenging publicly available dataset. Compared to other datasets used in previous studies, the METU TR benchmark dataset is very realistic, both at size and types of the trademarks aspects. 

There is also a raw dataset, called USTPO Trademark application bulk dataset\footnote{Available at https://www.google.com/googlebooks/uspto-trademarks-application-images.html}, which also contains millions of trademarks. However, before using it in trademark retrieval, it needs substantial amount of preprocessing (for removing duplicates, non-cropping cases and getting additional useful information like types, texts of trademarks, etc.). 
\begin{table}[hbt!]
	\caption{A comparison of trademark datasets available in the literature.}
	\small
	\begin{tabular}{l c c c c c}
		\hline
		\multirow{2}{*}{\bf Dataset}  & \bf{Number of } &\bf{Requires} 		& \bf{Image}  & \bf{Image} 			&\multirow{2}{*}{ \bf Ref.} \\ 
									  & \bf logos 		& \bfseries{preprocessing?}						& \bf type 	  & \bf{size (px)}  	& \\ 
		\hline \hline	
		UM   				& 106 	 & No 	& BW 	& various 		&\cite{UMD} \\
		MPEG7 CE2B			& 3,621  & No	& BW 	& - 			& \cite{Wang2012}\\
		Wei \etal			& 1,003  & No	& BW 	& $200\times200$ 		& \cite{Wei2009}\\
		$Alwis$ \etal  		& 210    & No	& BW 	& 				&\cite{Alwis1998} \\ 
		$Alwis$ \etal  		& 1,000 	 & No	& BW 	& -				&\cite{Alwis2000} \\
		$abdel$ \etal       & 63,718 & No	& BW 	& -				& \cite{abdel2000fast} \\
		MPEG7 CE2B			& 1,400	 & No	& BW	&256$\times$256	& \cite{qi2010effective}\\
		MPEG7 				& 3,000	 & No	& BW	& -				& \cite{jiang2006gestalt}\\
		Jain \etal			& 1,100	 & No	& BW	& $200\times200$ & \cite{jain1998shape, ciocca1999similarity}\\
		UKTR 				& 10,745 & No   & BW	& -				&\cite{Eakins1998} \\
		Leung \etal	    	&2,000	 & No 	& BW	& -				&\cite{Leung} \\
		Her \etal			& 2,020	 & No 	& RGB	& $64\times64$	&\cite{her2011hybrid} \\ 	
		USPTO				& $\sim$1,500,000 & Yes 	&RGB	& various		&\cite{USTPODATA}\\
		\bf{METU} 	& \bf{923,343} 	& \textbf{No} & \bf{RGB} & \bf{various} & \bf{\cite{tursun2015metu}}\\
		
		\hline
	\end{tabular}
	\label{tbl:datasets}
\end{table}

%% file: section/methods.tex
\section{Methods}
\label{sect:methods}

In this section, we introduce the visual features tested on our dataset. We group the features into two broad categories based on whether they are hand-crafted or learned using deep learning methods. Moreover, we present how we can fuse the best features to obtain better results. 

%
\subsection{Hand-crafted Features}
Hand-crafted features are designed based on ``expert" knowledge and experience on the problem at hand. These designed features try to capture different aspects of what is available in an image. These aspects include color, shape, texture etc., which can be analyzed locally or globally.

In the following, we first introduce color features, then discuss global shape and layout-based features. After that, we will describe the key-point features, which are good at capturing partial similarity.    

\subsubsection{Color Feature: Color Histogram}

Color is a widely-used integral property of trademarks, giving them an extra dimension for expressing information. As pointed out by Her \etal \cite{her2011hybrid}, color schemes of trademarks are not only attractive to customers, but also protected through additional registration processes \cite{kesidis2014logo}. 

Color similarity of trademarks is determined usually by comparing their color histograms. Color histogram is a short summary of the distribution of color in the trademarks. It is translation and scale invariant (when normalized properly). However, most of the time, color is not sufficient to identify similarity, which is mostly due to shape similarities; therefore, color is generally used together with other features  \cite{lam1996star, phan2010content, rusinol2011interactive, zeggari2010trademarks, rusinol2010perceptual}.

The efficiency and effectiveness of the color histogram method is dependent on the color space, quantization, distance measures and normalization methods used. In our previous work \cite{tuerxun2015comparison}, we experimented with two most widely used color spaces: RGB and HSV. 

Due to these crucial differences between the two color spaces, two different quantization methods are used: The RGB color space is uniformly quantized into 64 or 512 different colors by dividing each of its color channels to 4 or 8 parts. However, our choice of quantizing the HSV color space is not uniform (to see the necessity for this better: looking at the 3D cylindrical model of the HSV color space, one finds that the bottom part is black while the top part is colorful. These colors in the black region make little difference to human eyes \cite{lei1999cbir}. According to this observation, nonuniform quantization methods have been proposed \cite{lei1999cbir, guohui1999image, smith1999integrated}.

As to the distance measures and normalization methods, we chose five different distance metrics: Euclidean, Cosine, Intersection, Quadratic, and Manhattan distances, and L1 and L2 normalizations -- see \ref{sec:appendixA} and \ref{sec:appendixB} for a definition of distance metrics and the normalization methods.

\begin{figure}[tp]
	\centering
	\subfloat[The PR graph of RGB color histograms of 64 bins]{
		\includegraphics[width = 0.45\linewidth]{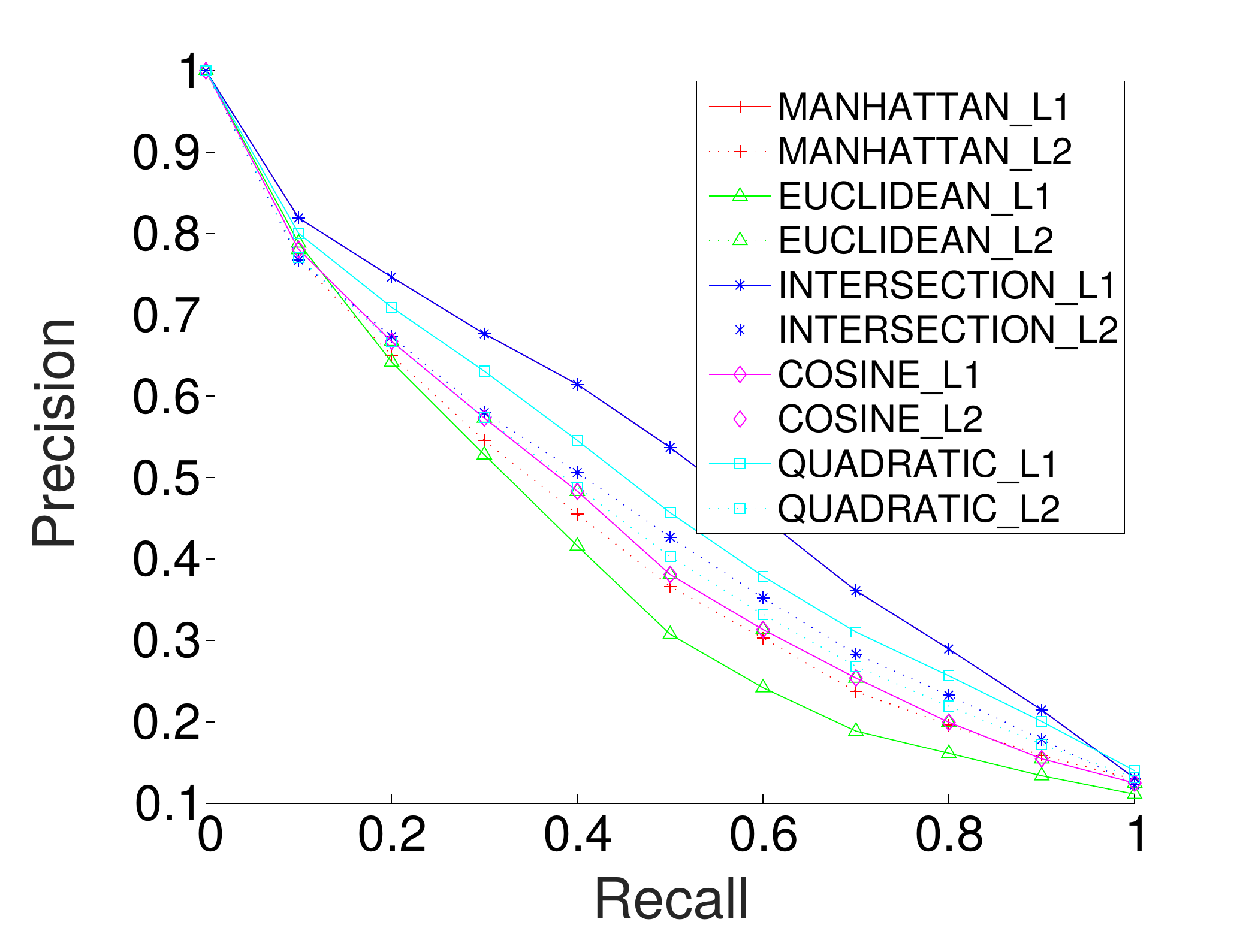}
		\label{fig:rgb_l1}}
	\hfill
	\subfloat[The PR graph of RGB color histograms of 512 bins]{
		\includegraphics[width = 0.45\linewidth]{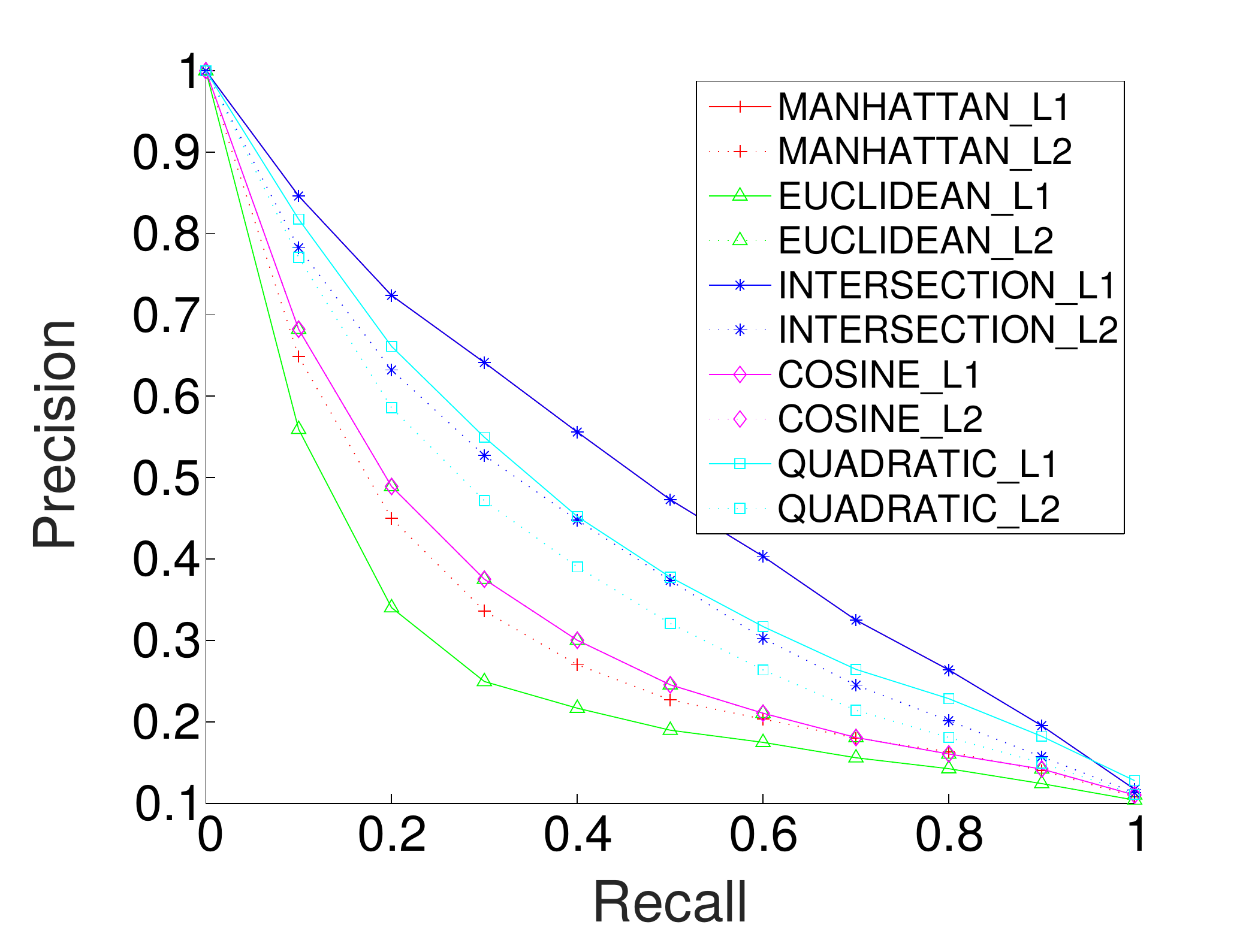}
		\label{fig:rgb_l2}} \\
	\subfloat[The PR of HSV color histograms of 36 bins]{
		\includegraphics[width = 0.45\linewidth]{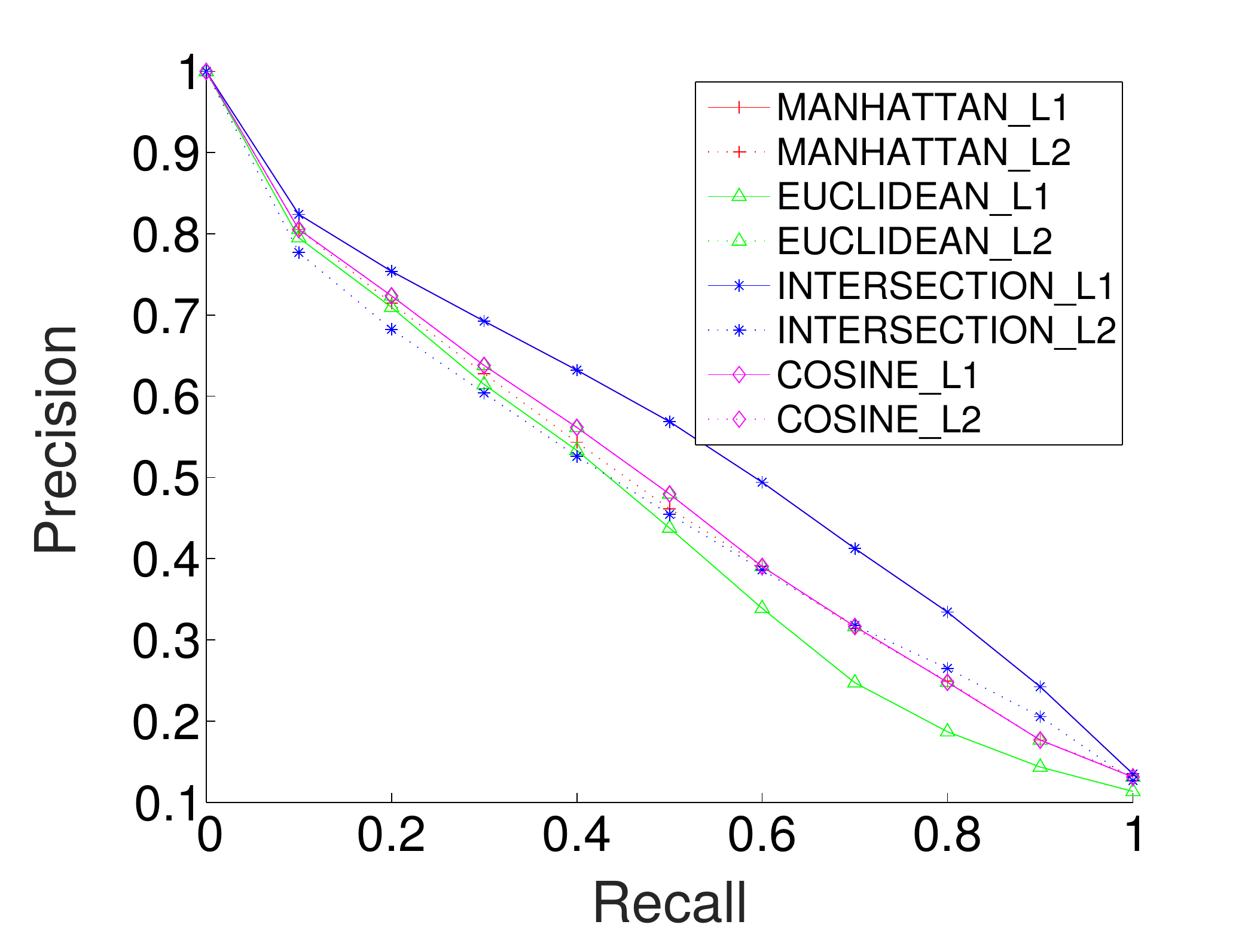}
		\label{fig:hsv_l1}} 
	\hfill
	\subfloat[The PR of HSV color histograms of 72 bins]{
		\includegraphics[width = 0.45\linewidth]{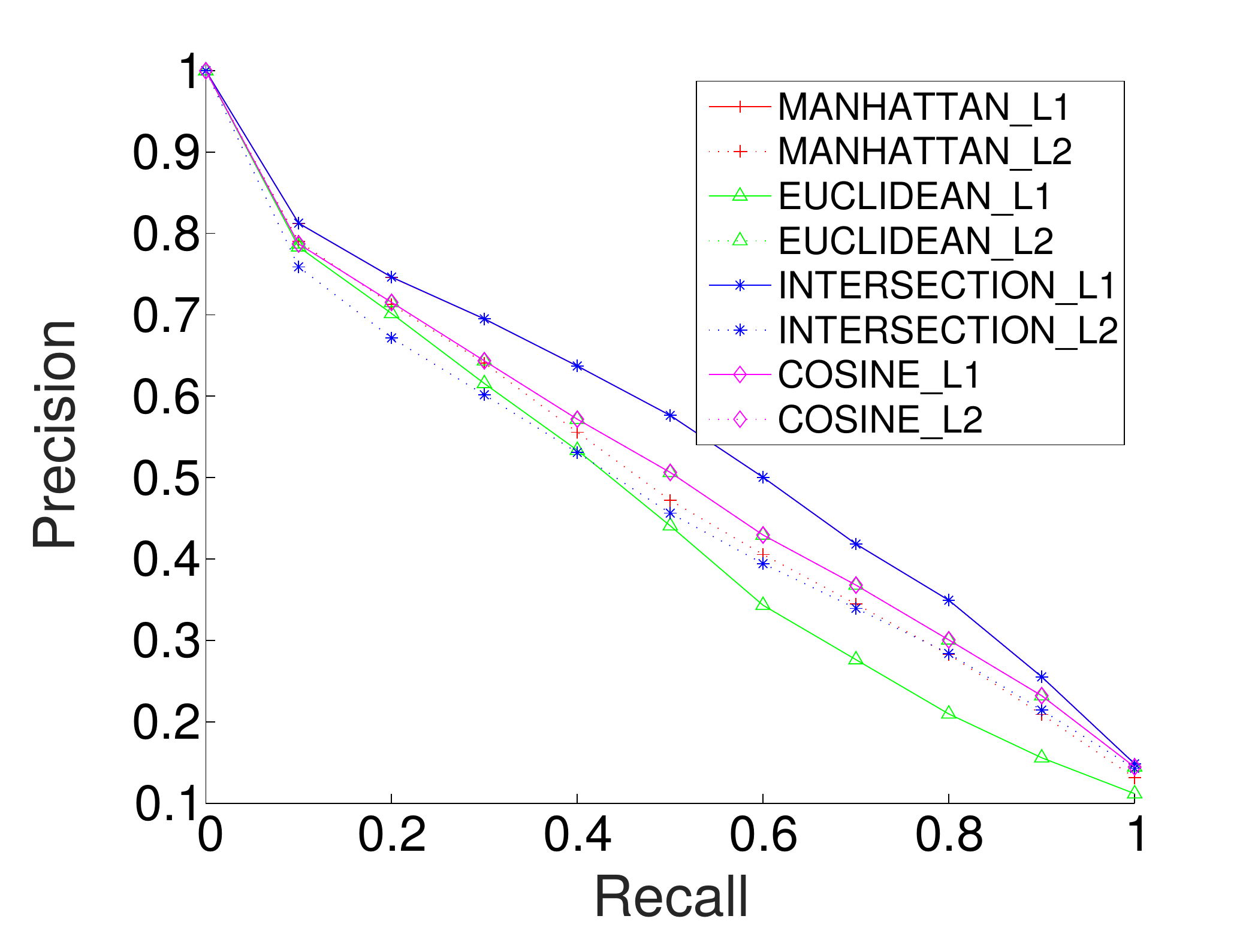}
		\label{fig:hsv_l2}} \\
	\subfloat[The comparison of outstanding schemes from (a-d)]{
		\includegraphics[width = 0.45\linewidth]{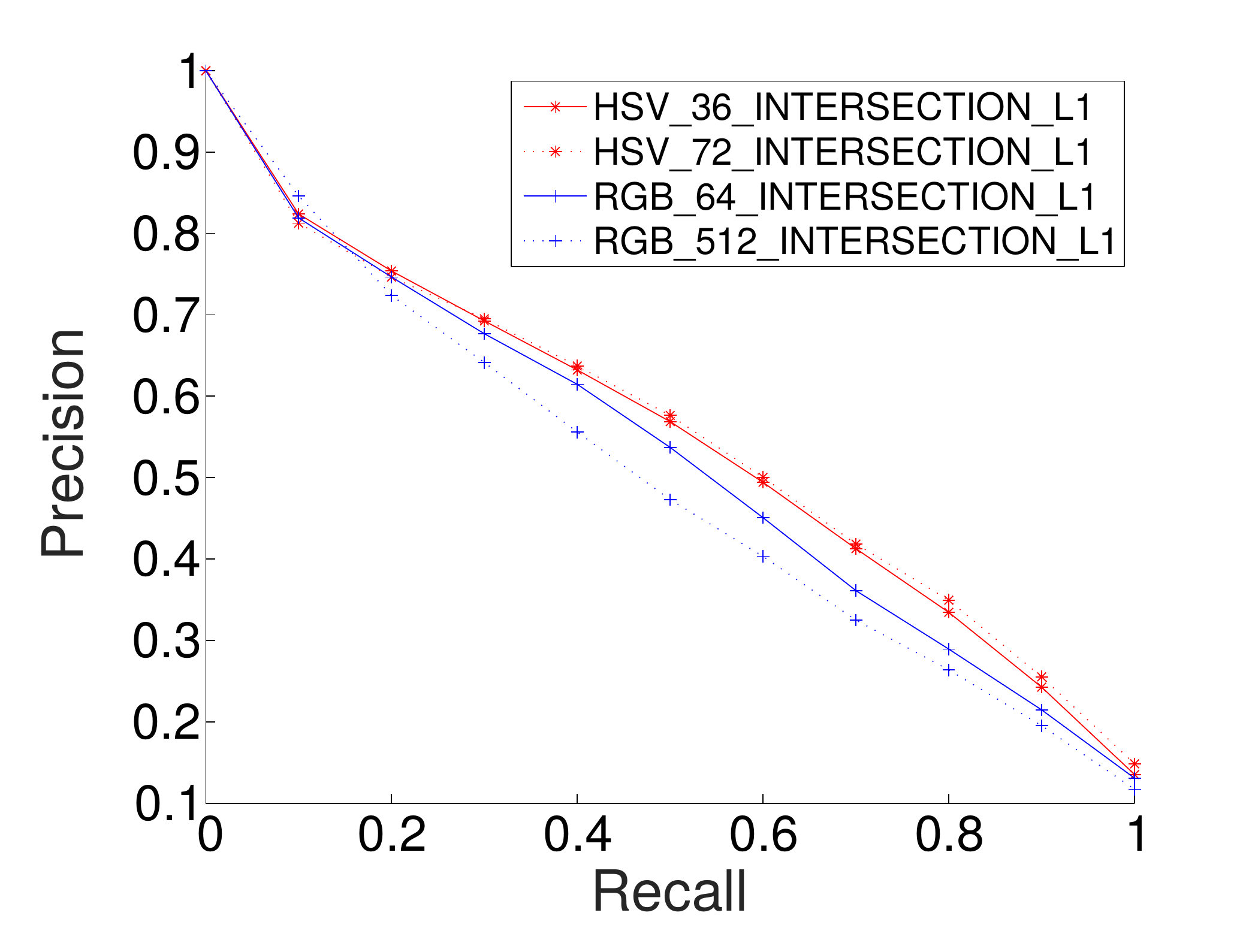}
		\label{fig:rgb_hsv_cp}}
	\caption{The effects of the parameters in color-based trademark retrieval in a small colorful subset of the METU dataset, grouped by the utilized normalization scheme and color space. (a-d) The results of RGB color histograms of 64 and 512 bins and HSV color histograms of 36 and 72 bins, compared for various distance metrics. (e) A comparison of the best overall results. The numeric prefixes in the legend entries denote the number of quantization bins, while the string suffixes indicate the utilized distance metric and normalization.}
	\label{fig:ls}
\end{figure}

As shown in Figure \ref{fig:ls}, we selected the best parameter settings on a small subset of our dataset. This subset includes 600 colorful trademarks in 10 different colors: red, green, blue, cyan, yellow, pink, black, gray, orange and brown. From this investigation, we found the following setting to perform best: HSV color space with 72 bin normalization (same as \cite{lei1999cbir}), intersection distance method, and L1 normalization return the best retrieval results. In the rest of the article, we adopt these settings for the color feature.

\subsubsection{Texture Feature: Local Binary Patterns (LBP)}

Texture is an important cue in evaluating similarity of trademarks, and for representing textural content of an image, Local binary patterns (LBP) \cite{ojala2001generalized, ojala2002multiresolution} is a popular, simple and efficient choice. LBP extracts structural patterns from images by comparing the intensity of a pixel with $N$ neighbors around it in a certain radius. Patterns are outcomes of comparisons in the $N$ bit binary number format. The statistics of occurrences of each pattern in an image is then expressed as a $2^N$-bin vector. Given the LBP vectors of two images (trademarks), their textural similarity can be queried using the distance between their LBP vectors.

Ojala \etal \cite{ojala2001generalized} generalized LBP with the following expression,
\begin{equation}
LBP_{P,R} = \sum_{p=0}^{P-1}{s(g_p - g_c)2^P},
\end{equation}
where $P$ is the count of neighbors in a circle with radius $R$, and $g_p$ and $g_c$ are intensities of pixel $p$ and the center pixel respectively, and $s(x)$ is equal to $0$ when $x$ is more than or equal to $1$, otherwise $0$. 

\begin{figure}[btp!]
	\centering
	\subfloat[The PR graph of $LBP_{P,r}$]{
		\includegraphics[width = 0.45\linewidth]{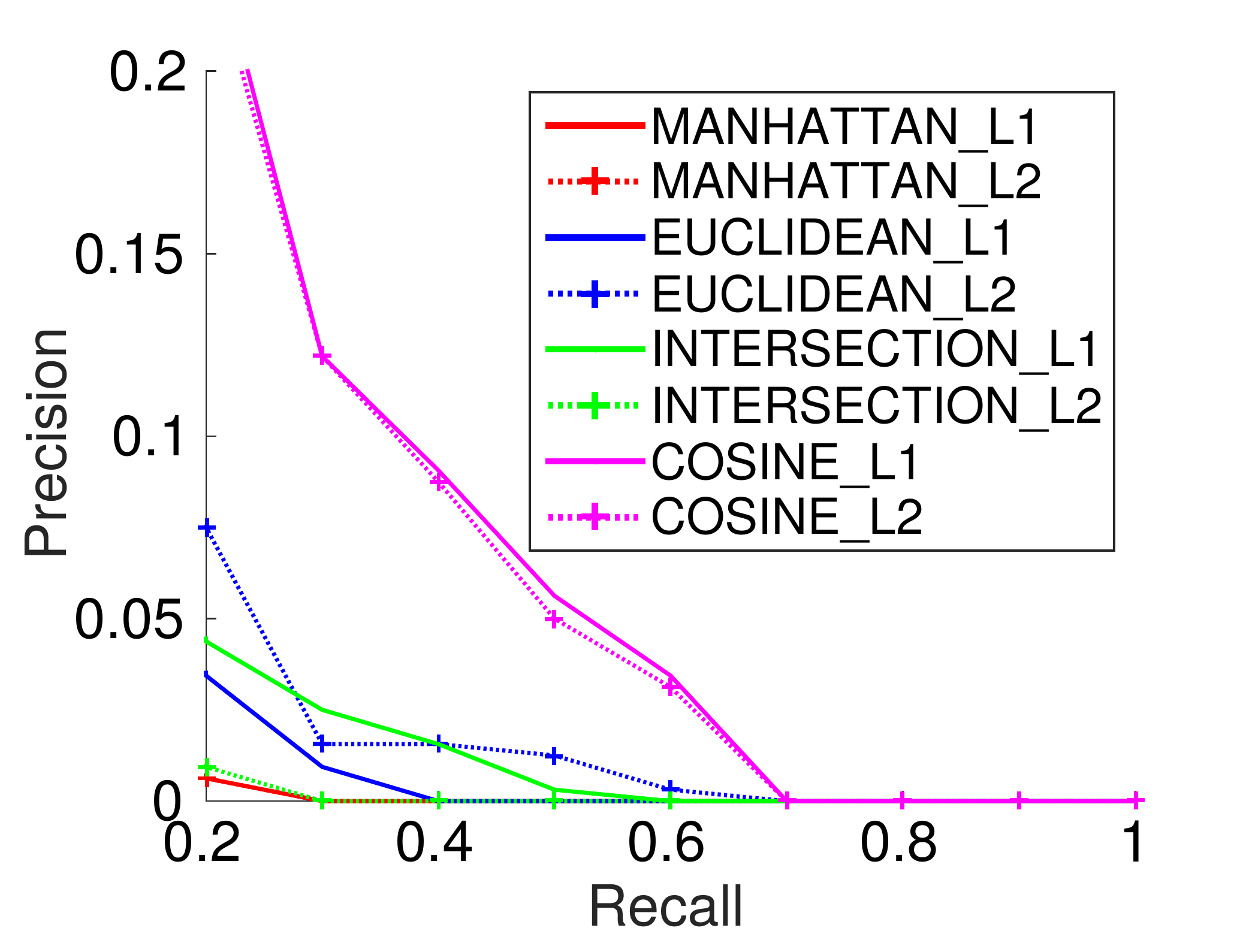}
		\label{fig:lbp_nr}} 
	\subfloat[The PR graph of $LBP_{P,r}^{ri}$]{
		\includegraphics[width = 0.45\linewidth]{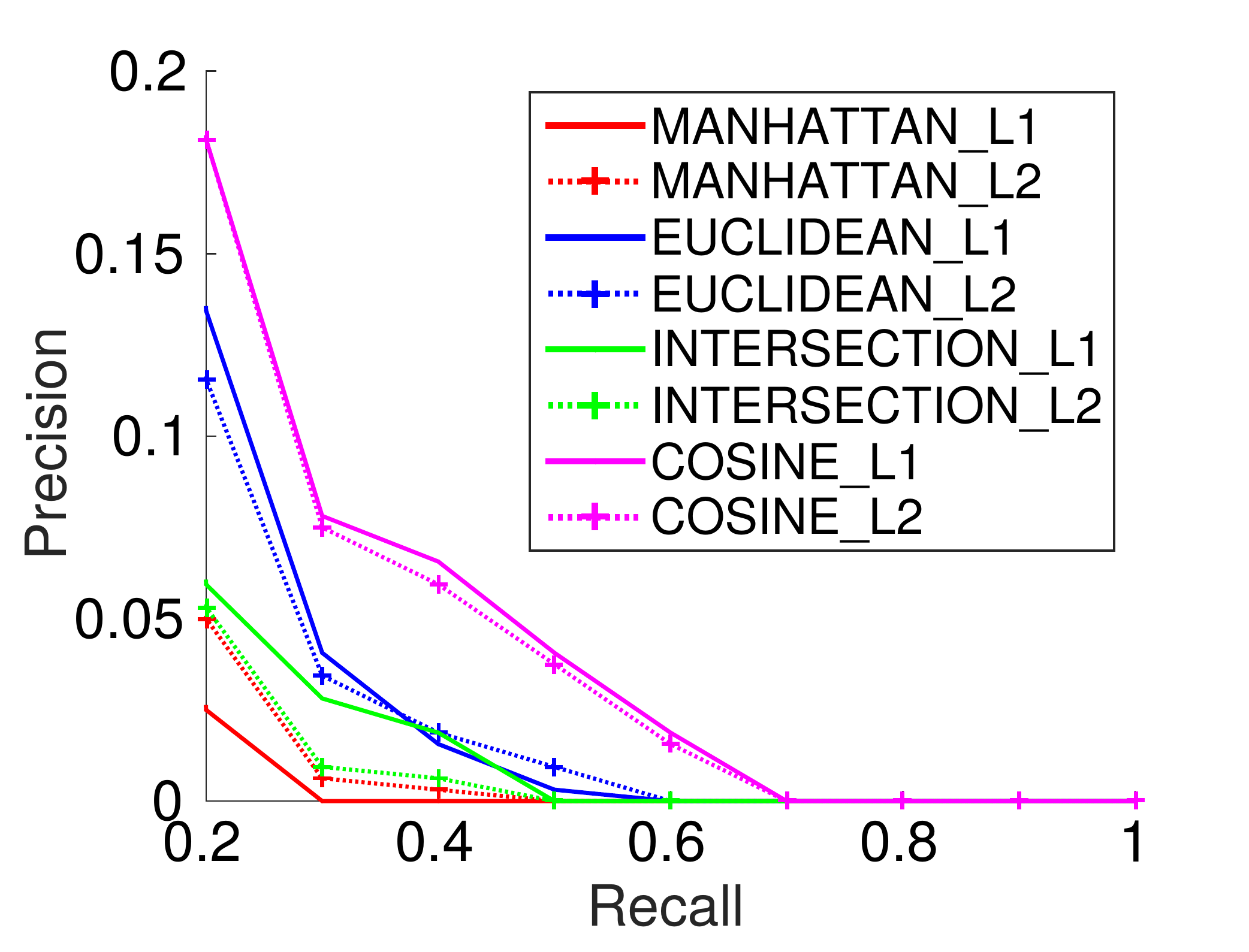}
		\label{fig:lbp_ri}} \\
	\subfloat[The PR graph of $LBP_{P,r}^{u2}$]{
		\includegraphics[width = 0.45\linewidth]{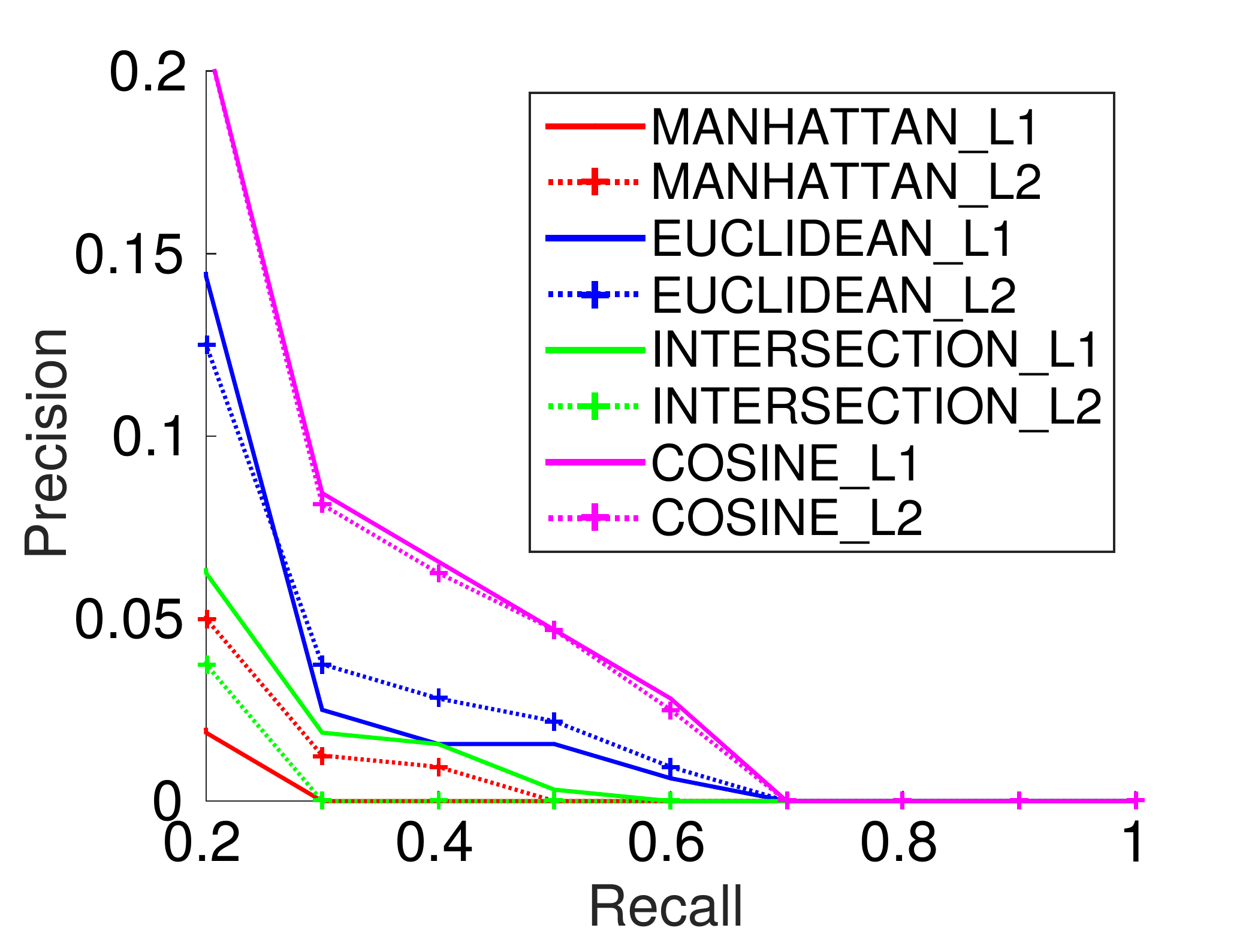}
		\label{fig:lbp_u2}} 
	\subfloat[The PR graph of $LBP_{P,r}^{riu2}$]{
		\includegraphics[width = 0.45\linewidth]{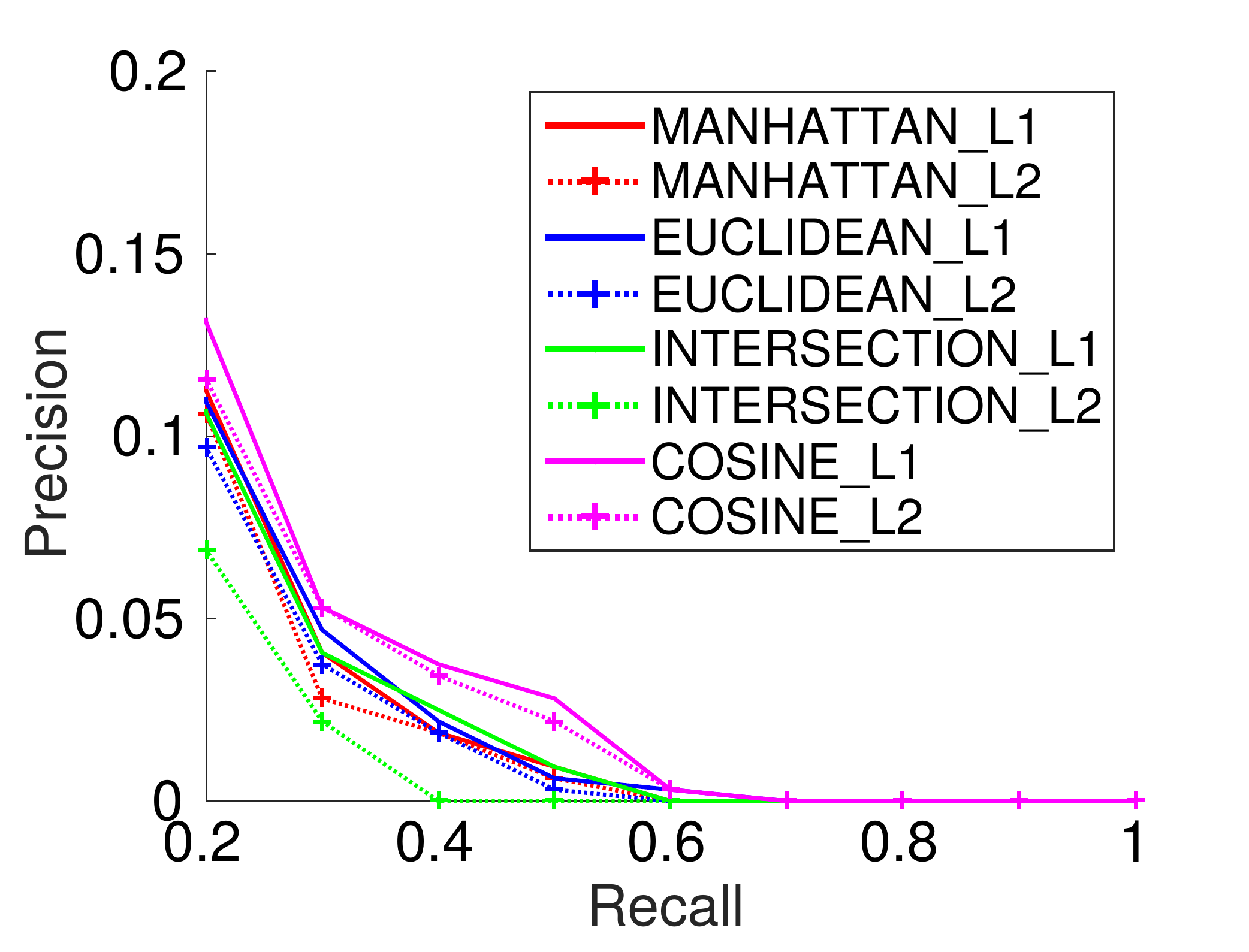}
		\label{fig:lbp_riu2}} \\
	\subfloat[The comparison of outstanding schemes from (a-d)]{
		\includegraphics[width = 0.45\linewidth]{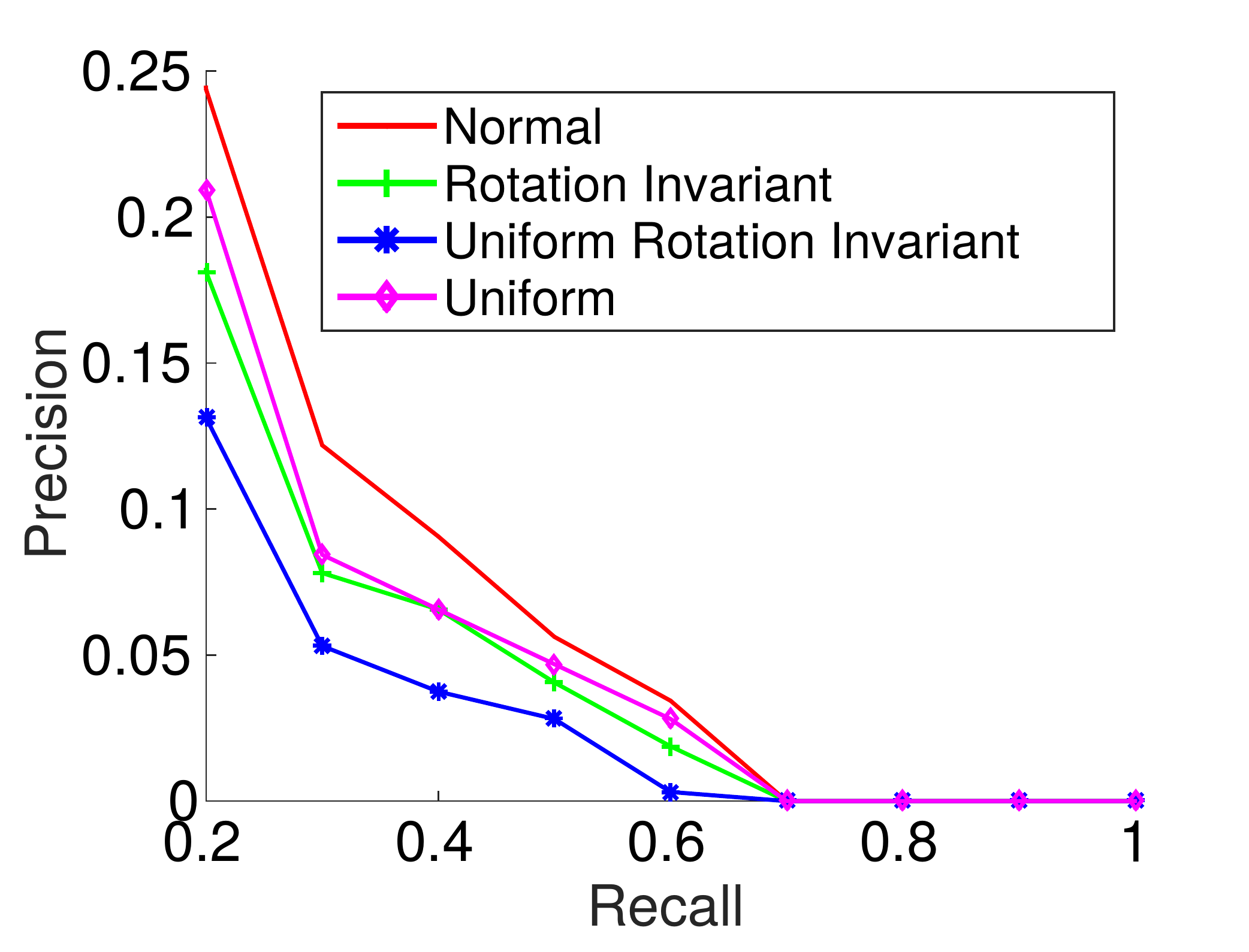}
		\label{fig:lbp_best}}
	\caption{Performance of different LBP variants on the METU dataset. (a-d) The results of $LBP_{P,r}$, $LBP_{P,r}^{ri}$, $LBP_{P,r}^{u2}$, $LBP_{P,r}^{riu2}$. (e) A comparison of the best overall results. In legends of (a-d), the string suffixes indicate the utilized distance metric and normalization type.
		\label{fig:lbp_rst}}
\end{figure}

The rudiment LBP method could achieve rotation invariance and robust discrimination ability with some modifications, such as bit-wise shifting and `uniform' operations  \cite{ojala2002multiresolution}. Similar to the color histogram method, the performance of the LBP method is dependent on the selected distance metric and normalization method. Figure \ref{fig:lbp_rst} displays the effect of the different settings, which shows the best LBP configuration to be  the original LBP method with the cosine distance metric and L1 normalization. Therefore, we will adopt these settings for LBP in the rest of the article.

\subsubsection{A Global Feature: GIST}

The GIST descriptor is initially designed for scene recognition  \cite{oliva2001modeling}. It describes objects with spatial envelope properties (a very low dimensional representation of the scene): the degree of naturalness, openness, roughness, expansion and ruggedness. These properties are computed by using the principal components of the global energy spectrum and the spectrogram. Since the descriptor uses only the mentioned spatial envelope properties, it projects images into a low dimensional feature space. This makes GIST a very compact and efficient descriptor for a global representation of an image.

Douze \etal \cite{douze2009evaluation} used GIST for large scale copyright detection. They found that GIST outperforms the most commonly used model, i.e., BoVW with local descriptors like SIFT, when searching duplicate images from a very large scale image dataset. We expect that GIST descriptor can be useful in trademark retrieval as well since it is known to be good at capturing the layout of a figure. 

\subsubsection{Bag of Visual Words (BoVW)}

The scaling problem is the bottle neck of large scale trademark retrieval, especially when methods extract multiple high-dimensional features from images as methods introduced in the following part. Storing and comparing tremendous key-point features extracted from large scale dataset is very challenging. Therefore, the method of bag of visual words (BoVW) \cite{Sivic2003} is adapted. In this approach, each feature is expressed with their unique cluster id, which is obtained by clustering all features to $k$ different classes. This BoVW model not only shrinks the feature spaces, but also grants the computational efficiency through applying the TF-IDF (term frequency-inverse document frequency) \cite{chowdhury2010introduction} and inverted file structures \cite{Sivic2003}. Through applying this model, high dimensional features space are mapped into vectors whose similarity is calculated with cosine vector distance metric in this study.

%
%

\subsubsection{Shape Context}
The main content of images, shapes, makes substantial impression on customers \cite{her2011hybrid}. It is, therefore, one of the most significant aspects considered for judging similarity. 

A robust shape feature is critical to trademark retrieval. Yang \etal \cite{Mingqiang2008} suggest that a robust shape feature should include most of the following properties: identifiability, translation, rotation, scale, affine and occlusion invariance, noise resistance, statistical independence, reliability. The \textit{shape context} method proposed by Belongie \etal \cite{belongie2000shape} is known to be a suitable shape descriptor, satisfying most of the properties aforementioned. The shape context of a shape is spatial distributions of all sample points to each sample point from it. The deformation energy necessary for matching shape contexts is the similarity degree of shapes.

\begin{figure}[hbt]
	\centering
	\subfloat[Logo of NIKE] {
		\includegraphics[width = 0.3\textwidth]{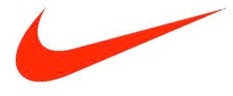}
		\label{fig:nike_logo_polar1}
	}
	\hspace{2cm}
	\subfloat[a polar histogram of a point on NIKE logo] {
		\includegraphics[width = 0.3\textwidth]{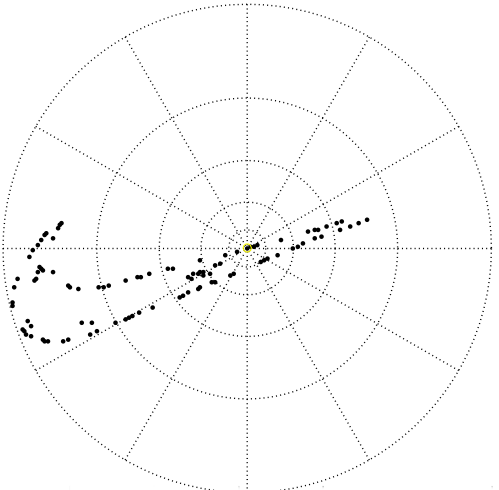}
		\label{fig:nike_logo_ploar2}
	}
	\caption{A polar histogram of a sample logo (Adapted from \cite{tuerxun2015comparison}).}
	\label{fig:shape-context_polar}
\end{figure}
\begin{figure}[htb]
	\centering
	\subfloat[NIKE] {
		\includegraphics[width = 0.25\textwidth]{image/shape-context/nike.jpg}
		\label{fig:nike_logo}
	}
	\hspace{2cm}
	\subfloat[NEWPORT] {
		\includegraphics[width = 0.25\textwidth]{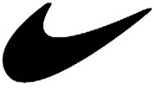}
		\label{fig:np_logo}
	}\\
	\subfloat[Shape-context of NIKE logo] {
		\includegraphics[width = 0.45\textwidth]{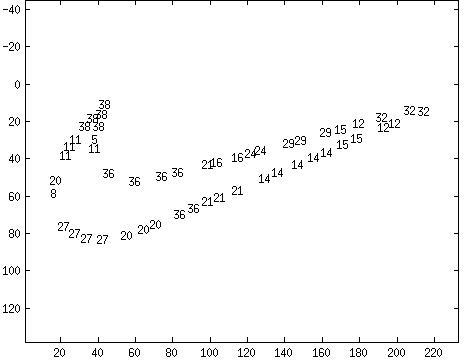}
		\label{fig:sc_nike}
	}
	\subfloat[Shape-context of NEWPORT logo] {
		\includegraphics[width = 0.45\textwidth]{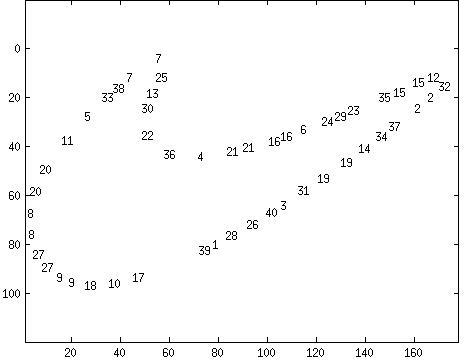}
		\label{fig:sc_np}
	}
	\caption{Shapeme of sample logos (Adapted from \cite{tuerxun2015comparison}).}
	\label{fig:shape-context_logo}
\end{figure}

The shape context of a shape is generated through the following steps: (1) Uniformly sample $n$ points from inner and outer outline of the shape. (2) Assign a log-polar histogram to each sample point. A sample log histogram, in which radius bins $\theta$ is 5 and angle bins $logr$ is 12, is shown in Figure \ref{fig:shape-context_polar}. (3) According to the allocation of sample points on each log-histogram, generate $n$ vectors of shape context.
%
%
Computing the deformation energy of shape contexts of million shapes is costly. Approximate solutions have been developed for this purpose. For example, Mori \etal \cite{mori2001shape} proposed two different approximation approaches: representative shape-context and shapeme histogram descriptor. The shapeme histogram method is similar to the BoVW method. It applies vector quantization to all descriptors as shown in Figure \ref{fig:shape-context_logo}. With this approach, the shape context becomes more efficient in terms of time and memory aspects. What is more, Rusino \etal \cite{rusinol2010efficient} achieved further scalability through organizing shapeme histogram descriptors by a local-sensitive hashing indexing structure for searching similar descriptors in a sub-linear time. In this article, we will employ the shape-context descriptor with the BoW model with a dictionary size of 10,000 (this is decided empirically).

%
  
\subsubsection{Keypoint-based Features}

If two trademarks are similar, they should be composed of similar key-points. To extract key-point descriptors from an image, the first step is detection: One of the most popular methods for this purpose, SIFT, takes as key-points the intensity changes overlapping in multiple scales in a multi-scale filtered representation of an image. While, for speeding up the detection process, SURF applies a Hessian-matrix-based blob detector to find key-points.

After detection, the second step is the description of the visual content at and around keypoints. Key-point descriptor methods generate a description of a key-point usually from the distribution of gradients and orientation of its nearby pixels.

In this study, we evaluate the most popular key-point descriptors: SIFT \cite{lowe2004distinctive}, SURF \cite{bay2008speeded}, and HOG \cite{dalal2005histograms}. In several studies \cite{kochakornjarupong2011trademark, wei2009trademark, lin2008trademark}, these features have been already applied for trademark retrieval.

\textbf{Triangular SIFT}:
Despite the fact that SIFT is an effective, stable and robust descriptor, it is not recommended for large scale datasets because of its computational complexity. To scale up SIFT, the local geometry information is usually incorporated. 

One such promising attempt, owing to Kalantidis \etal \cite{kalantidis2011scalable}, showed that grouping SIFT features at the same scale as triplets and comparing only triplets of SIFTs at the same scale at the matching phase both improves the accuracy and running-time.

\subsection{Learned Features Using Convolutional Neural Networks (CNN)}

To the best of our knowledge, this is the first study using deep learning for trademark retrieval. Deep learning (networks) relies on finding an end-to-end mapping directly from the raw input to the required output, whereby the best representation for the problem at hand is obtained from the data directly, leading to distributed, compositional, hierarchical representations.

One of the prominent methods in deep learning is Convolutional Neural Networks (CNNs), which exploit local connectivity and weight sharing mechanisms (see, e.g., \cite{lecun1989generalization}). CNNs mainly learn filters for convolution operation at different layers and scales, together with complementary operations like non-linear transformation, pooling (down-sampling) etc. These filters are trained using back propagation for various problems such as classification, detection and recognition. 

In this work, we evaluated the widely used pre-trained networks, namely AlexNet \cite{NIPS2012_4824}, VGGNet \cite{Simonyan14c}, and GoogLeNet \cite{GoogLeNet} -- see Table \ref{tab:dlcm} for a comparison of the architectures. We extracted features from trademarks through these models, then compared these features with cosine vector distance. 

We have also trained two different comparatively shallow denoising autoencoders \cite{vin2008}. These two autoencoders use $3\times3$ convolutional kernels, following the work of \cite{Simonyan14c}. The encoder structure of the autoencoders, $ae^1$ and $ae^2$, are [16 ($3\times3$), 8 ($3\times3$), 8 ($3\times3$)] and [128 ($3\times3$), 64 ($3\times3$), 64 ($3\times3$)] respectively -- see also Table \ref{tab:dlcm}.

\begin{table}[tbh]
	\caption{A comparison of the deep networks. For the number of layers, only weighted layers are counted. In the architecture descriptions, I represents input; C, convolution layer; P, pooling layer; D, dropout layer; F, fully connected layer; and N, inception network described in \cite{GoogLeNet}. \label{tab:dlcm}} 
	\vspace*{-0.2cm}
	\begin{center}
		\resizebox{\textwidth}{!}{
			\begin{tabular}{l l l l l}
				\hline
				\multirow {2}{*}{\bf Network} 	& \bf \# of & \bf \# of & \bf Feature & \bf Overall\\
				& \bf layers	                  &	\bf parameters	                & \bf dimension	       	& \bf architecture	         \\ 		
				\hline
				\hline
				
				AlexNet \cite{NIPS2012_4824}  & 8  & 61M  		& 4,096       & $I-[CP]^2-C^2-[CP]-F^3$  \\ \hline 
				VGGNet16 \cite{Simonyan14c}   & 16 & 138M 		& 4,096 (FC7) & $I-[CCP]^2-[CCCP]^3-F^3$ \\ \hline
				VGGNet16 \cite{Simonyan14c}   & 16 & 138M 		& 1,000 (FC8) & $I-[CCP]^2-[CCCP]^3-F^3$ \\ \hline
				GoogLeNet \cite{GoogLeNet}    & 22 & 6.9M 		& 1,024       & $I-[CP]-[CCP]-N^9-P-F$    \\ \hline
				Autoencoder ($ae^1$)		  & 8  & 4,963		& 288	  	 & $I-[CP]^3-[CU]^3-C$		\\ \hline
				Autoencoder ($ae^2$)		  & 8  & 200,899	& 8,192	   	 & $I-[CPD]^3-[CU]^3-C$		\\ \hline					  					  
			\end{tabular}}
		\end{center}
	\end{table}

\subsection{Summary}

Overall, we have selected a wide range of features representing different aspects of content in trademarks, both hand-designed and learned from data directly. These features have different advantages and disadvantages, as shown in Table \ref{tab:metcom}, which indicates their fusion might perform better than the individual methods.

\begin{table}[tbh]
	\caption{Comparison of the feature extraction methods.\textit{Robustness} means robustness to translation, scaling, rotation, and occlusion, and \textit{efficiency} pertains to time and memory efficiency. \label{tab:metcom}}
	\vspace*{-0.2cm}
	\begin{center}
		\resizebox{\textwidth}{!}{
		\begin{tabular}{l l l l l l l l l}
			\hline
			\multirow {2}{*}{\bf Algorithm} 	& \multirow {2}{*}{\bf Shape} & \multirow {2}{*}{\bf Color} & \multirow {2}{*}{\bf Texture} & \multirow {2}{*}{\bf Layout}  & \bf Partial 	& \multirow {2}{*}{\bf Efficiency} & \multirow {2}{*}{\bf Robustness} & \multirow {2}{*}{\bf Type}\\
							&		&		&		  &			& \bf matching	&			 &			  &\\ 		
			\hline 			\hline
			Color 			&-		&$*****$	&-	  	  & -	& -  		&$****$		 & $*$ 	& Global	\\ \hline 
			LBP				&-		&-		&$****$	  &	-	  	& -  		&$****$		 & $***$	& Global	\\ \hline
			GIST			&$***$	&$***$		&$***$	      &	$****$	& -		  	&$****$		 & $**$		& Global 	\\ \hline
			SHAPEMES		&$****$	& -		& -	      &	$***$	& $***$		&$***$ 		 & $***$	& Local 	\\ \hline
			HOG				&$**$	& -		& $**$	  &	-		& $***$		&$**$ 		 & $***$	& Local		\\ \hline		
			SIFT			&$**$	& -		& $**$	  &	-		& $****$	&$*$ 		 & $***$	& Local	 	\\ \hline				
			SURF			&$**$	& -		& $**$	  &	-		& $****$	&$**$ 		 & $***$	& Local	 	\\ \hline
			DCNN			&$****$	& $***$		& $***$	  &	$***$		& $***$			&$**$ 		 & $****$	& Local	\\ \hline
		\end{tabular}}
	\end{center}
	
\end{table}

\section{Enhancing and Fusing Features}
\label{sect:enhancing_features}

We noticed that the overall performance of some features could be improved by (i) leveraging contrast change and removing text, which is irrelevant for trademarks not including text, and (ii) fusing the features, combining their advantages.

\subsection{Detecting and Removing Text in Trademarks}
\begin{figure}[htb]
	\centering
	\subfloat[]{
		\includegraphics[width = 0.45\textwidth]{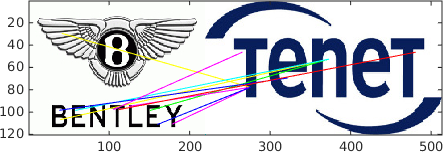}
		\label{fig:text_sift_1}}
	\qquad
	\subfloat[]{
		\includegraphics[width = 0.45\textwidth]{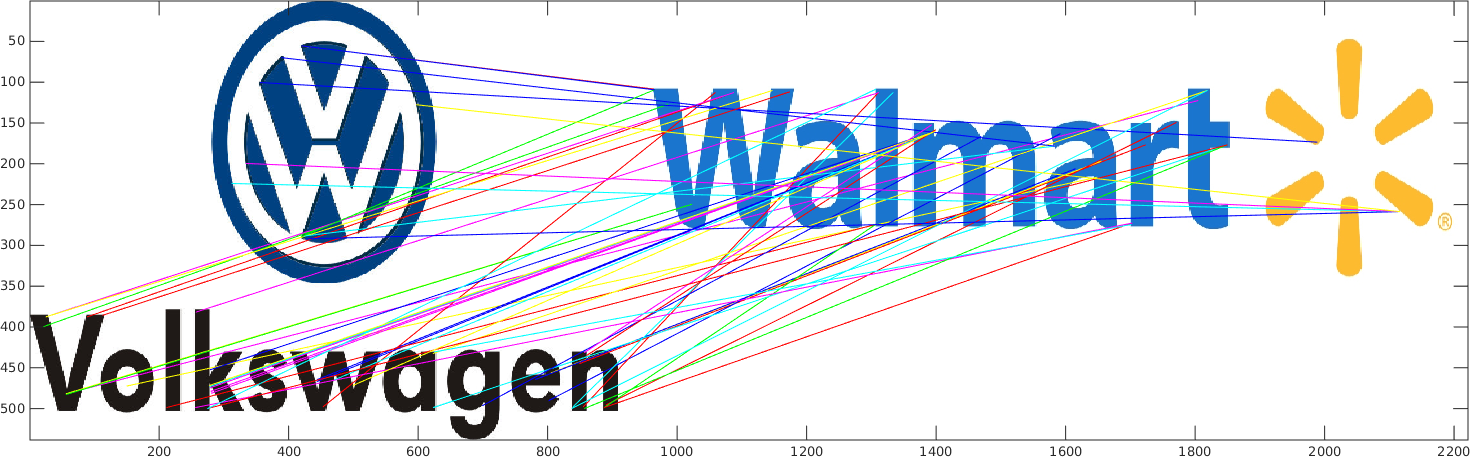}
		\label{fig:text_sift_2}}
	\caption{Example of the influence of characters on key-point based detection (shown for the SIFT features).}
	\label{fig:text_sift}
\end{figure}

%
Text is a misleading element for retrieval if the query logo does not include any text. Text in trademarks leads to many keypoints and features, which significantly affect the overall matching performance -- see Figure \ref{fig:text_sift}. If the query logo includes text, a good strategy is to recognize the text and evaluate similarity based on the recognized text.

\begin{figure}[htb]
	\centering
	\subfloat[]{
		\includegraphics[width = 0.23\textwidth]{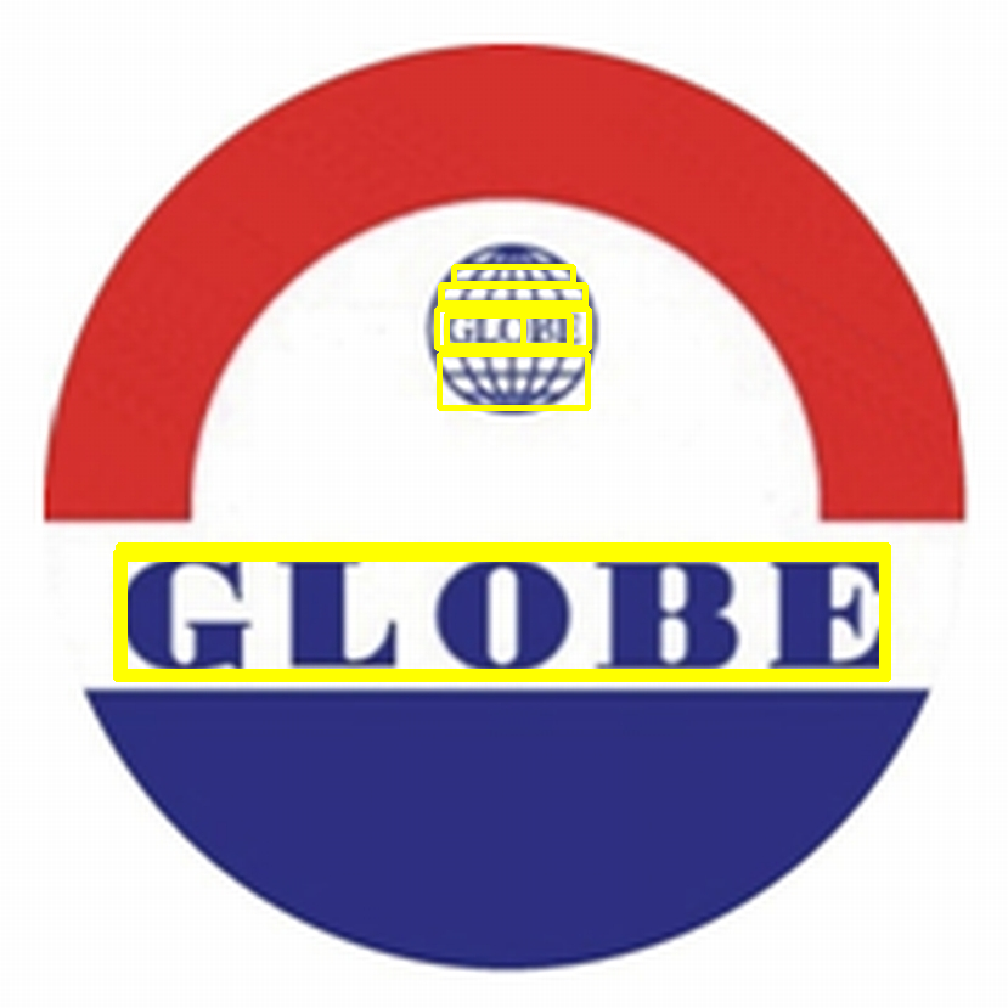}
		\label{fig:text_detect_1}}
	\subfloat[]{
		\includegraphics[width = 0.23\textwidth]{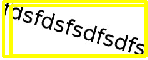}
		\label{fig:text_detect_2}}
	\subfloat[]{
		\includegraphics[width = 0.23\textwidth]{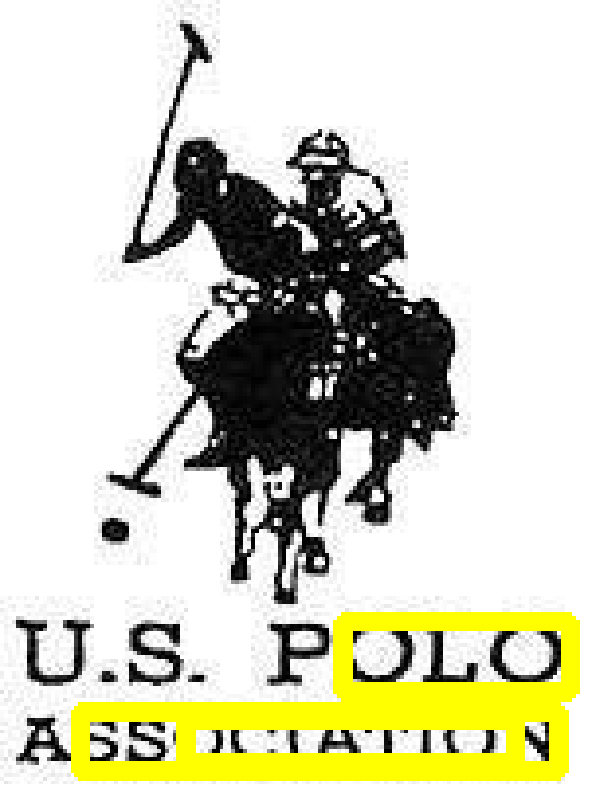}
		\label{fig:text_detect_3}}
	\subfloat[]{
		\includegraphics[width = 0.23\textwidth]{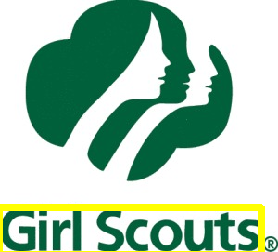}
		\label{fig:text_detect_4}}
	\caption{Text detection results on sample trademarks (detection shown in yellow) using the method by Neumann \etal \cite{neumann2012real}.}
	\label{fig:text_detect}
\end{figure}

For locating text in trademarks, we use a state-of-the-art method proposed by Neumann \etal~\cite{neumann2012real}, which performs real-time text localization by detecting characters by using the Extremal Region (ER) detector, which is robust and stable to illumination, blur, and color and texture variation -- see Figure \ref{fig:text_detect} for some results.
%

%
\subsection{Contrast Enhancement}
\begin{figure}[htb]
	\centering
	\subfloat[SIFT]{
		\includegraphics[width = 0.4\textwidth]{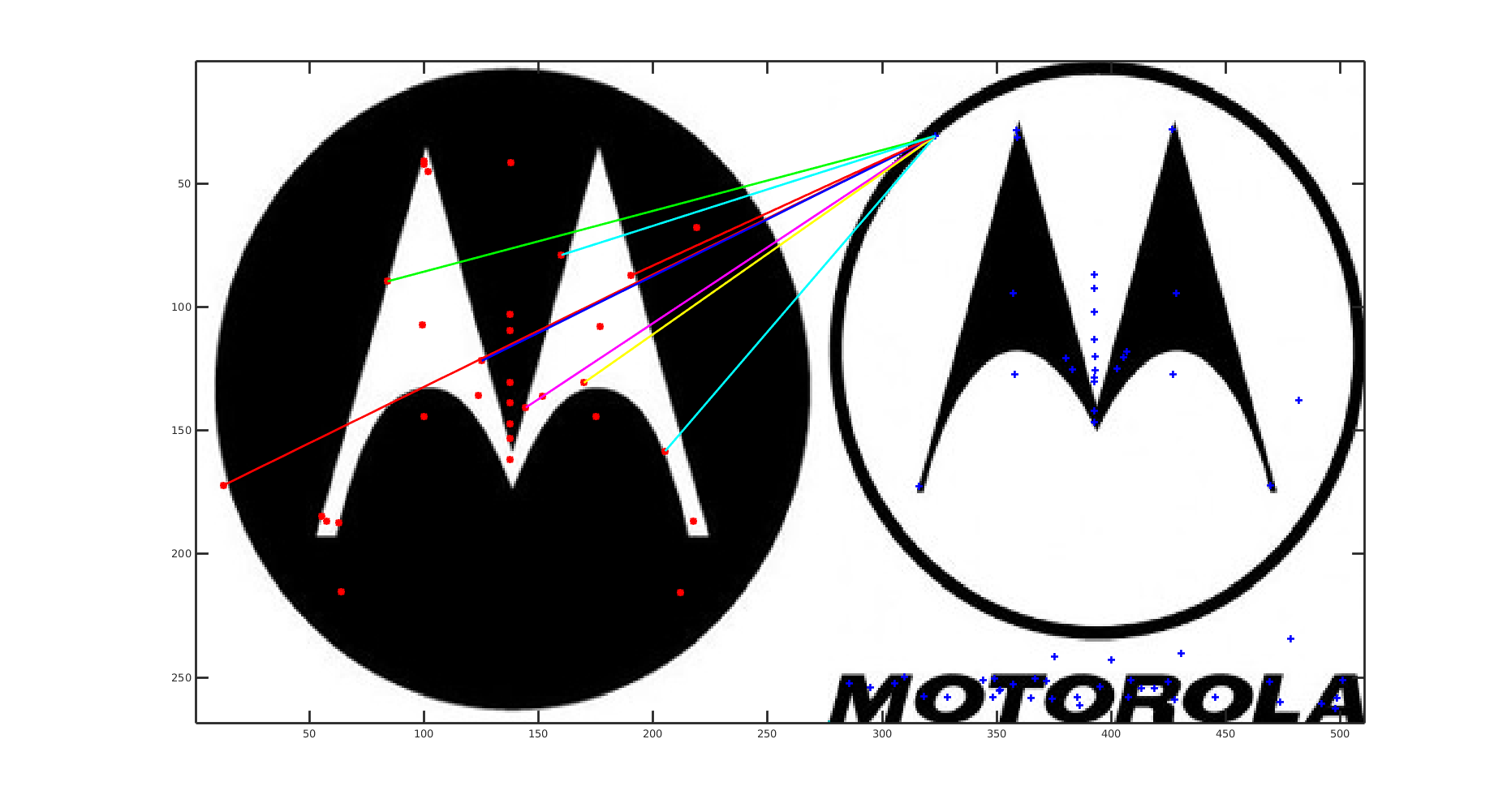}
		\label{fig:con_sift_1}}
	\hspace{1cm}
	\subfloat[OR-SIFT]{
		\includegraphics[width = 0.4\textwidth]{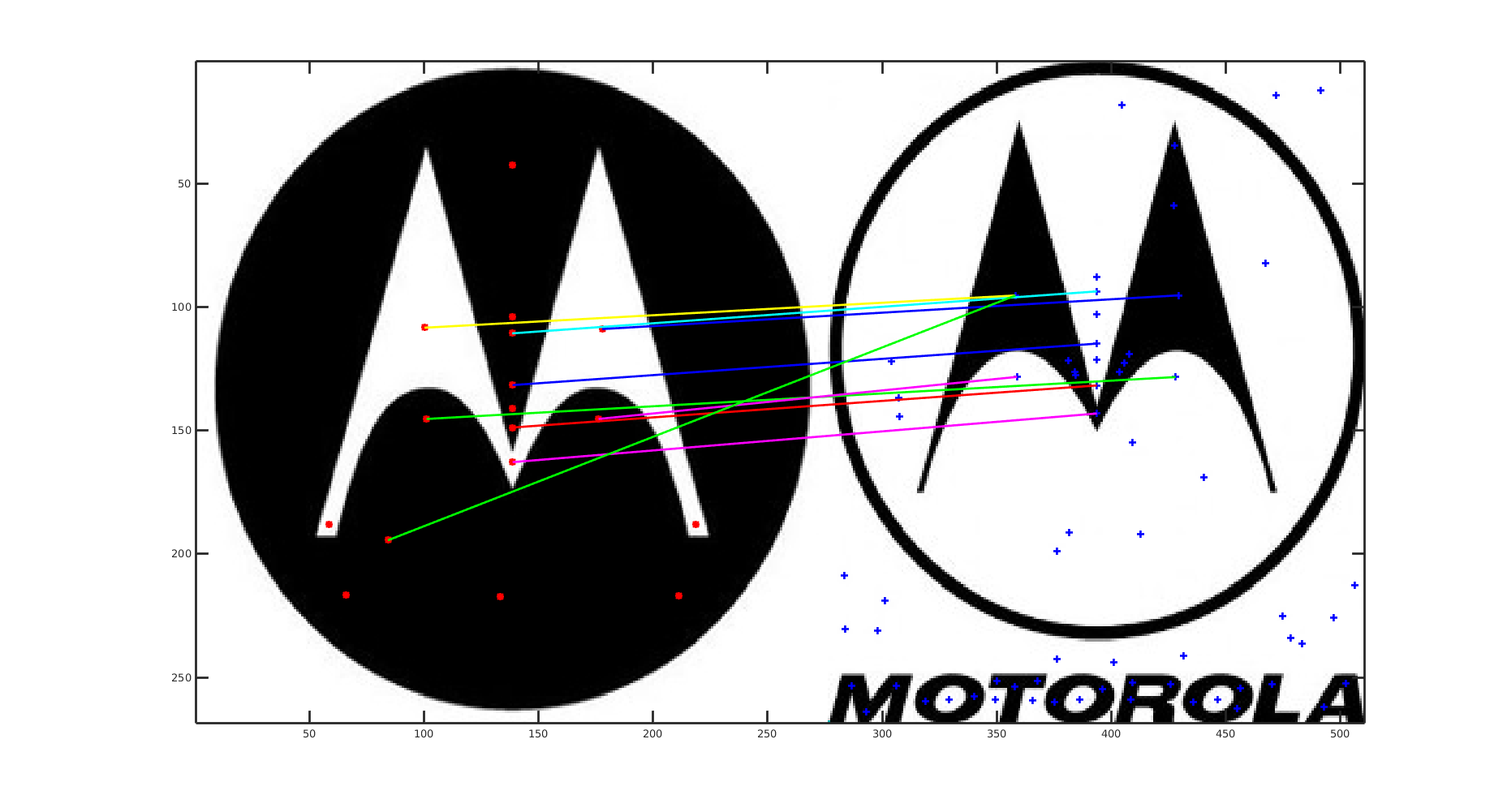}
		\label{fig:con_orsift_1}}
	\\
	\subfloat[SIFT]{
		\includegraphics[width = 0.4\textwidth]{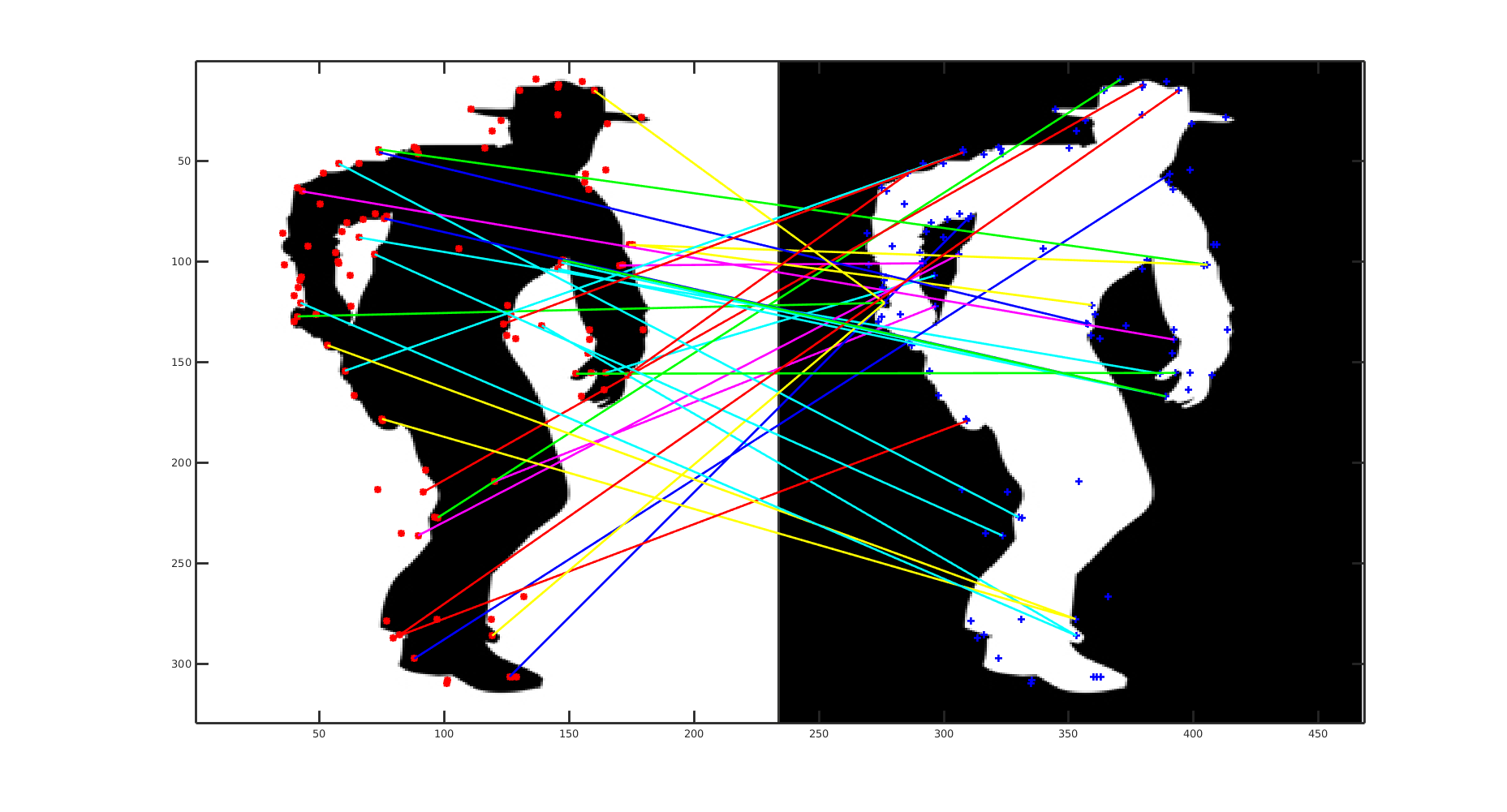}
		\label{fig:con_sift_2}}
	\hspace{1cm}
	\subfloat[OR-SIFT]{
		\includegraphics[width = 0.4\textwidth]{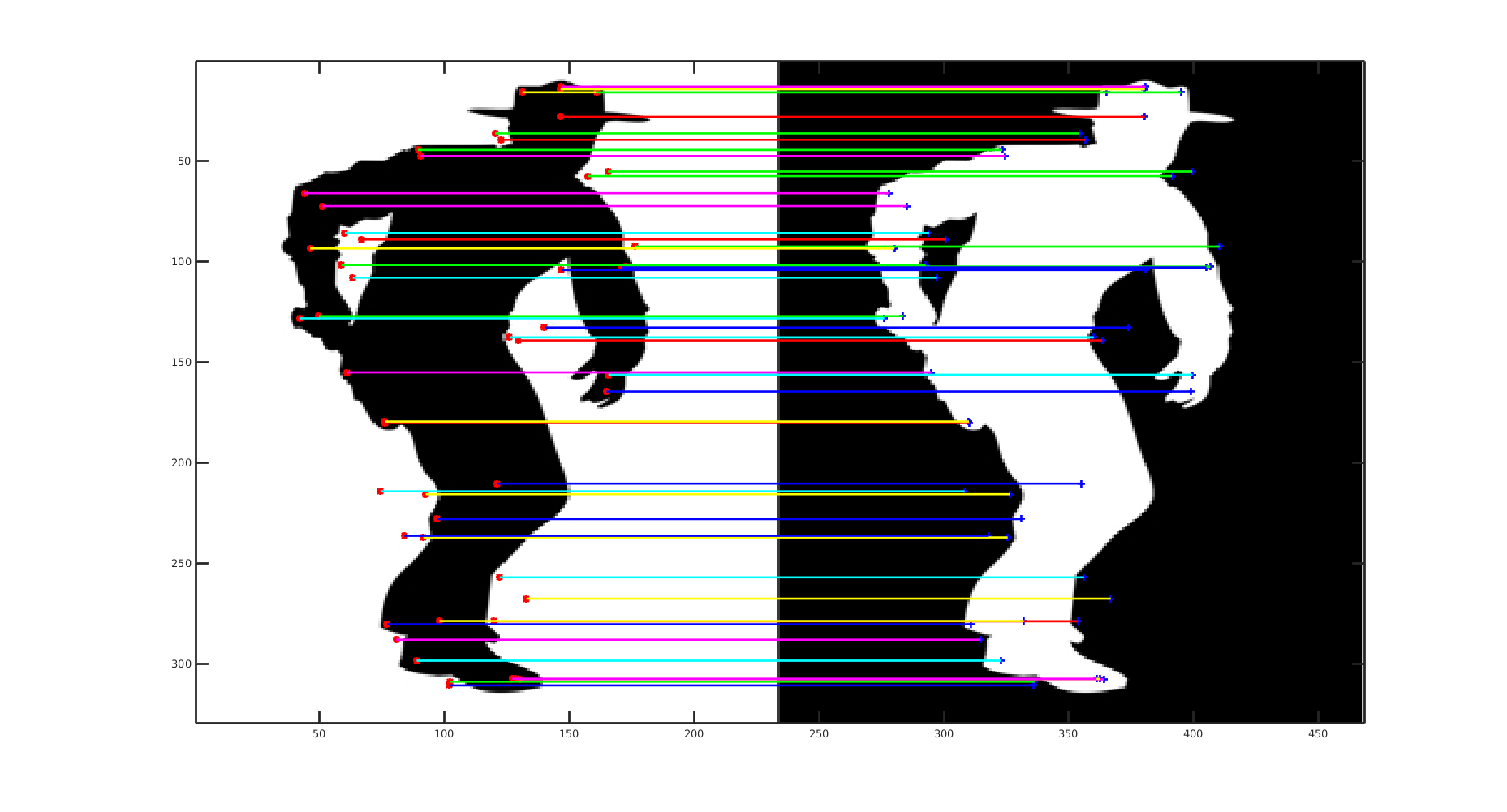}
		\label{fig:con_orsift_2}}
	\caption{Matching images with different contrast changes using SIFT and OR-SIFT\cite{vural2009registration}. In (a-b), similarity between the trademarks is missed using naive SIFT. In (c-d), a modification of SIFT, OR-SIFT, captures the similarity despite contrast change. Lines are colored randomly only for the sake of visibility.}
	\label{fig:con_pro}
\end{figure}

Key-point feature descriptors like SIFT are sensitive to contrast change, which causes TR systems to be ignorant to infringements that include a different contrast change -- see Figure \ref{fig:con_pro} for examples. Extensions of SIFT, namely Orientation-Restricted SIFT (OR-sift) and GOM-SIFT presented in \cite{vural2009registration, yi2008multi}, are made robust to this contrast issue. GOM-SIFT achieves this by restricting orientation values of each feature between $0^{\circ}$ and $180^{\circ}$ for increasing the performance against contrast cases. GOM-SIFT leads to improvement though sacrificing rotation invariance. To keep contrast robustness with rotation invariance, Vural \etal~\cite{vural2009registration} proposed OR-SIFT, which merges directions who are $180^{\circ}$ apart. For this reason, we employ OR-SIFT in this article. See Figure \ref{fig:con_pro} for results of OR-SIFT key-point comparisons on trademarks having contrast differences. 
%

\subsection{Fusion of Features}
In trademark retrieval, fusion of features has been applied successfully already \cite{rusinol2011interactive, ravela2005multi, Wei2009, zhang2012technique, eakins2003shape}. In this study, we have also fused the best performing methods. The fusion method we have applied is Inverse Rank Position (IRP) \cite{jovic2006image}. It takes inverse of the sum of inverse of similarity ranks.
\begin{equation}
IRP(q,i) = {1}/{\sum_j^n{\frac{1}{rank_j}}},
\end{equation}
where $j$ represents the $j^{th}$ feature, $q$ is query image, $i$ is $i^{th}$ image.


%% file: section/experiments_and_results.tex
\section{Experiments and Results}

\label{sect:experiments}

In this section, we first introduce the experimental setup, the evaluation method, and then the results with an analysis.

\subsection{Experimental Setup}
Most of the experiments are conducted on a PC with an Intel i7-4770K 3.50GHz CPU with 32GB DDR3 memory, and a GeForce GTX 760 graphics card. However, we used the Tesla K40 GPU card for developing autoencoder models.  

\begin{figure}[tbh]
	\centering
	\includegraphics[width = 1\textwidth]{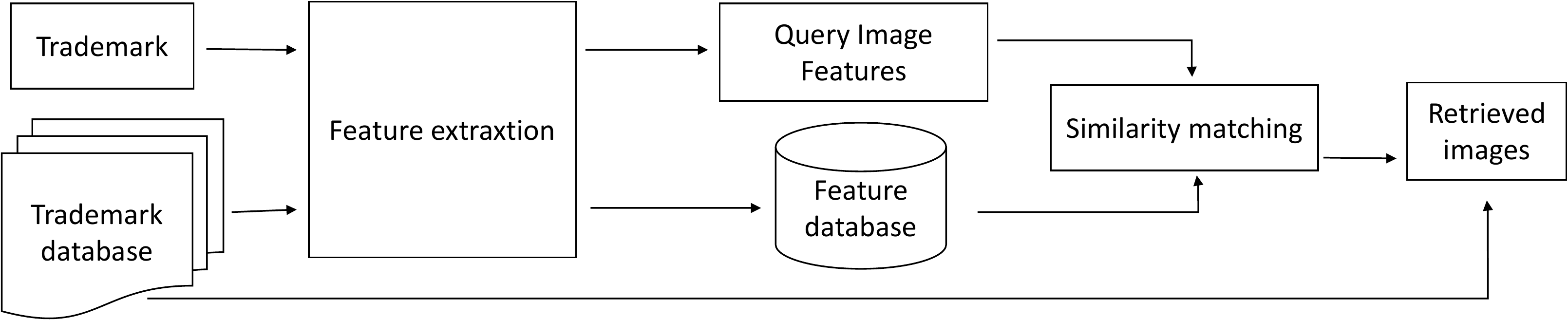}
	\caption{The overall view of how the experiments are performed.}
	\label{fig:cbir}	
\end{figure}

Our main experiment flow is visualized in Figure \ref{fig:cbir}.

\subsection{Evaluation Method and Metrics}

As discussed in Section \ref{sect:dataset}, the dataset includes 45 query groups, and in each group there are around 10-15 logos for which similarities have been identified by an expert. For each image in the query set, we ``inject" the other images in the query group into the test-set, apply the trademark retrieval method and look at the rank of the logos that have been injected.

For evaluating the retrieval performance, we use precision-recall (PR) graphs, and average ranks of the ``injected'' known trademarks as performed in the CBIR literature. Precision and recall are defined as follows:
\begin{equation}
Precision = \frac{\emph{No. of relevant retrieved Trademarks }} {\emph{No. of retrieved Trademarks}},
\end{equation}
\begin{equation}
Recall = \frac{\emph{No. of relevant retrieved Trademarks}}{\emph{No.of relevant Trademarks}}.
\end{equation}
A PR graph offers an intuitive comparison of the retrieval ability of a set of methods for various levels of sensitivity.
Besides PR, we use average rank and normalized rank for evaluating ranking ability of the methods. The normal rank metric returns actual average ranks of relevant logos (following the notation and the definition by Sivic \& Zisserman \cite{Sivic2003}): 
\begin{equation} \label{eq2}
Rank = \frac{1}{N_{rel}}\sum\limits_{i=1}^{N_{rel}}{R_i},
\end{equation}
where $N_{rel}$ is the number of relevant images for a particular query image, $N$ is the size of the image set, and $R_i$ is the rank of the $i^{th}$ relevant ``injected" image. In contrast, the normalized rank metric returns a score for evaluating the robustness of the retrieval method:
\begin{equation} \label{eq1}
\widetilde{Rank} = \frac{1}{N\times N_{rel}}\left(\sum\limits_{i=1}^{N_{rel}}{R_i-\frac{N_{rel}(N_{rel}+1)}{2}}\right).
\end{equation}
Average rank ranges from $1 + \frac{N_{rel}}{2}$ to $N - \frac{N_{rel}}{2}$ s.t. the smaller the rank is, the better the performance is. In contrast, the normalized rank measure lies in the range $[0, 1]$. Zero ($0$) corresponds to the best performance, and $0.5$ to random performance. These two ranking scores capture a global view of retrieval ability of the methods. However, in our experiments, methods exhibit different retrieval performances to different queries. In order to capture this, we also visualize the ranking results of all queries as a graph.

Last but not the least, in the ranking process, we realized that \textit{tie cases} may occur due to same similarity scores when descriptors failed to extract sufficient information from the trademarks. For resolving the tie cases, the average of original ranks  is used in the following ranking results (similar to \cite{saha2007image, jin2005improving}). 

\subsection{Results}

In this section, we analyze the methods in terms of performance and efficiency. Sample queries are provided at the following page:  \url{http://kovan.ceng.metu.edu.tr/~osman/dataset_webpage/query.html}.

\subsubsection{Precision-Recall and Average Rank Results}

\begin{table}[!htb]
	\caption{Comparison of the results of the traditional individual methods.}
	\label{tab:rank_mean}
	\vspace*{-0.2cm}
	\begin{center}
		\resizebox{\textwidth}{!}{
		\begin{tabular}{l c c c c}
			\hline
			\multirow {2}{*}{\bf Algorithm (id)}  & \bf BoW & \bf Without & \bf Average	 & \textbf{Normalized }\\
				 			&\bf cluster			& \bf{text?}		 &	\bf rank	 & \bf{average rank}\\ \hline \hline
			Color ($cl$)			&-		&	& 369,598.3 $\pm$ 161,895.1			& 0.400 $\pm$ 0.175\\ 
			LBP ($lp$)			&-	 	&	& 254,971.8 $\pm$ 131,399.5			& 0.276 $\pm$ 0.142\\		
			GIST ($gs$)				&-		&	& 234,087.1  $\pm$ 159,585.2		& 0.254 $\pm$ 0.173\\ 
			SHAPEMES ($sh$) 			&10k	 &	& 203,408.2 $\pm$ 171,317.4			& 0.220 $\pm$ 0.186\\ 
			HOG	($hg$)				&10k	 &	& 242,166.1 $\pm$ 118,686.6			& 0.262 $\pm$ 0.129\\
			\hline
			SIFT ($si^1$) 				&10k	 &	& 164,837.7 $\pm$ 133,932.5     	& 0.179 $\pm$ 0.145\\
			SIFT ($si^2$)				&999	 &	& 192,881.1 $\pm$ 144,359.4			& 0.209 $\pm$ 0.156\\
			SIFT ($si^3$)				&9		 &	& 321,268.8  $\pm$ 132,487.4		& 0.348 $\pm$ 0.143 \\
			TRI-SIFT ($ts$) 		 &9	 	 &	& 298,744.3 $\pm$ 148,279.1			& 0.324 $\pm$ 0.161\\
			OR-SIFT ($os$)			&10k	 &	& 175,482.6 $\pm$ 139,185.6			& 0.190 $\pm$ 0.151\\
			SIFT ($si^4$)			&10k	 & \checkmark	& \bf 141,840.9 $\pm$ 117,705.3	& \bf 0.154 $\pm$ 0.127 \\
			\hline
			SURF ($su$)				&10k	 &	& 191,304.1 $\pm$ 139,696.4		   & 0.207 $\pm$ 0.151 \\
			\hline
		\end{tabular}
	}
	\end{center}
\end{table}

\begin{figure}[!htb]
	\centering
	\subfloat[Original]{
		\includegraphics[width = 0.45\textwidth]{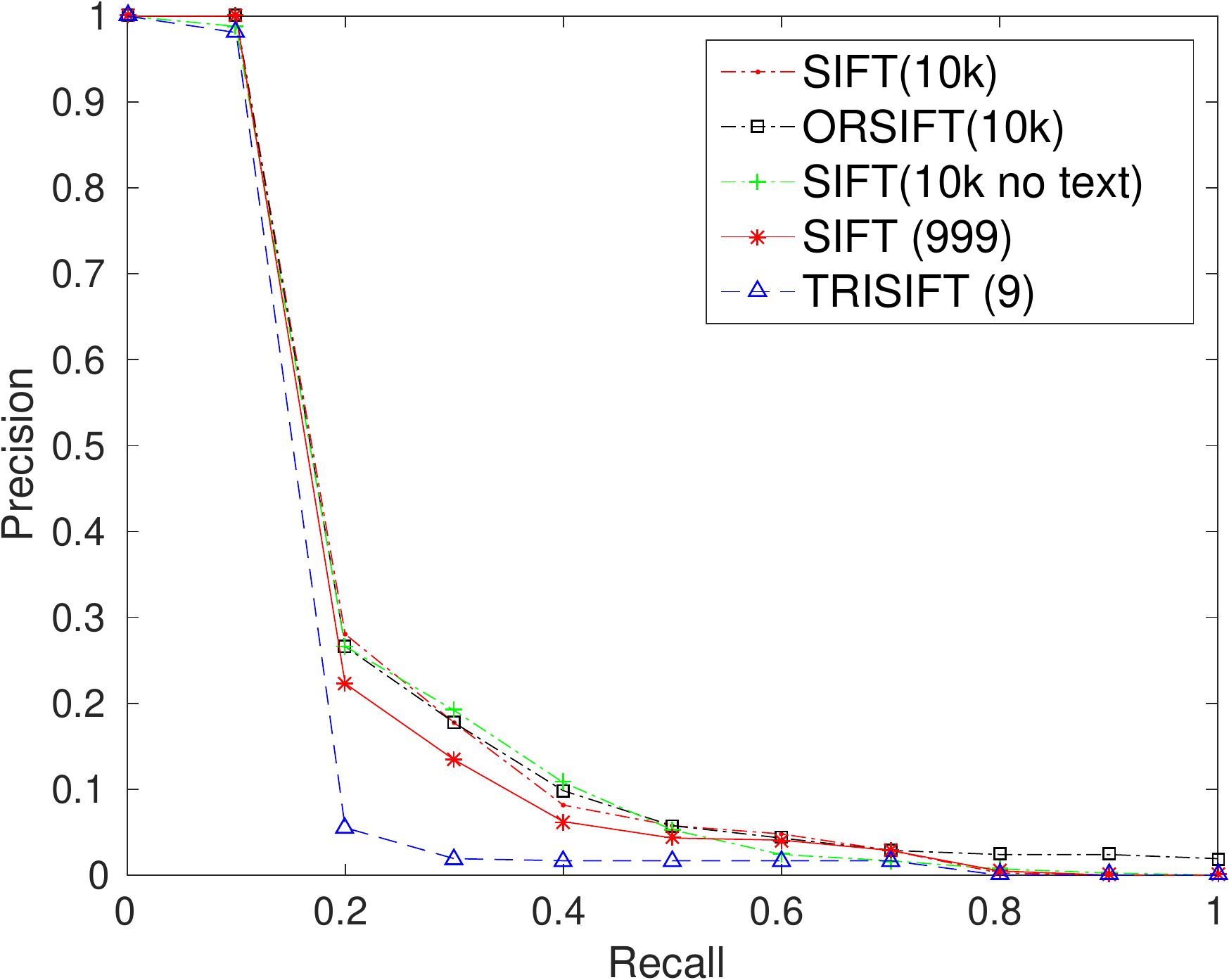}}
	\hspace{0.2cm}
	\subfloat[Zoomed]{
		\includegraphics[width = 0.45\textwidth]{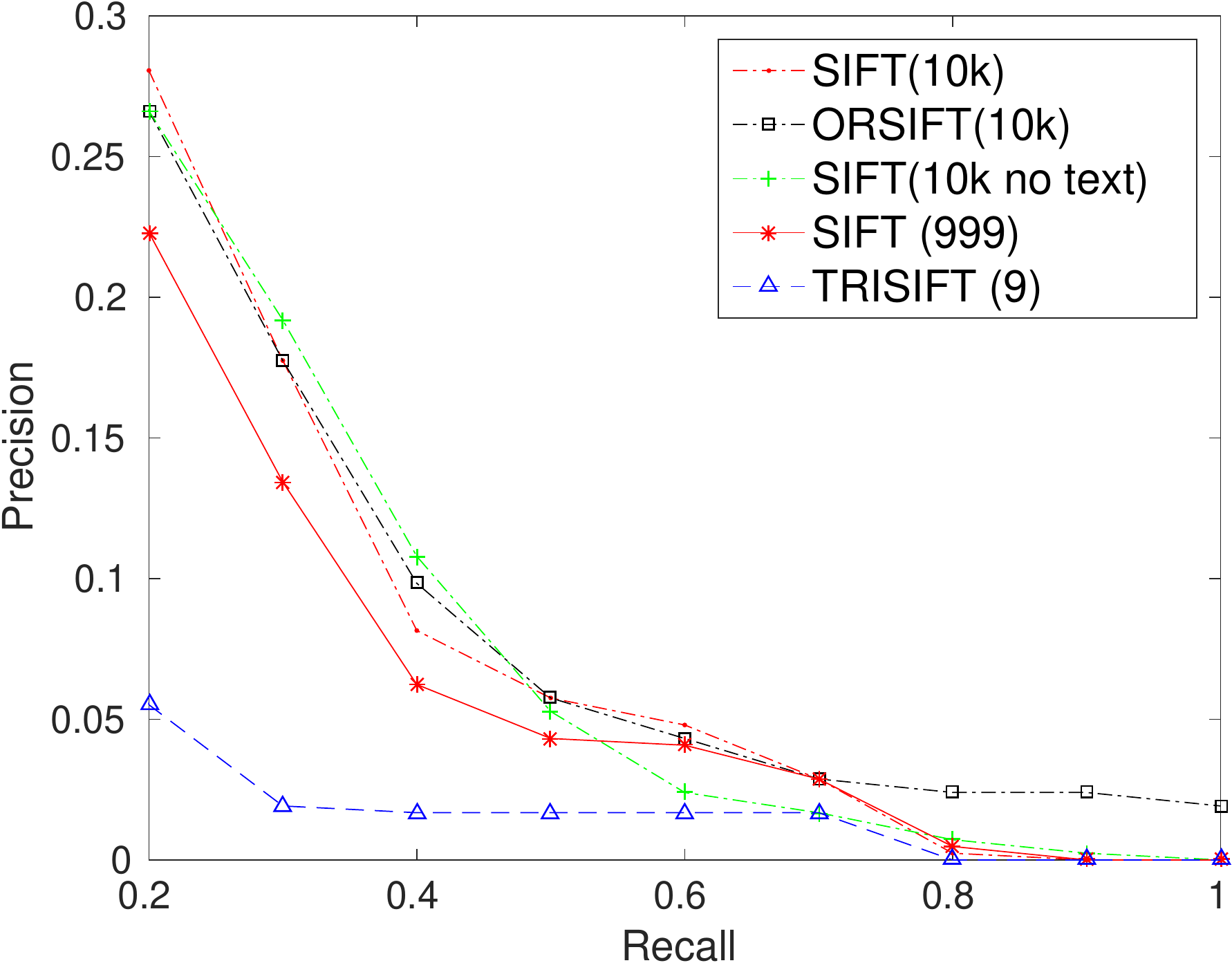}}
	\caption{Precision-recall results of SIFT and its variants.}
	\label{fig:pr_result1}
\end{figure}

\begin{figure}[!htb]
	\centering
	\subfloat[Original]{
		\includegraphics[width = 0.45\textwidth]{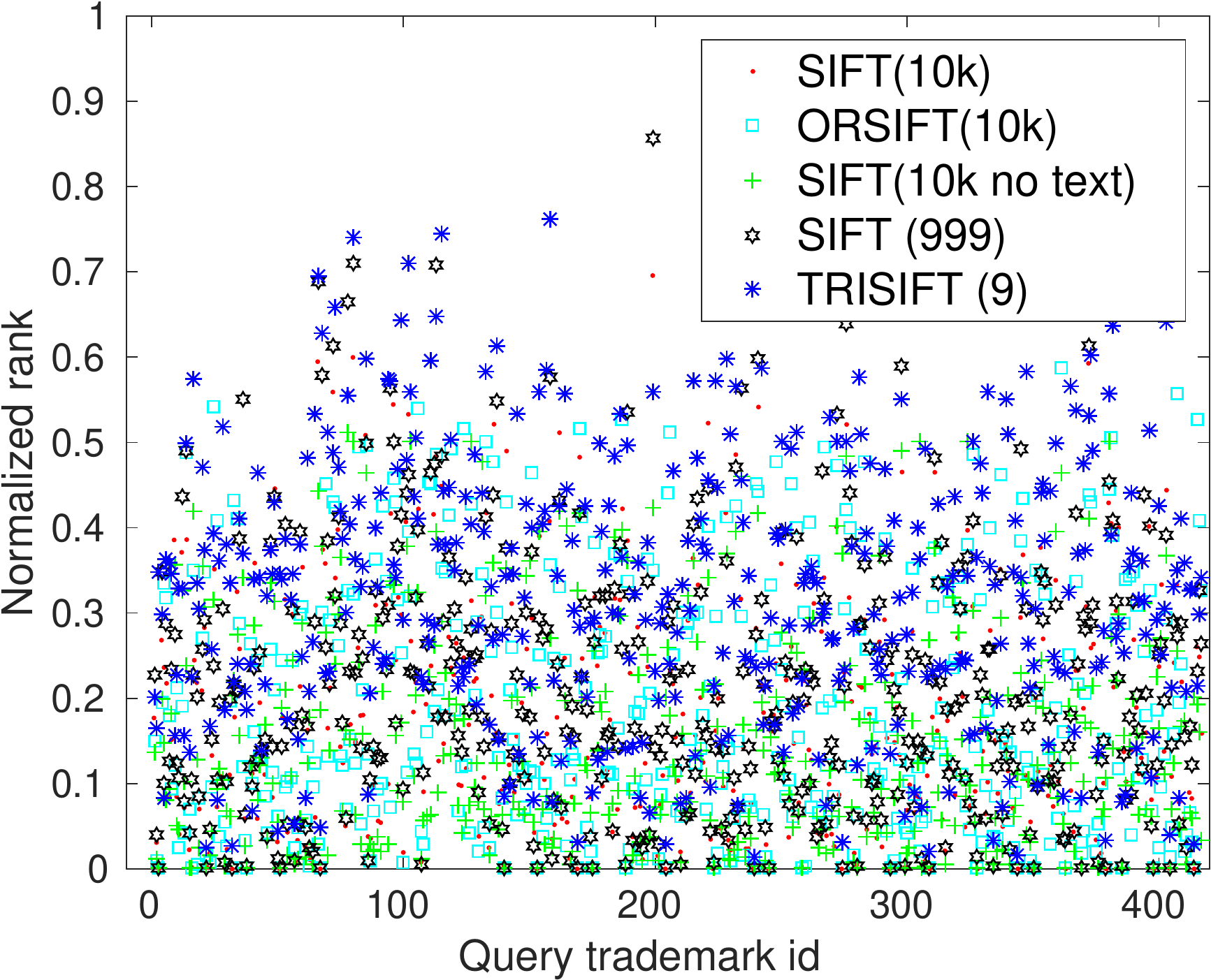}}
	\hspace{0.2cm}
	\subfloat[Zoomed]{
		\includegraphics[width = 0.45\textwidth]{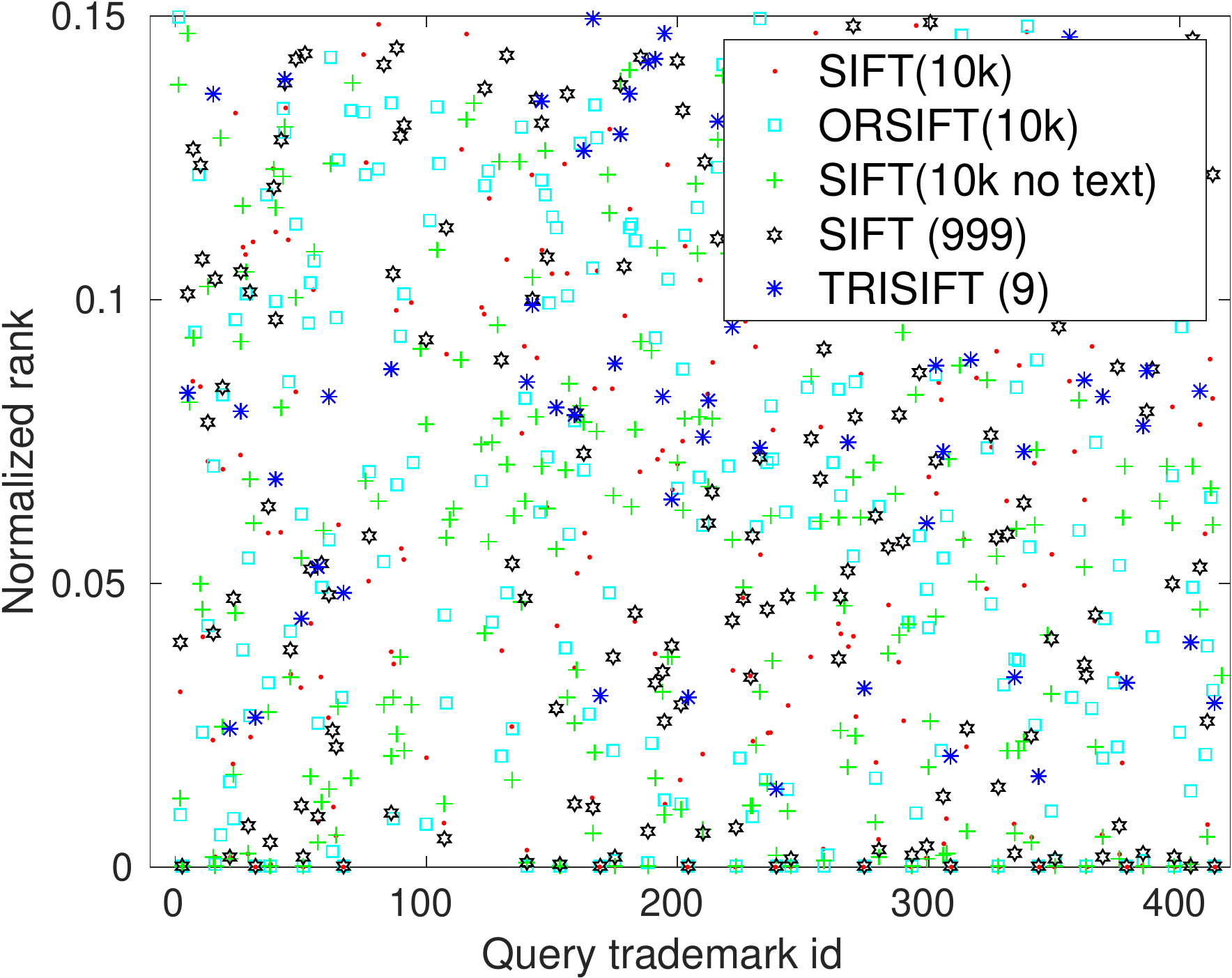}}
	\caption{Normalized average ranking results of SIFT and its variants.}
	\label{fig:point_rank1}
\end{figure}

\begin{figure}[!htb]
	\centering
	\subfloat[Original]{
		\includegraphics[width = 0.45\textwidth]{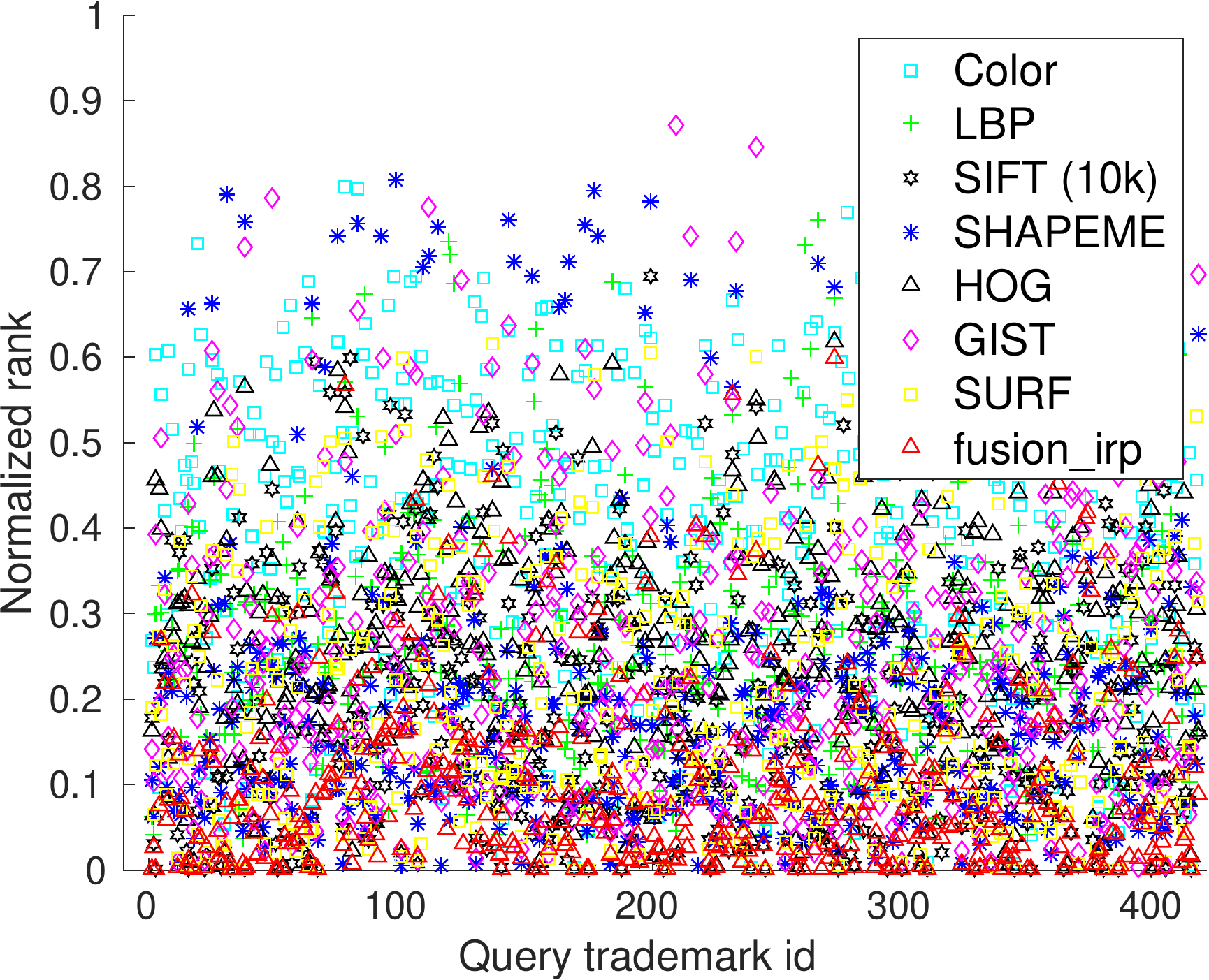}}
	\hspace{0.2cm}
	\subfloat[Zoomed]{
		\includegraphics[width = 0.45\textwidth]{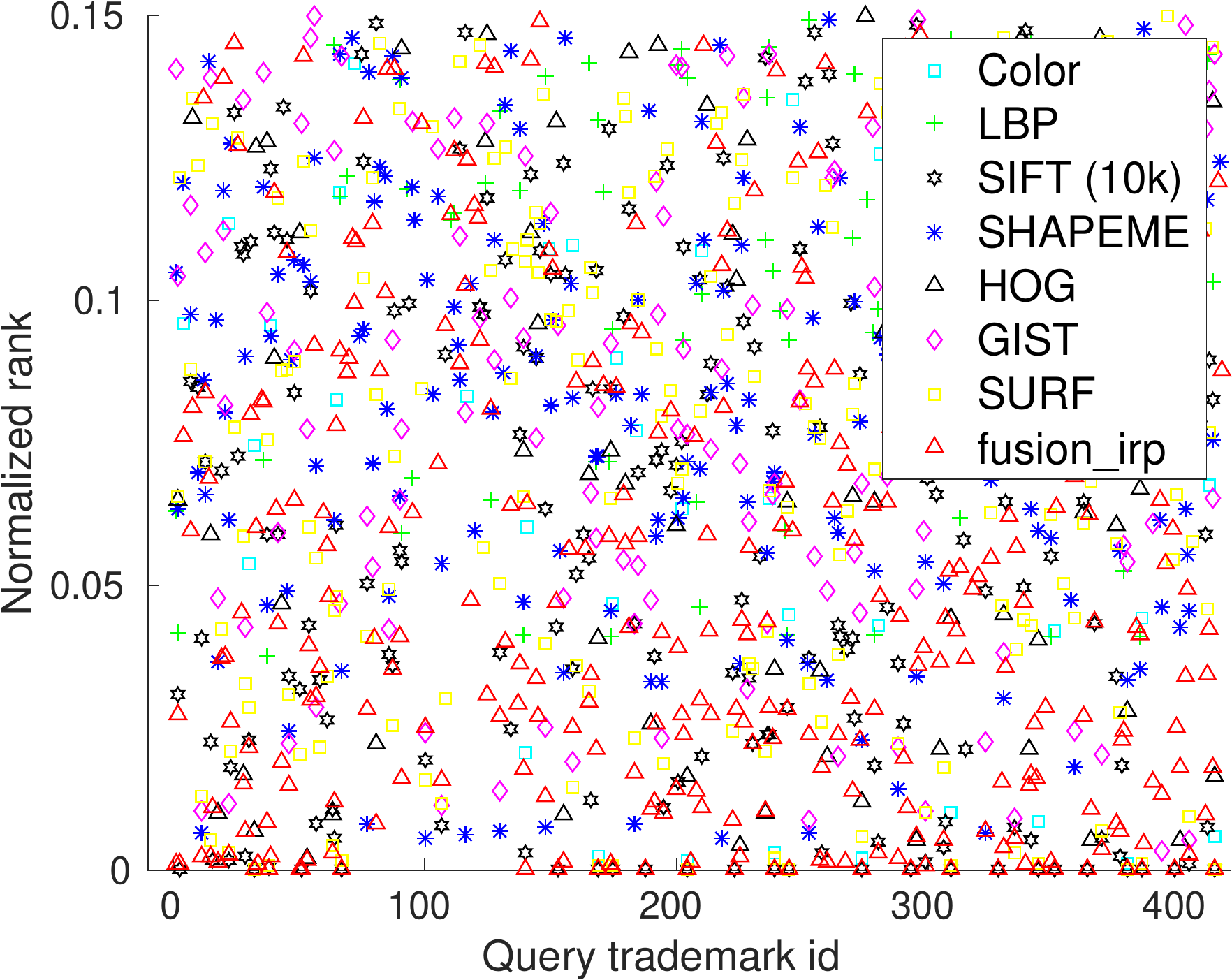}}
	\caption{Normalized average ranking results of hand-crafted features. (a) Original view, (b) Zoomed view.}
	\label{fig:point_rank2}
\end{figure}

\begin{figure}[!htb]
	\centering
	\subfloat[Original]{
		\includegraphics[width = 0.45\textwidth]{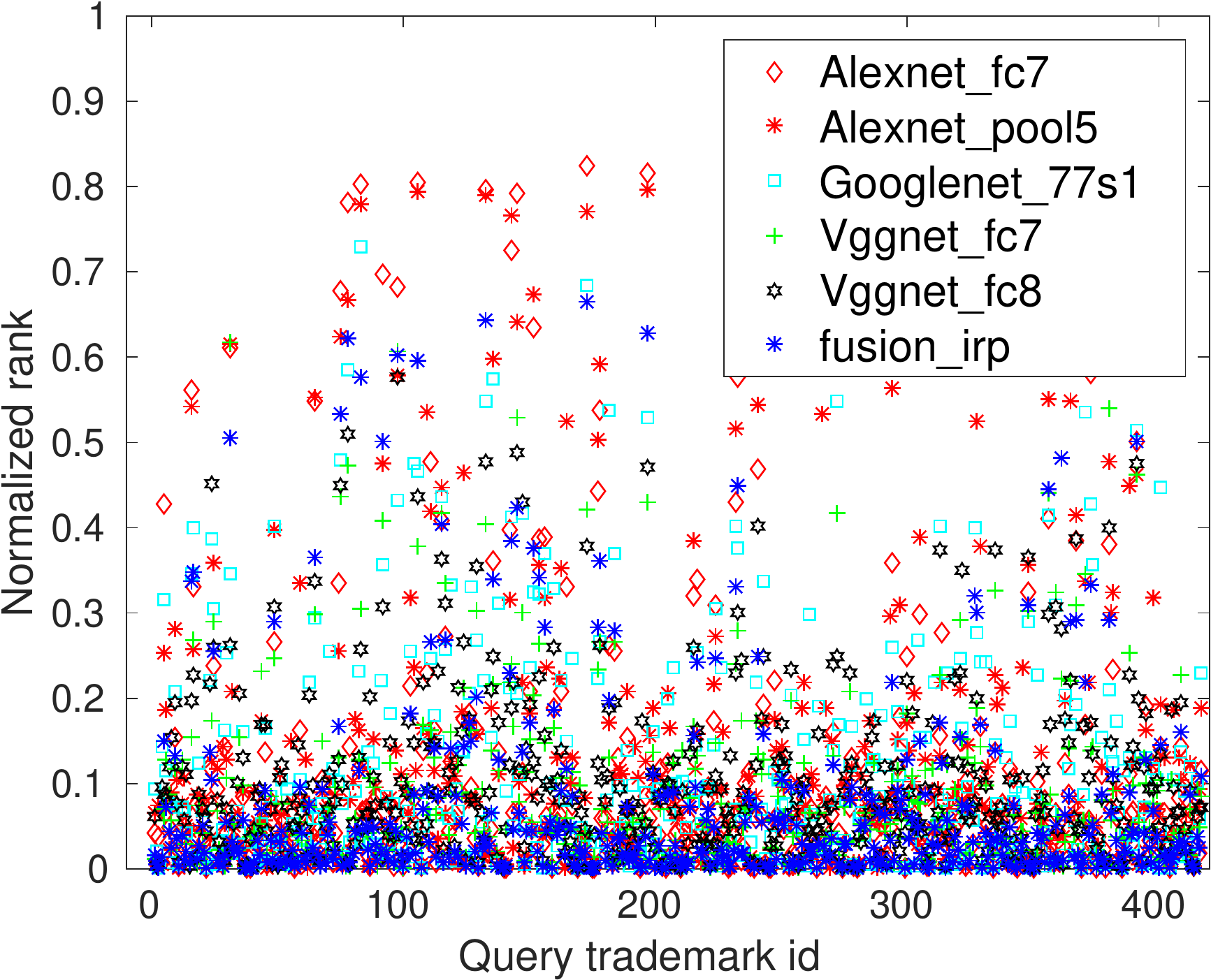}}
	\hspace{0.2cm}
	\subfloat[Zoomed]{
		\includegraphics[width = 0.45\textwidth]{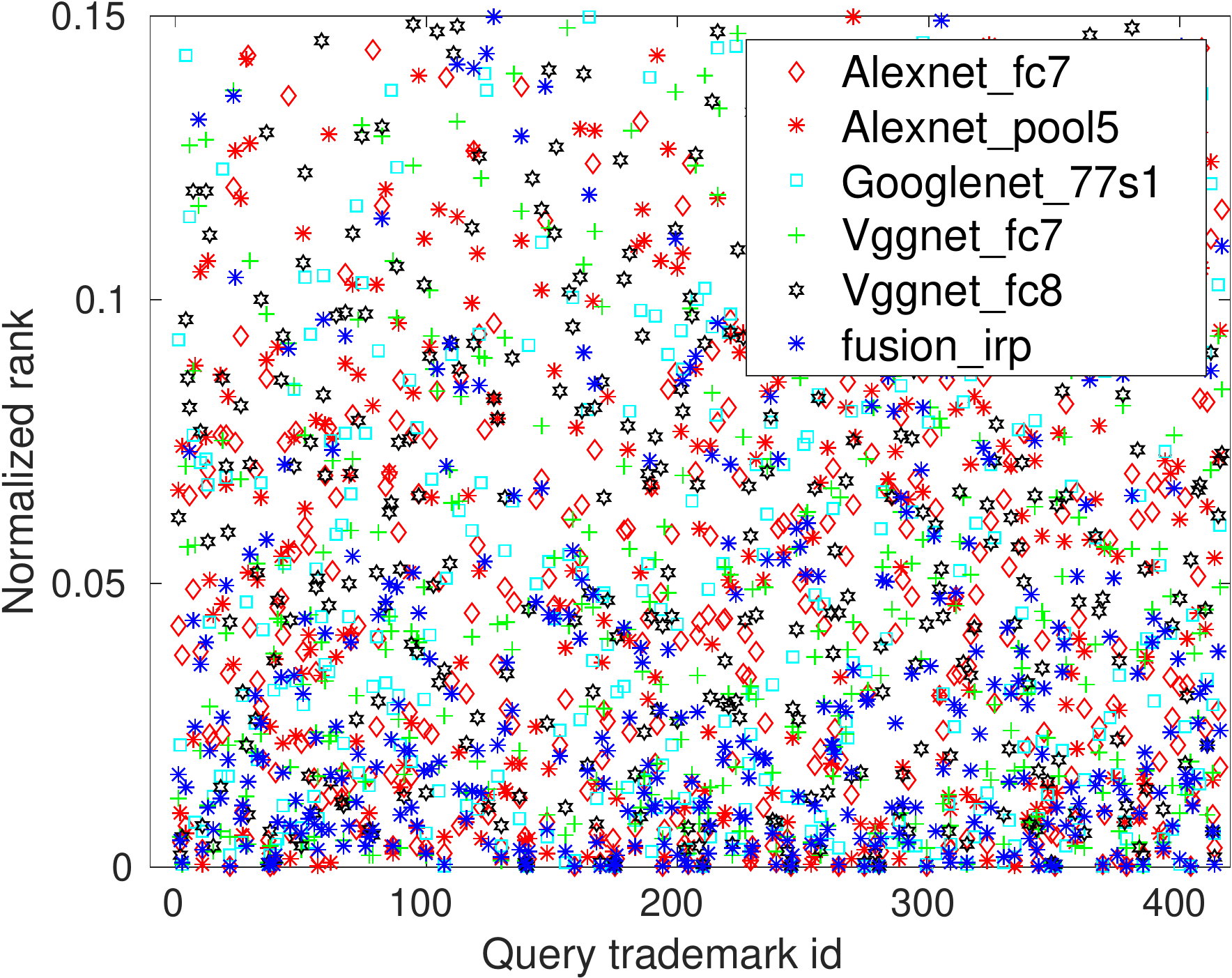}}
	\caption{Normalized average ranking results of DCNN features.(a) Original view, (b) Zoomed view.}
	\label{fig:point_rank3}
\end{figure}

\begin{figure}[!htb]
	\centering
	\subfloat[Original]{
		\includegraphics[width = 0.45\textwidth]{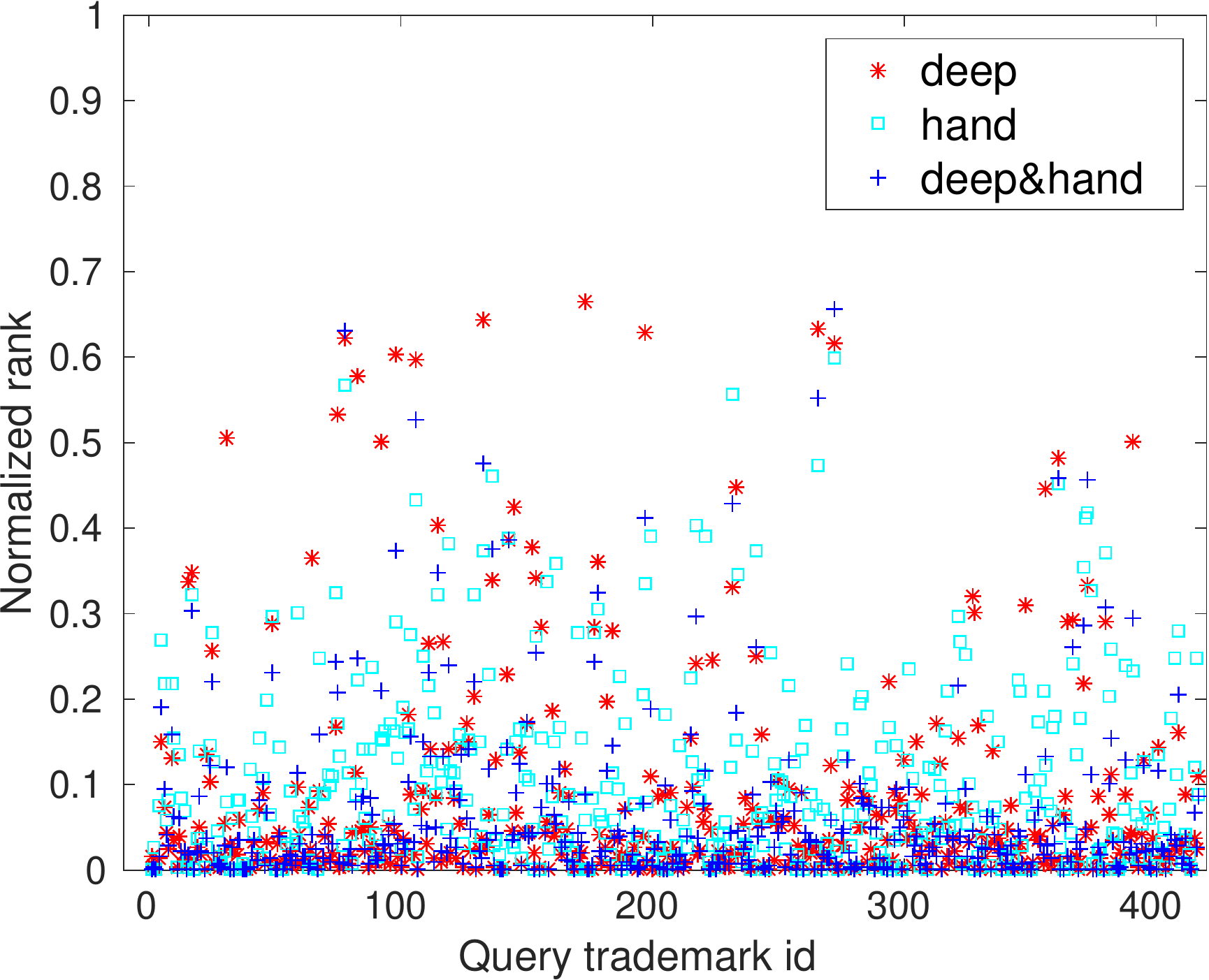}}
	\hspace{0.2cm}
	\subfloat[Zoomed]{
		\includegraphics[width = 0.45\textwidth]{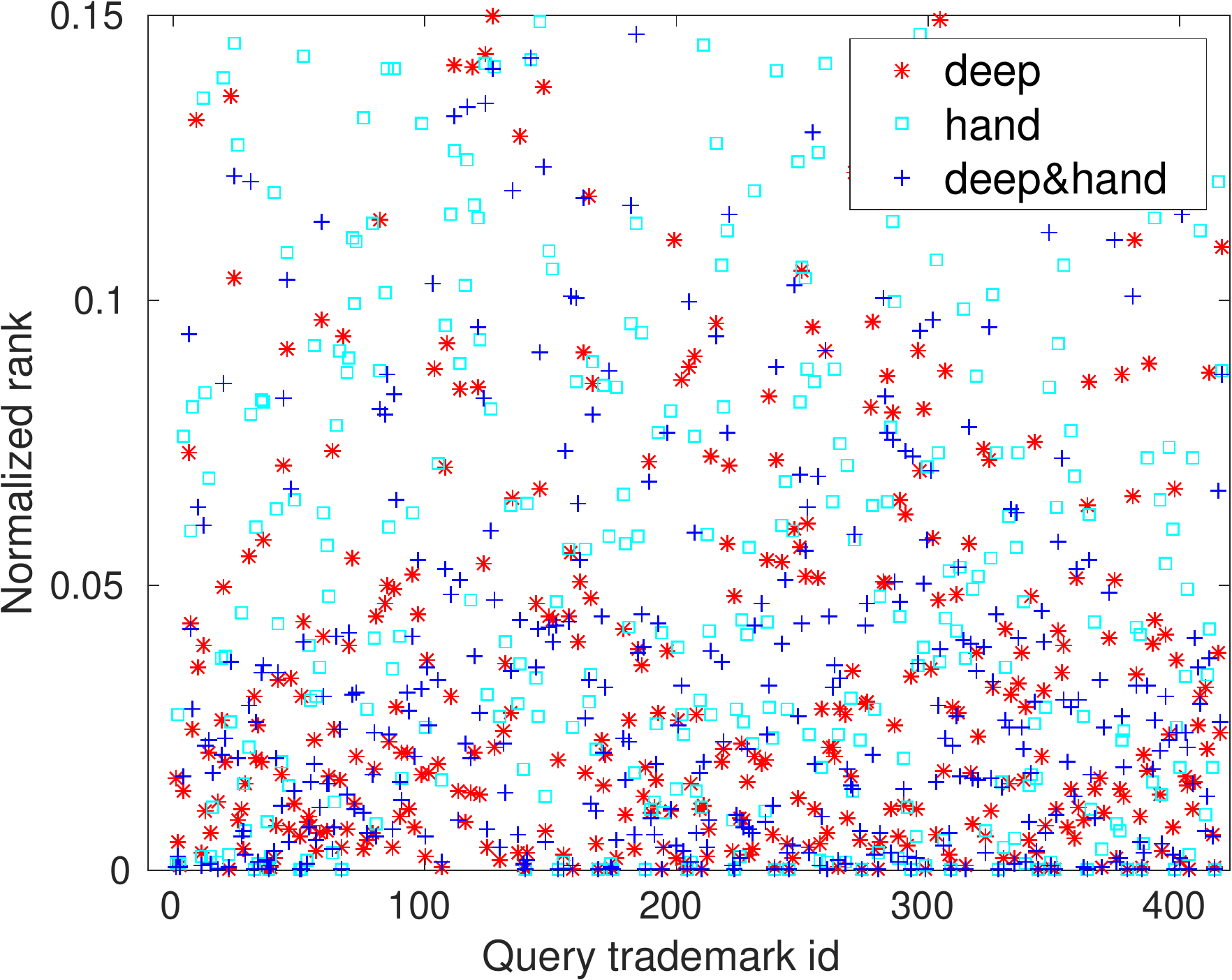}}
	\caption{Normalized average ranking results of fusion features.(a) Original view, (b) Zoomed view.}
	\label{fig:point_rank4}
\end{figure}

We display the rank results of the hand-crafted features and the CNN features in Tables \ref{tab:rank_mean} and \ref{tab:dl} respectively. These tables show mean and standard deviation values of $Rank$ and $\widetilde{Rank}$ of the implemented methods. The best method should have the smallest $Rank$ and $\widetilde{Rank}$ values. Figures \ref{fig:pr_result1}, \ref{fig:pr_result2}, \ref{fig:pr_result3} and \ref{fig:pr_result4} show the PR graphs. In these figures, each PR curve includes also a zoomed version for the sake of better visibility. Although rank results and PR curves indicate the overall performance of the method, they fail to highlight a method's performance on individual queries. For this end, we provide a display of performance on individual queries in Figures \ref{fig:point_rank1}, \ref{fig:point_rank2}, \ref{fig:point_rank3} and \ref{fig:point_rank4}. In an individual rank graph, the X-axis is the query id. The length of X-axis is 417, since we have 417 queries. The Y-axis is rank value of the each queries. When a query is given, the optimal method will return expected results with a priority. Therefore, the marks of the optimal method will be very close to X-axis, and the density of the zoomed version will be high at nearby the X-axis.  

From Table \ref{tab:rank_mean}, we see that the worst retrieval result is due to the color histogram method. This is expected since color is not sufficient for providing an overall judgment for trademark similarity. What is worse, half of the dataset are text-only trademarks, and mostly black and white. However, as shown in Figure \ref{fig:point_rank2}, we see that, although color is not sufficient, it is necessary for determining similarity for some trademarks: In fact, in some cases, color histogram results are very close to the X-axis, which means it works well on some of the queries. 

Looking at the performance of the hand-crafted features, we see that the performance of global-features (LBP, GIST) are more or less the same. However, based on our experience gained by visualizing query results, we found that the GIST method is better at capturing layout similarity, while the LBP performs better on texture similarity (not reported here). What is more, we can see that BoVW model based local feature methods yield better results than the global features. Among them, SIFT without text features ($si^4$) perform best. SIFT with 10k visual words is the second performing method. Surprisingly, TRI-SIFT does not perform better than the original SIFT since most trademarks yield insufficient number of keypoints for TRI-SIFT to make difference. This is in contrast to our previous results \cite{tuerxun2015comparison, tursun2015metu}, which is due to the fact that we handle the tie cases differently in this paper (following the literature - \cite{saha2007image, jin2005improving}), and that TRI-SIFT produces a large number of tie cases. Similarly, OR-SIFT does not outperform the original SIFT neither; however, Figure \ref{fig:point_rank1} suggests that it is better than SIFT in certain queries. 

Table \ref{tab:dl} lists the $Rank$ and $\widetilde{Rank}$ performance of the CNN based methods. We can see that their performances are far better than those of the hand-crafted features. Among the individual methods, the features extracted from FC7 layer of VGG-Net16 returns the best result. This is expected since VGG-Net is known to have learned more generic representations than GoogleNet or AlexNet (see, e.g., \cite{simonyan2014very}). However, Figure \ref{fig:point_rank3} shows that these models perform differently on individual queries, for example, AlexNet outperforms other networks on certain queries.

\subsubsection{Fusion Results}

\begin{figure}[!htb]
	\centering
	\subfloat[Original]{
		\includegraphics[width = 0.45\linewidth]{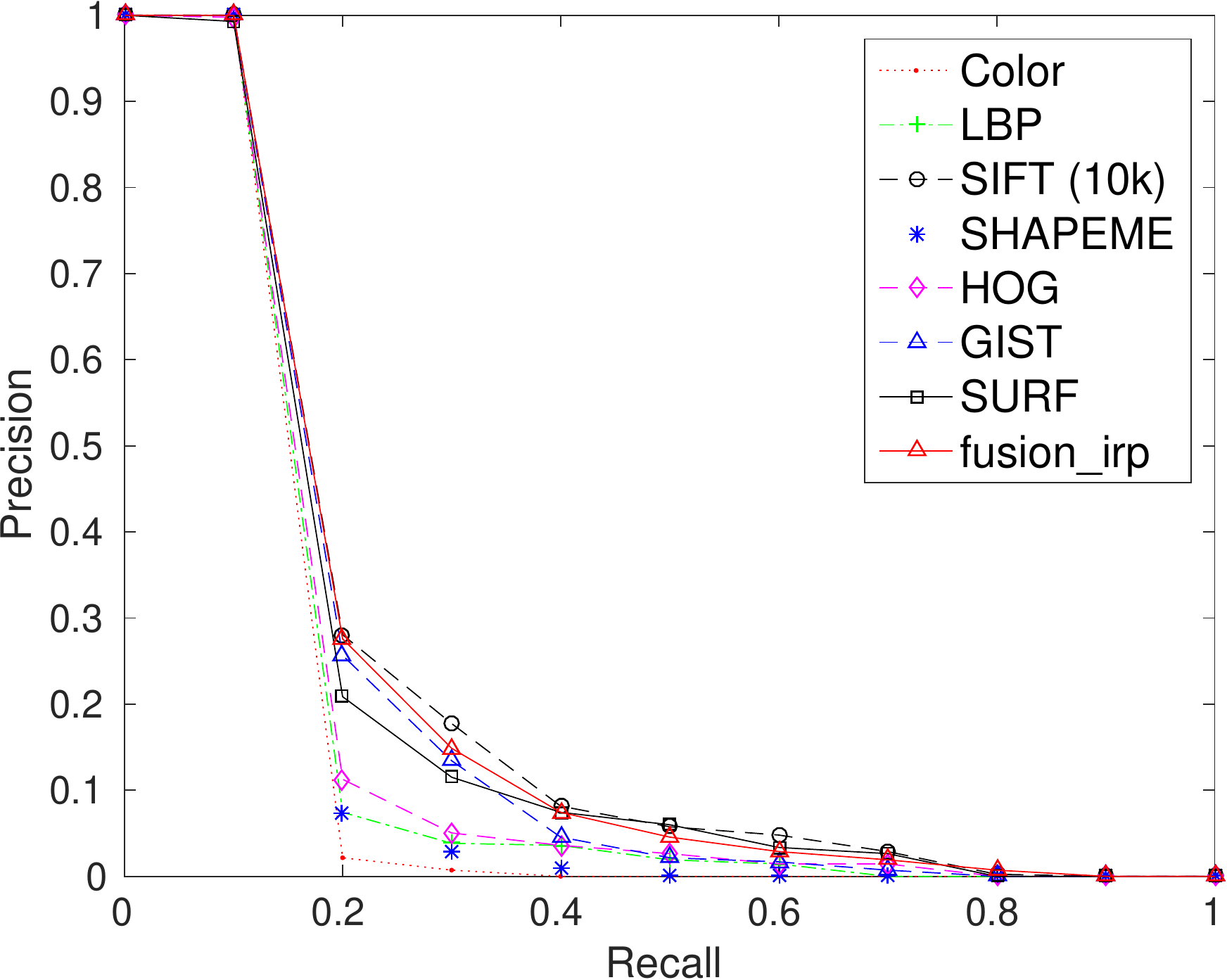}}
	\hspace{0.2cm}
	\subfloat[Zoomed]{
		\includegraphics[width = 0.45\linewidth]{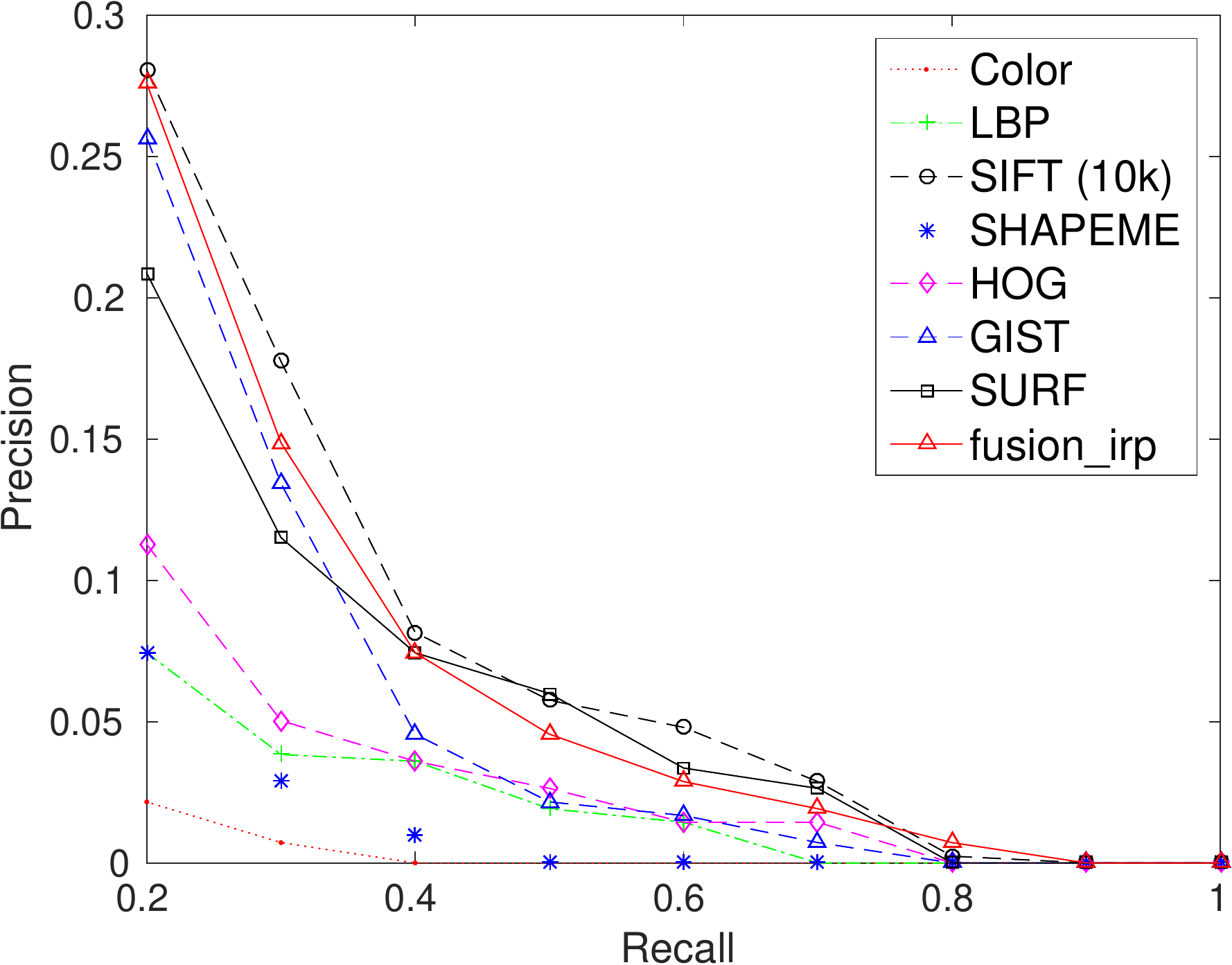}}
	\caption{Precision-recall results of hand-crafted features. (a) Original view, (b) Zoomed view.}
	\label{fig:pr_result2}
\end{figure}

\begin{figure}[!htb]
	\centering
	\subfloat[Original]{
		\includegraphics[width = 0.45\textwidth]{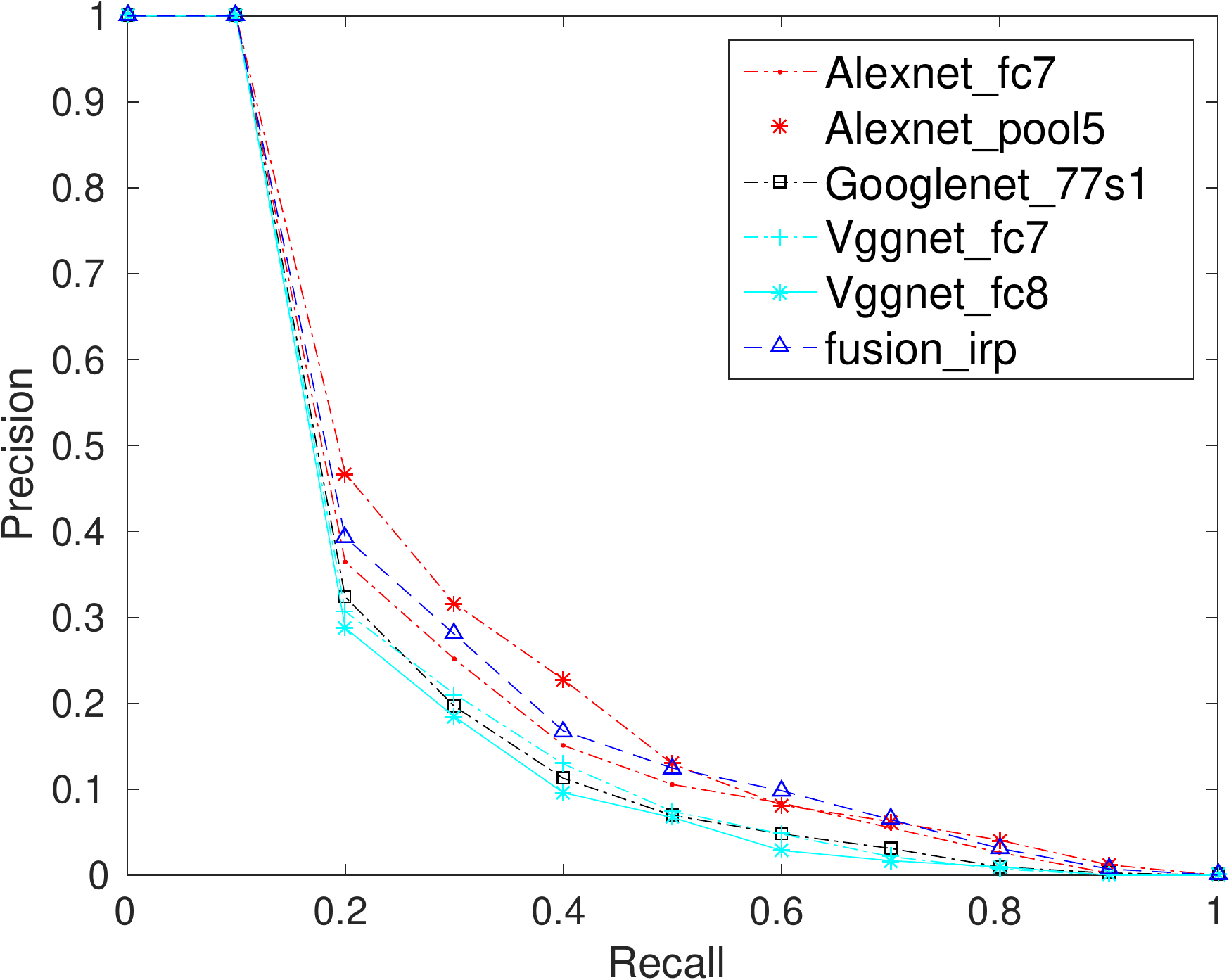}}
	\hspace{0.2cm}
	\subfloat[Zoomed]{
		\includegraphics[width = 0.45\textwidth]{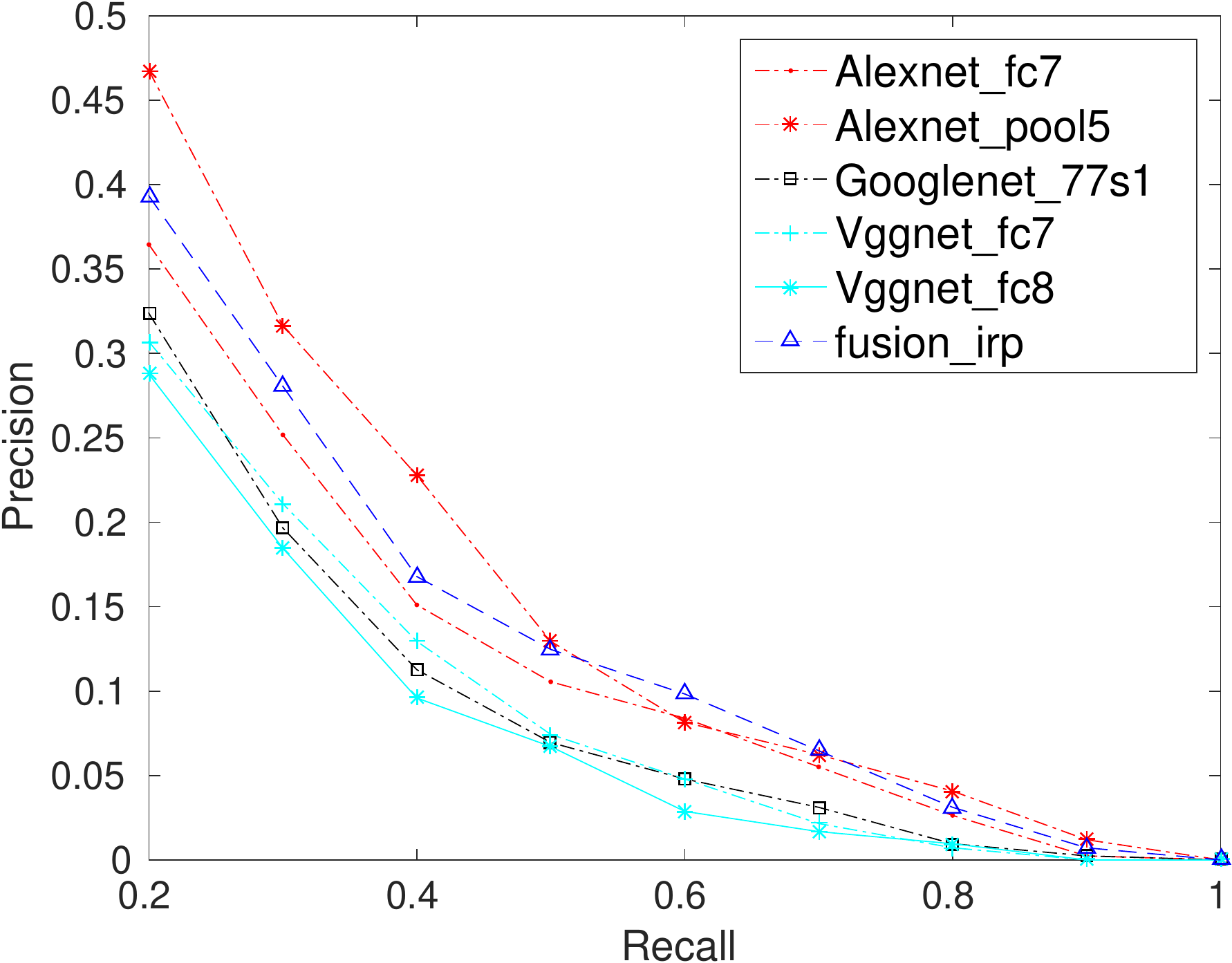}}
	\caption{Precision-recall results of DCNN features. (a) Original view, (b) Zoomed view.}
	\label{fig:pr_result3}
\end{figure}

\begin{figure}[!htb]
	\centering
	\subfloat[Original]{
		\includegraphics[width = 0.45\textwidth]{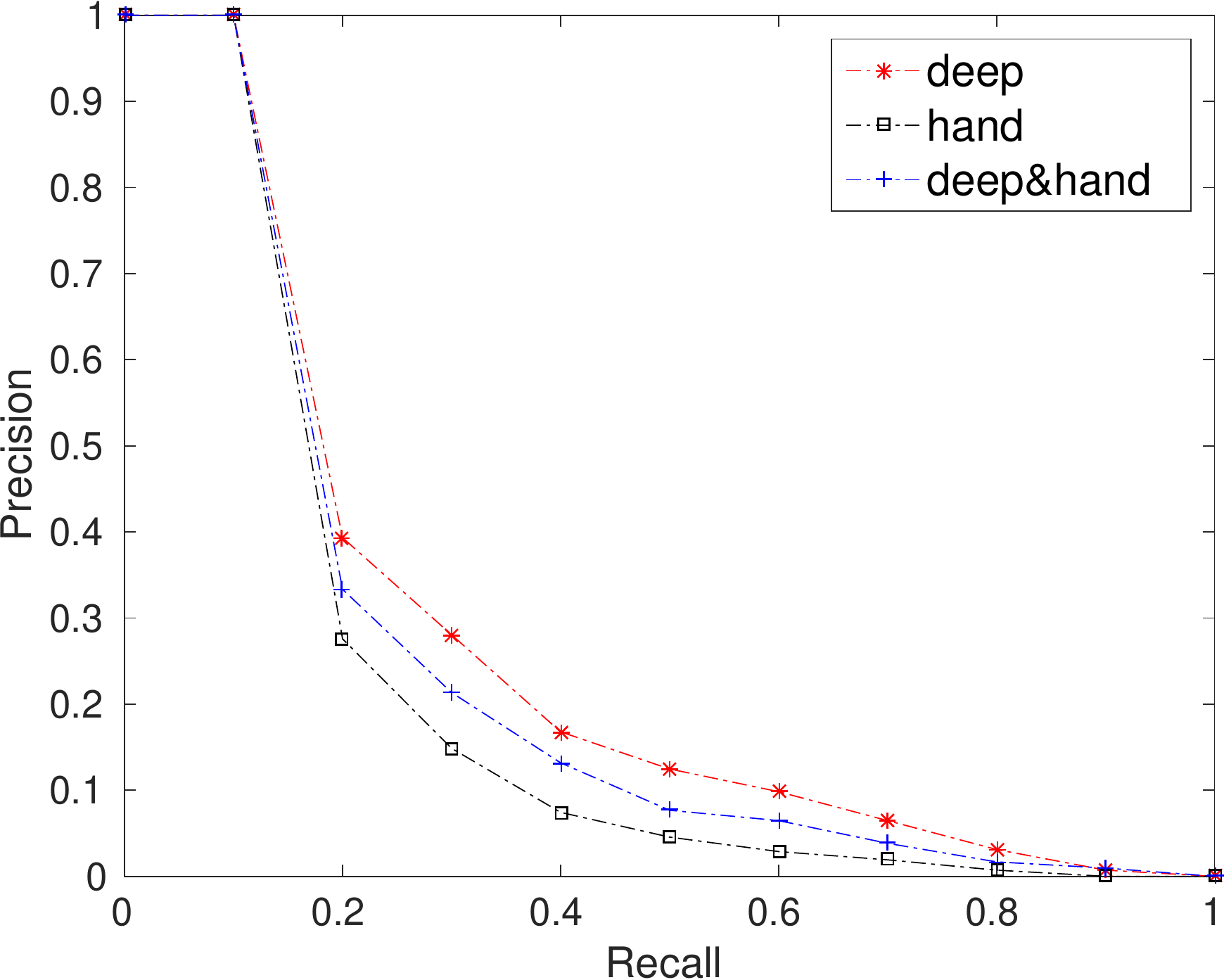}}
	\hspace{0.2cm}
	\subfloat[Zoomed]{
		\includegraphics[width = 0.45\textwidth]{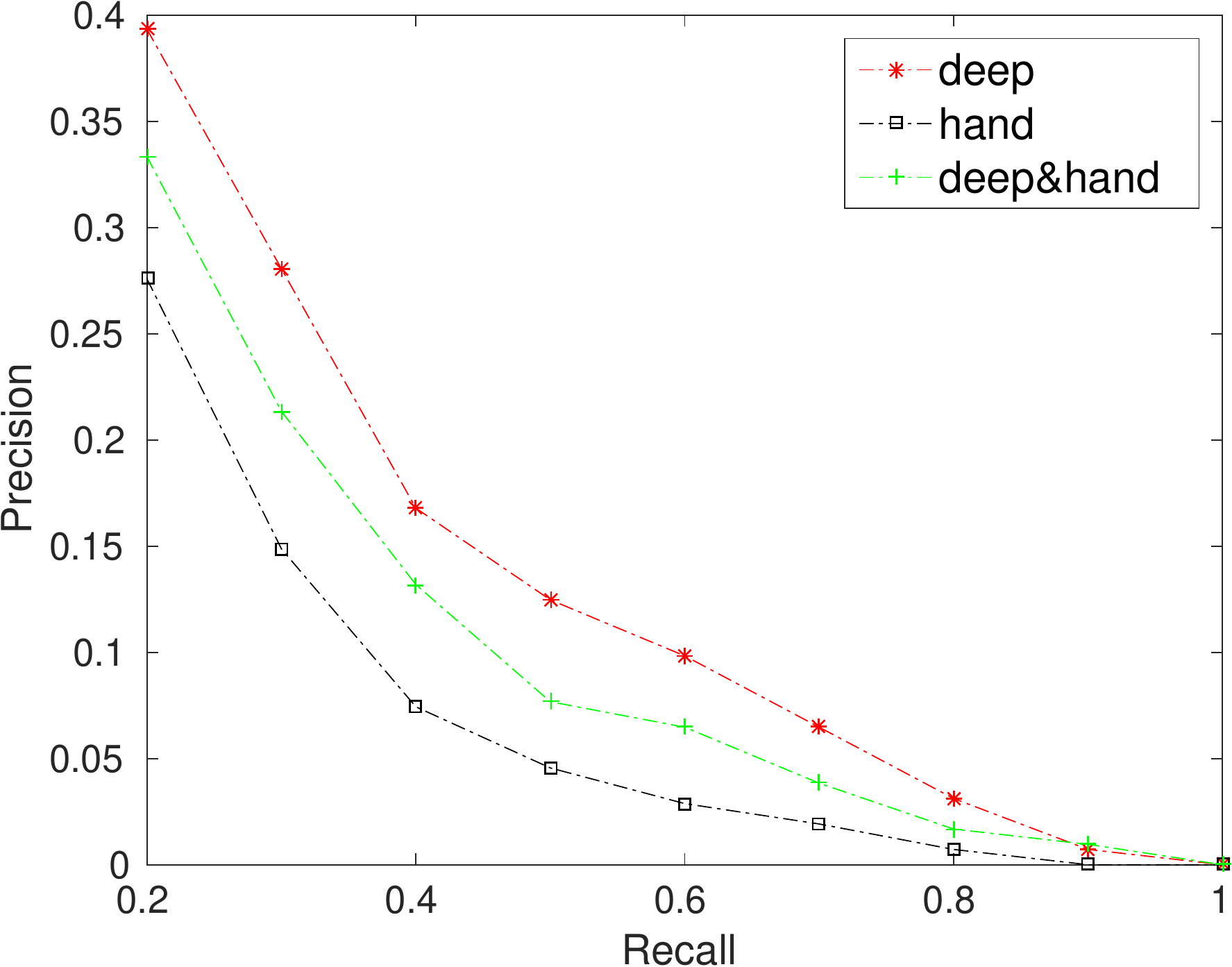}}
	\caption{Precision-recall results of fusion features. (a) Original view, (b) Zoomed view.}
	\label{fig:pr_result4}
\end{figure}

We have selected the best performing methods under each category and fused them. Looking at the fusion results in Table \ref{tab:fs}, fusion improves the performance substantially. With a simple and efficient fusion method like IRP, we observe a clear improvement in both hand-designed features and learned features. In fact, fusing together the fusion of hand-crafted and deep features (denoted $f^3$ in the Table \ref{tab:fs}) yields the best performance among the tested methods. However, looking at the precision and recall values in Figures \ref{fig:pr_result2}, \ref{fig:pr_result3} and \ref{fig:pr_result4}, we see that fusion leads to slight decrease in precision. This is mainly due to the fact that fusion discovers similar logos not anticipated by us.

\begin{table}[tbh]
	\caption{Comparison of the results of the CNN methods.}
	\label{tab:dl}
	\vspace*{-0.2cm}
	\begin{center}
		\resizebox{\textwidth}{!}{
		\begin{tabular}{c|c|c| l r}
			\hline
			\multirow {2}{*}{\bf Net} 	& \multirow {2}{*}{\bf Layer}  & \multirow {2}{*}{\bf Size} 	& \multirow {2}{*}{\bf Average rank} 	 & \textbf{Normalized }\\
						 					&			 					&							& 										& \bf{average rank}\\ \hline
			\hline
			AlexNet	($ax^1$)		& FC7	& 4,096	 & 103,549.2 $\pm$ 157,877.9			 & 0.112 $\pm$ 0.171 \\
			AlexNet	($ax^2$)		& Pool5	& 9,216	 & 125,300.9 $\pm$ 157,739.5			 & 0.136 $\pm$ 0.171 \\  
			GoogLeNet ($gl^1$)		   	& 77S1	& 1,024	 & 108,662.5 $\pm$ 127,619.1			 & 0.118 $\pm$ 0.138\\
			VggNet16 ($vg^1$)			& Pool5	& 25,088 & 88,829.1 $\pm$  112,370.7			 & 0.096 $\pm$ 0.122\\ 
			VggNet16 ($vg^2$)		   	& FC7	& 4,096	 & \bf 79,538.5 $\pm$ 98,961.3			 &\bf 0.086 $\pm$ 0.107\\
			VggNet16 ($vg^3$)		  	& FC8		& 1,000	 	 & 98,716.9 $\pm$ 100,910.4			 	 & 0.107 $\pm$ 0.109\\
			Autoencoder ($ae^1$)	 	& Last	&288	 &287,884.8 $\pm$ 157,787.7	 &0.312 $\pm$ 0.171 \\
			Autoencoder ($ae^2$)		& 	Last	&8,192	 &209,029.0 $\pm$ 142,507.9	 &0.226  $\pm$ 0.154\\
			\hline
			
		\end{tabular}
		}
	\end{center}
\end{table}

\begin{table}[tbh]
	\caption{Comparison of the results of fusions. 	\label{tab:fs}}
	\vspace*{-0.2cm}
	\begin{center}
		\resizebox{\textwidth}{!}{
		\begin{tabular}{c|c|c| c c}
			\hline
			\multirow {2}{*}{\bf Fusion} 	& \multirow {2}{*}{\bf Method}  & \multirow {2}{*}{\bf Items} 	& \multirow {2}{*}{\bf Average rank} 	 & \textbf{Normalized }\\
			&			 					&							& 										& \bf{average rank}\\ \hline
			\hline
			Fusion ($f^1$)				& IRP	 		& $cl, lp, sh, gs, si^1, su$ 	&   96,545.1 $\pm$ 100,474.7		& 0.105   $\pm$ 0.109 \\
			Fusion ($f^2$)			   & IRP		   & $ax^1, gl^1, vg^2$ 			&  73,239.0 $\pm$ 11,7881.2			 & 0.079 $\pm$ 0.128 \\
			Fusion ($f^3$)			   & IRP		    & $f^1, f^2$ 						&	\bf 56,844.1 $\pm$	87,794.1		  & \bf 0.062 $\pm$	0.095								\\	
			\hline
		\end{tabular}
		}
	\end{center}
\end{table}

%
%

\subsubsection{Time and Memory Aspects}

Tables \ref{tab:fetime} and \ref{tab:fea_mem} compare running time and memory aspects of the tested methods respectively. Running time measures three phases: feature extraction, feature processing, and ranking. CNN features have the fastest feature extraction phase because of GPU parallelization. Feature processing time is the extra time we spend for steps like vectorization, text removal, and feature grouping etc. The ranking time contains similarity calculation and sorting times. 

In our experiments, each query is compared with all other trademarks in the dataset, and the trademarks are sorted by similarity for retrieving the top $m$ results. Looking at Table \ref{tab:fetime}, we see that the maximum time for querying a trademark in our dataset is about 17 seconds. Although this is a realistic figure, it can be improved even further since our tests were conducted in MATLAB. Moreover, we see opportunities for further improvement by parallelizing the feature matching phase.

In large-scale trademark retrieval, the descriptor size becomes an important factor. Table \ref{tab:fea_mem} compares the size of the descriptors as a measure for the require memory. We see that the key-point based methods have large descriptor sizes whereas global features have smaller sizes. CNN features have sizes between those of the local and the global features, depending on the number of detected key-points.

\begin{table}[tbh]
	\caption{Comparison of running times of the tested methods (in seconds).}
	\vspace*{-0.2cm}
	\begin{center}
		\resizebox{\textwidth}{!}{
			\begin{tabular}{l l l r r r r}
				\hline
				\multirow{3}{*}{\bf Algorithm} & \multirow {3}{*}{\bf Cluster} & \bf Feature 		& \bf Feature	& \bf Get Rank  & \bf Total \\
													&								& \bf extraction  	& \bf process	 & \bf results  & \bf calculation \\
													&								& \bf time 			& \bf time	 	&\bf	time  	& \bf time \\
				\hline
				\hline
				Color			&-				& 0.0364	& -			& \bf0.2034	& \bf0.2398\\
				LBP				&-				& 0.0309	& -			& 1.6609	& 1.6918\\
				GIST 			&-				& 0.1638 	& -			& 2.0623	& 2.2261\\
				HOG 			& 10k		 	& 0.0545 	& 0.0076	& 16.5227	& 16.5849\\
				SIFT 			& 10k			& 0.2232	& 0.0265	& 16.5227    & 16.7725\\
				SIFT 			& 999			& 0.2232	& 0.0030	& 16.5227 	& 16.7490\\
				Tri-SIFT 		& 9				& 0.2232	& 0.3477	& 2.3770	& 2.9479\\
				OR-SIFT 		& 10k			& 0.0540	& 0.0118	& 16.5227	& 16.5886\\
				SIFT (WoT) 		& 10k			& 0.2232	& 0.2029  	& 16.5227	& 16.9489\\
				SURF 			& 10k			& 0.0440	& 0.0120 	& 16.5227	& 16.5786\\
				SHAPEMES 		& 10k			& 0.1197	& 0.0110	& 16.5227	& 16.6534\\
				\hline
				Alexnet			& FC7			&0.0123		& -			& 6.0389	& 6.0512\\
				Alexnet			& Pool5			&\bf0.0111 	& -			& 15.8960	& 15.9066\\
				GoogLenet		& 77s1			&0.0240		& -			& 2.4430	& 2.4670\\
				Vggnet			& FC7			&0.0678		& -			& 6.0389	& 6.1067\\
				Vggnet			& FC8			&0.0692		& -			& 2.3770	& 2.4462\\
				\hline
			\end{tabular}
		}
	\end{center}
	\label{tab:fetime}
\end{table}

\begin{table}[tbh]
	\caption{Comparison of the sizes of various features. The ``single size" is the size of original features. The BoVW feature size is the size after BoVW quantization. $n$ denotes the number of keypoints detected.}
	\vspace*{-0.2cm}
	\begin{center}
			\begin{tabular}{l c c c}
				\hline
				\multirow{2}{*}{\bf Algorithm} & \bf Cluster 	& \textbf{Size} & \textbf{Size}\\
											   & \bf /Type   & (single)	   & (BoVW)		\\	  
				\hline
				Color			&-		& $1\times72$ 	&-		 \\
				LBP				&-		& $1\times256$ 	&-		 \\
				GIST 			&-		& $1\times512$ 	&-		 \\
				HOG 			& 10k	& $n\times36$ 	& $1\times10,000$ \\
				SIFT 			& 10k	& $n\times128$ 	& $1\times10,000$ \\
				SIFT 			& 999	& $n\times128$	& $1\times999$	 \\
				Tri-SIFT 		& 9		& $n\times128$	& $1\times998$	 \\
				OR-SIFT 		& 10k	& $n\times64$ 	& $1\times10,000$ \\
				SURF 			& 10k	& $n\times64$ 	& $1\times10,000$ \\
				SHAPEMES 		& 10k	& $n\times60$ 	& $1\times10,000$ \\
				\hline
				Alexnet			& FC7	& $1\times4096$	&-	     \\
				GoogLenet		& 77s1	& $1\times1024$	&-		 \\
				Vggnet			& FC7	& $1\times4096$	&-	     \\
				Vggnet			& FC8	& $1\times1000$	&-	     \\
				\hline
			\end{tabular}
	\end{center}
	\label{tab:fea_mem}
\end{table}

%% file: section/conclusion.tex
\section{Conclusion}
\label{sect:conclusion}

In this work, we introduced a large scale dataset and benchmark for trademark retrieval, and provided a baseline for the problem by evaluating the state of the art hand-crafted and CNN features. We found that CNN features are the best for logo retrieval problem in terms of not only performance but also running-time and memory. However, our results suggest that the performances of the existing methods are far from \textit{replacing} human experts in trademark retrieval, if not helping them.

We hope that the benchmark solicits further research into the trademark retrieval problem, improving the performances of the current systems, addressing the challenges addressed in the paper. We also suggest that trademark retrieval should be one of the challenges that the computer vision and pattern recognition community pays more attention to since it bears challenges and issues that have not been yet addressed properly.

\section{Acknowledgments}
We would like to thank Usta Bilgi Sistemleri A.\c{S}. and Grup Ofis Marka Patent A.\c{S}. for kindly providing nearly 1 million logos for this research and making it available to the community. This work is supported by the Ministry of Science, Turkey, under the project SANTEZ-0029.STZ.2013-1. 

We also gratefully acknowledge the support of NVIDIA Corporation with the donation of the Tesla K40 GPU used for this research.  


%% file: section/appendix_dist.tex
\section{Distance metrics}
\label{sec:appendixA}

The definition of the evaluated distance metrics are provided below for the sake of simplicity and completeness ($\bf{p}$, $\bf{q}$ are two vectors in $\Re^n$):

\begin{description}
\item[Euclidean] \hfill \\
	\begin{equation}
		\label{eq:euclidean}
		d\left (\bf{p} , \bf{q} \right ) =  \sqrt{\sum_{i=1}^n(p_i^2 - q_i^2)}.
	\end{equation}
\item[Cosine] \hfill \\
	\begin{equation}
		\label{eq:cosine}
		d\left (\bf{p} , \bf{q} \right ) =  \frac{\sum_{i=1}^n(p_i \cdot q_i)}{\norm{\bf{p}} \cdot\norm{\bf{q}}}.
	\end{equation}
\item[Intersection (L1)] \hfill \\
\begin{equation}
	\label{eq:intersection1}
	{d}\left(\bf{p}, \bf{q} \right ) =  1- \frac{\sum_{i=1}^n\min\left(p_i, q_i\right)}{\min\left(\norm{\bf{p}}, \norm{\bf{q}}\right)}.
\end{equation}

\item[Intersection (L2)] \hfill \\
\begin{equation}
	\label{eq:intersection2}
	{d}\left(\bf{p}, \bf{q} \right ) =  1- \sqrt{\sum_{i=1}^n\min\left(p_i^2, q_i^2\right)}.
\end{equation}
%
\item[Quadratic] \hfill \\
\begin{equation}
	\label{eq:quadratic}
	d \left (\bf p,\bf q \right ) = {\left (\bf p - \bf q \right )} ^ {t} \bf A(\bf p - \bf q).
\end{equation}

\item[Manhattan] \hfill \\
\begin{equation}
	\label{eq:manhattan}
	d\left (\bf{p} , \bf{q} \right ) =  \sqrt{\sum_{i=1}^n(p_i - q_i)}.
\end{equation}

\end{description}

\section{Normalization}
\label{sec:appendixB}
To calculate the distance between two vectors at various scales, appropriate normalization methods are necessary. Here, we present the definitions of the two normalization methods implemented in the paper:
\begin{description}
	\item[L1 normalization] \hfill \\
	\begin{equation}
	\label{eq:l1-norm}
	\textbf{h}_1 = \textbf{h}/{\sum_{i=1}^n {h(i)}}.
	\end{equation}

	\item[L2 normalization] \hfill \\
	\begin{equation}
	\label{eq:l2-norm}
	\textbf{h}_2 =  {\textbf{h}}/{\sqrt {\sum_{i=1}^n {h(i)^2}}}.
	\end{equation}	
\end{description}